\documentclass{article}
\usepackage{jfrExamplee}
\usepackage{graphicx}
\usepackage{apalike}
\usepackage{setspace}
\usepackage{scalerel}
\usepackage{algorithm}
\usepackage{algorithmic}
\usepackage{amsmath} 
\usepackage{bm}
\usepackage{amssymb}
\usepackage[dvipdfmx]{color}
\usepackage{cite}
\usepackage{mathtools}

\newcommand{\figref}[1]{{Fig.~\ref{#1}}}

\newcommand{\equref}[1]{{Eq.~\ref{#1}}}

\newcommand{\refsec}[1]{Sec. \ref{sec:#1}}
\newcommand{\secref}[1]{\refsec{#1}} 

\newif\ifdraft
\drafttrue 

\title{Versatile Multilinked Aerial Robot with Tilting Propellers: Design, Modeling, Control and State Estimation for Autonomous Flight and Manipulation}

\author{
Moju Zhao, Tomoki Anzai, Fan Shi, Toshiya Maki, Takuzumi Nishio,  \\ {\bf Keita Ito, Naoya Kuromiya, Kei Okada, Masayuki Inaba} \\
Department of Mechano-Infomatics\\
The University of Tokyo\\
7-3-1 Hongo, Bunkyo-ku, Tokyo 113-8656, Japan \\
\texttt{chou@jsk.t.u-tokyo.ac.jp}
}

\begin{document}

\maketitle

\begin{abstract}
  Multilinked aerial robot is one of the state-of-the-art works in aerial robotics, which demonstrates the deformability benefiting both maneuvering and manipulation. 
  However, the performance in outdoor physical world has not yet been evaluated because of the weakness in the controllability and the lack of the state estimation for autonomous flight. Thus we adopt  tilting propellers to enhance the controllability. The related design, modeling and control method are developed in this work to enable the stable hovering and deformation. Furthermore, the state estimation which involves the time synchronization between sensors and the multilinked kinematics is also presented in this work to enable the fully autonomous flight in the outdoor environment. Various autonomous outdoor experiments, including the fast maneuvering for interception with target, object grasping for delivery, and blanket manipulation for firefighting are performed to evaluate the feasibility and versatility of the proposed robot platform. To the best of our knowledge, this is the first study for the multilinked aerial robot  to achieve the fully autonomous flight and the manipulation task in outdoor environment. We also applied our platform in all challenges of the 2020 Mohammed Bin Zayed International Robotics Competition, and ranked third place in Challenge 1 and sixth place in Challenge 3 internationally, demonstrating the reliable flight performance in the fields.

\end{abstract}

\section{Introduction}
\label{sec:introduction}

In recent years, the development of aerial robot in field robotics has been greatly enhanced, and applications are diverse, ranging from autonomous exploration \cite{disaster-response-JFR} and data collection \cite{aerial-water-sampling-JFR} to provision of commercial services such as cinematography \cite{aerial-cinematography-JFR}. Furthermore, the advanced grasping ability has been focused, which enables the fully autonomous delivery without the human interference.
In the last MBZIRC 2017 challenge, many teams succeeded to use their multiorotor aerial robots equipped with electromagnetic gripper to pick up and deliver the treasure autonomously \cite{mbzirc2017-task3-uppen-JFR, mbzirc2017-task3-eth-JFR, mbzirc2017-task3-nimbro-JFR, mbzirc2017-task3-kaist-JFR}. However, these task-specific grippers can only  pick up the magnetic object, indicating the lack of versatility in grasping.

\begin{figure}[h]
  \begin{center}
    \includegraphics[width=\columnwidth]{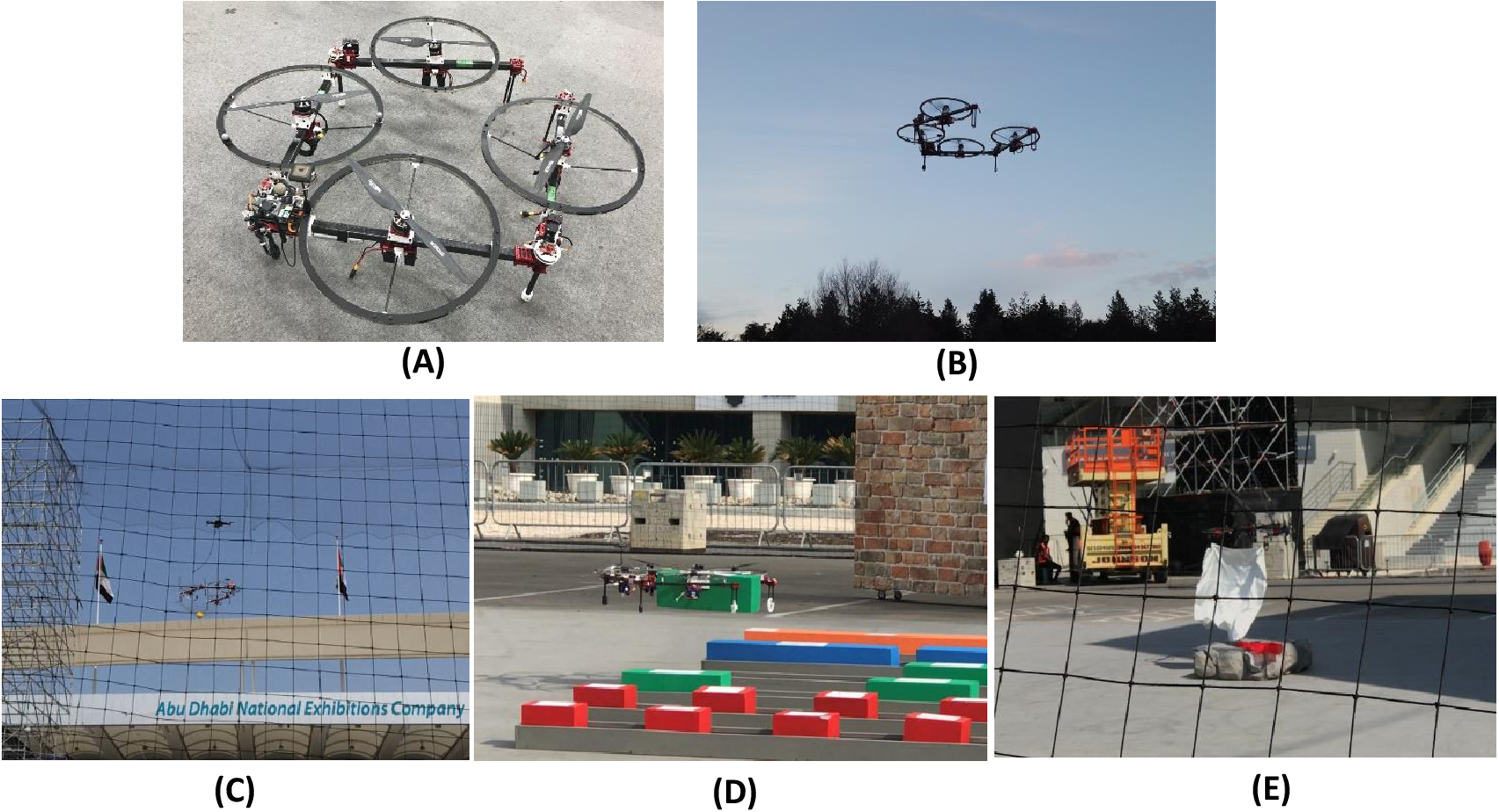}
    \caption{{\bf (A)}: the versatile multilinked aerial robot platform with tilting propellers.  {\bf (B)}: fully autonomous flight in outdoor environment. {\bf (C) $\sim$ (E)}: participation in MBZIRC 2020 which involved the interception with fast moving target, grasping and delivering object,  and firefighting by manipulating blanket.}
    \label{figure:abst}
  \end{center}
\end{figure}

In order to achieve a versatile aerial robot platform without an additional gripper or manipulator, the multilinked structure which enables grasping and manipulation by self-deformation has been proposed in our previous works \cite{hydrus-icra2017, hydrus-ijrr2018, hydrus-regrasping-shi-icra2020}.
Furthermore, the adoption of tilting propellers for the  enhancement on  the controlability   in a model composed from eight links is proposed \cite{anzai-halo-icra2018}. However, such an over-actuated model has a relatively large inertia indicating the difficulty to perform aggressive maneuvering.
In addition, the large number of mechanical components would also increase the maintenance cost.
Therefore, the integration of tilting propellers with a four-link model which is the minimum configuration for the multilinked platform is investigated in this work for the achievement of various autonomous task as shown in \figref{figure:abst}.
The main issues addressed in this work are the design of the multilinked structure with the tilting propellers, the modeling and control for such special under-actuated system, and the state estimation for the fully autonomous flight in the fields.

\subsection{Related Works}

\subsubsection{Deformable Aerial Robot}
Deformability is one of the cutting-edge studies in the field of aerial robot. Various deformable structures are developed based on the quad-rotor model. One of the advantages of deformability is an advanced maneuvering to pass through a narrow space by morphing \cite{soft-robotics-deformable, zurich-deformable, iros2017-deformable}, leading to a potential for exploration in a confined environment such as disaster site. On the other hand, several modular structures are also proposed \cite{upenn-deformable-gripper-icra2018, odar-lasdra-manipulation-icra2018}, which afford an advanced ability of aerial manipulation without using additional manipulator or gripper. However, in these works, there are more than four propellers in each module indicating the redundancy in structural design.
Thus, a minimized modular structure which only contains a single propeller is presented in our previous works \cite{hydrus-ar2016}, and a deformable platform with four links is developed to achieve the grasping ability \cite{hydrus-ijrr2018}. In this platform, there are three joints connecting four links. However, it is possible to further reduce the joint number to two which can still achieve the grasping ability. Such a simplified design can decrease the structural complexity and thus improve the robustness against crash.

\subsubsection{Design and Control for Model with Tilting Propellers}
Regarding the general (undeformable) multirotor model, the rotational axes of rotors are all parallel, leading to the under-actuation in terms of control since the collective thrust force is always vertical in the body frame, and thus the robot cannot track full pose in SE(3). Moreover, the torque in z axis of the body frame is solely dependent on the drag moment generated by propeller rotation, which is significantly weak compared with torques in other axes. This is the reason of the low controlability in yaw rotational motion. Thus, the tilting propeller design is introduced to overcome these difficulties, and several undeformable models composed from more than six tilting propellers are proposed to achieve full pose tracking \cite{se3-hexrotor-control, eth-odar-icra2016, odar-tasme2018}. On the hand, the tilting propeller design is also adopted in the multilinked model by \cite{anzai-halo-icra2018}, which is composed from eight links. However, the model is over-actuated resulting in a large inertia which is difficult to perform  fast motion.

Then the integration of tilting propellers with a four-link model is studied. Prior to the multilinked structure, the tilting propeller design for a general quadrotor is presented by \cite{quadrotor-tilt-inwards-JIRS2015}, where rotors are all tilted inwards regarding the center of body to enhance the stability of the horizontal motion. However, the insufficiency of the torque in z axis still remains in such tilting design, since torque in z axis still depends solely on the drug moment. Then, alternative tilting design same with \cite{anzai-halo-icra2018} is applied in the four-link model by \cite{shi-tilt-hydrus-mpc-ar2019}, where the propeller is tilted along the link direction. With this tilting design,  an additional moment much larger than the drag moment can be generated resulted from the multiplication by the tilted thrust force and the distance to the center of gravity.
On the other hand, the decision of the tilting angle is very important. The relationship between the tilting angle and the force efficiency is revealed by \cite{ryll-sync-tilting-hexarotor-iros2016}. However, there are more factors which should be considered, such as the quantity of the torque in z axis and the horizontal force generated by the tilted thrust force.
In this work, a comprehensive investigation of the influence of the tilting angle on the statics, dynamics and the aerodynamics interference is presented.

In terms of the control for model with tilting propellers, several control methods for fully-actuated model have been proposed \cite{se3-hexrotor-control, odar-tasme2018, anzai-halo-icra2018},  which are not available for the under-actuated model with  four tilting propellers . Then, a nonlinear model predictive control method is proposed by \cite{shi-tilt-hydrus-mpc-ar2019} to address the under-actuation, and the stability of the translational and yaw rotational motion is achieved. However, this control method highly depends on the accurate dynamics model, thus the model error is not allowed. Such condition would lead to a difficulty in stabilizing during the grasping task, since the model offset resulted from the additional inertial parameter of the grasped object can not be perfectly compensated. Thus, a more robust control method is developed in this work which is based on the cascaded control flow similar to \cite{hydrus-ar2016}.

\subsubsection{State Estimation}
The autonomous flight is achieved by the state estimation fusing mutiple sensors. In addition to GPS which is a useful sensor to get a global position in outdoor environment, visual odometry (VO) and visual-inertial-odometry (VIO)  are also the effective methods to obtain the robot motion  \cite{vins-mono-tro, svo-tro, rovio-ijrr, orb-slam-tro, robust-stereo-vio-kumar-tro}.
Several VIO based state estimation methods used in MBZIRC 2017 are introduced by \cite{mbzirc2017-imperial-college-JFR, mbzirc2017-task3-eth-JFR}.

Sensors have measurement delays. Ignoring the measurement delay would significantly decrease the estimation performance. Therefore, a time-synchronized algorithm in sensor fusion is necessary. A time-synchronized extended Kalman filter framework which can handle multiple sensors with different measurement delays is first proposed by \cite{time-sync-ekf-eth}.
In our work, a similar time-synchronized framework with a limited buffer is applied.

On the other hand, the consideration of the  kinematics is significantly important in a multilinked model,  since the relative pose of each sensor changes according to the joint angles. Thus, the proper transformation for both input and output of extended Kalman filter is necessary to guarantee the accurate estimation and stable control during deformation.

\subsection{Main Contribution}
To the best of our knowledge, this is the first study for the multilinked aerial robot  to achieve the fully autonomous flight and the manipulation task in outdoor environment. In short, the main contribution of this study which can benefit the field robotics community is summarized as follows:
\begin{itemize}
\item we design a multilinked aerial robot containing two joints and four tilting propellers to enhance the controlability in yaw rotational motion, and also introduce an optimal design method for the tilt angle.
\item we derive the modeling of the under-actuated model with tilting propellers and also develop the control method. The stability analysis  based on the Lyapunov theory is also presented.
\item we develop the state estimation framework for the multilinked model which takes the kinematics and sensor measurement delay into account to enable autonomous flight in outdoor environment.
\item we perform various experiments including fast maneuvering and aerial manipulation to demonstrate the feasibility of our design, modeling, control and state estimation method for fully autonomous flight.
\end{itemize}

\subsection{Notation}
All the symbols in this paper are explained at their first appearance. Boldface symbols (most are lowercase, e.g., $\bm{r}$) denote vectors, whereas non-boldface symbols (e.g., $m$ or $I$) denote either scalars or matrices.
A coordinate regarding a vector or a matrix is denoted by a left superscript, e.g., ${}^{\scalebox{0.4}{$\{C\}$}}\bm{r}$ expresses $\bm{r}$ with reference to (w.r.t.) the frame $\{C\}$. Subscript are used to express a relation or attribute, e.g., ${}^{\scalebox{0.4}{$\{C\}$}}r_{x}$ represents the value of position on the $x$ axis w.r.t. the frame $\{C\}$.


\subsection{Organization}
The remainder of this paper is organized as follows: the design and modeling for the multilinked structure along with the optimal design method for the propeller tilt angle are presented in \secref{design}. Then the control method and the state estimation are presented in \secref{control} and  \secref{estimation} respectively, followed by the description on the hardware and software system for a real robot platform in \secref{platform}. Last, we show the experiment results in \secref{experiments} before concluding in \secref{conclusion}.

\section{Design and Modeling}
\label{sec:design}
Although the multilinked aerial robot with tilting propellers has been already developed in our previous work \cite{anzai-halo-icra2018, shi-tilt-hydrus-mpc-iros2019, shi-tilt-hydrus-mpc-ar2019}, none of theses works state the concrete design methodology regarding the tilting propeller. Thus, in this section, we will reveal the influence of the tilt angle on the flight performance to further derive an optimal design method for a multilinked aerial robot.

\subsection{Mechanical Design}
\begin{figure}[h]
  \begin{center}
    \includegraphics[width=\columnwidth]{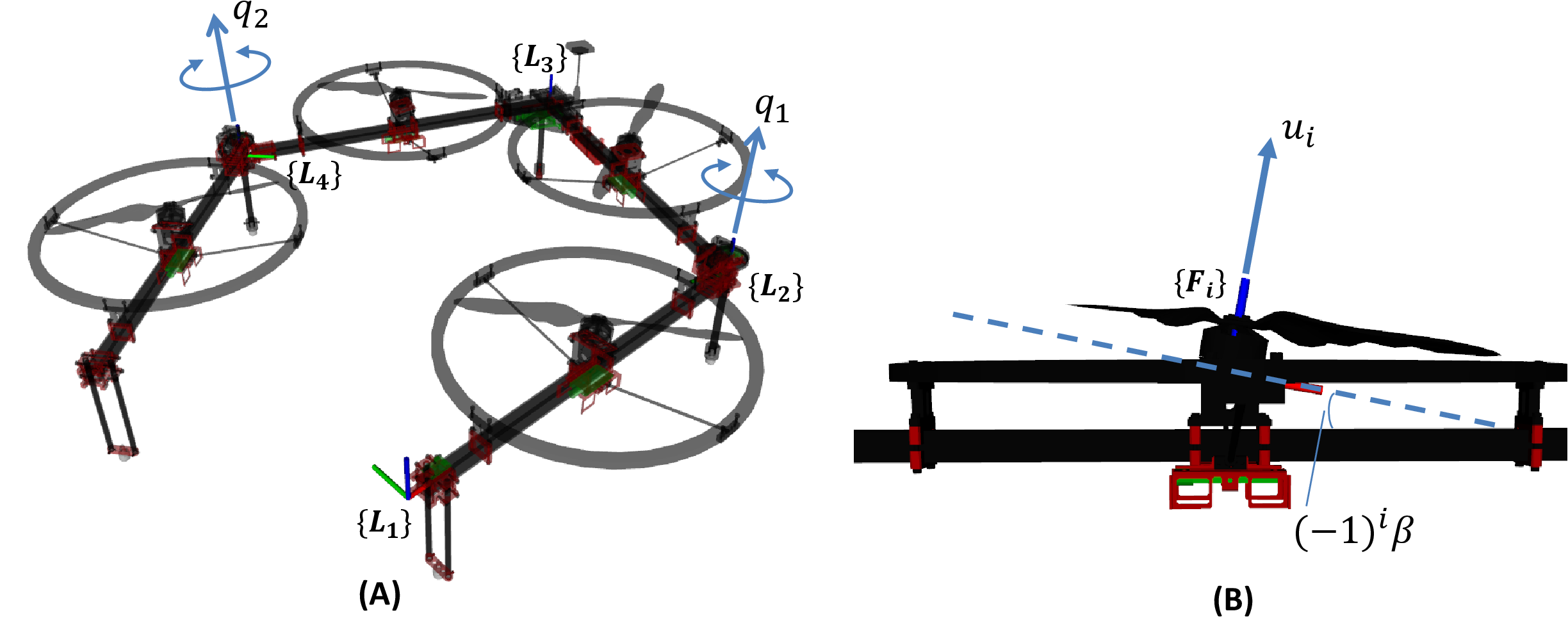}
    \caption{{\bf A}: the mechanical structure of the multilinked aerial robot composed from four links (the central part is separated into two links: link2 and link3). $\{L_i\}$ corresponds to the origin of each link. {\bf B}: the structure of the link module with the embedded propeller which has a fixed tilting angle $(-1)^i\beta$. $u_i$ denotes the thrust force acting on the rotor frame $\{F_i\}$.}
    \label{figure:design}
  \end{center}
\end{figure}

The proposed multilinked aerial robot as shown in \figref{figure:design}(A)  is composed from three parts which are connected by two joints, namely $\bm{q} = [q_1 \ q_2]^{\mathrm{T}}$. Joints are actuated by servos and the rotating axes are parallel, resulting in an ability of two-dimensional deformation. A critical difference from the four-link model in our previous work is that the joint between link2 and link3 is replaced by a fixed connection which can greatly enhance the entire rigidity. Although the freedom-of-dimension (dof) of deformation decreases, such simplified design can still achieve various manipulation task such as grasping object by regarding the whole body like a gripper. For convenience in modeling, we still regard the central part as two separated links, namely link2 and link3.

Regarding each link module as shown in \figref{figure:design}(B), there is a fixed angle $(-1)^i\beta$ to tilt the propeller for the enhancement of the controllabiltiy in yaw rotational motion. The tilting direction is opposite between neighboring links to increase the z axis torque in  both direction. Besides, the propeller spins oppositely regarding neighboring propellers. Furthermore, the propeller duct is designed not only for safety, but also for grasping.

\subsection{Basic Modeling}

\begin{figure}[h]
  \begin{center}
    \includegraphics[width=0.8\columnwidth]{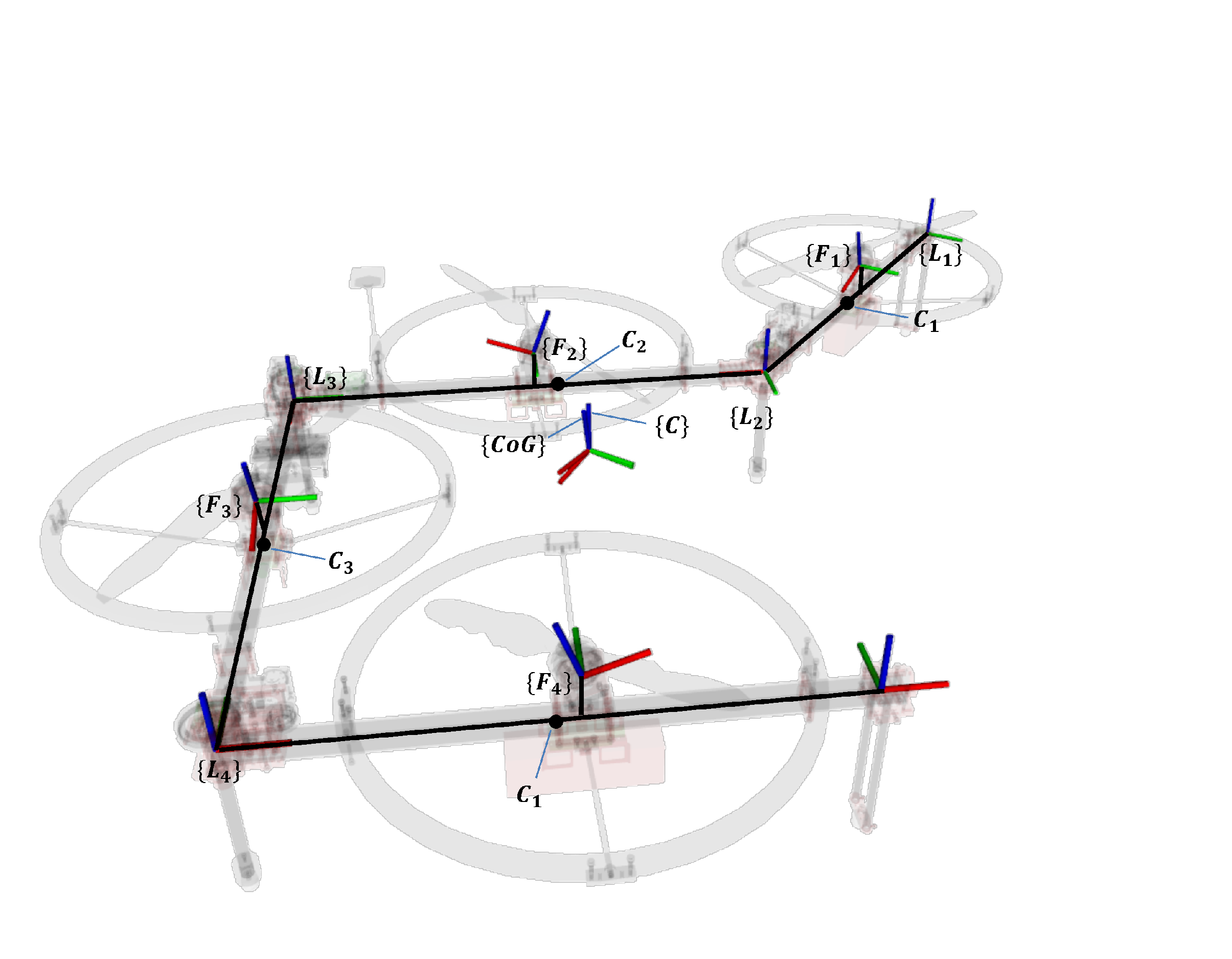}
    \caption{The kinematics model of the multilinked structure. The frame $\{C\}$ and $\{CoG\}$ have the same origin which are all identical to the center of gravity of the whole model. However, the orientation of the frame $\{C\}$ is identical to the frame $\{L_i\}$, while the orientation of the frame $\{CoG\}$ is specially defined in this work for flight control which is presented in \secref{control}.}
     \label{figure:kinematics}
  \end{center}
\end{figure}

We first define an intermediate frame, namely $\{C\}$ as shown \figref{figure:kinematics}, which has following attributes:
\begin{eqnarray}
\label{eq:c_frame_define_pos}
      {}^{\{L_1\}}\bm{p}_{\{C\}} &=& \frac{1}{{m_{\Sigma}}}\displaystyle \sum_i^4 m_{L_i} {}^{\{L_1\}}\bm{p}_{C_i}(\bm{q}) \\
\label{eq:c_frame_define_rot}
      {}^{\{L_1\}}R_{\{C\}} &=& E_{3 \times 3}
\end{eqnarray}
where, $\{L_1\}$ is the frame of link1 which is the root of whole model, and ${}^{\{L_1\}}\bm{p}_{C_i}$ is the position of center of gravity (CoG) of $i$-th link which is variant because of the joint angles $\bm{q} \in \mathcal{R}^2$. ${m_{\Sigma}}$ is the overall mass, i.e., ${m_{\Sigma}} = \sum m_{L_i}$.
\equref{eq:c_frame_define_pos} and \equref{eq:c_frame_define_rot}  indicate that the origin of $\{C\}$ is identical to the CoG of the whole model, and the direction of frame axes of $\{C\}$ is identical to that of $\{L_1\}$. Note that, the frame $\{C\}$ is not suitable to serve as the reference frame in flight control, and thus a further derived frame  $\{CoG\}$ will be introduced in \secref{control}.

Then, the wrench generated by each rotor w.r.t the frame  $\{C\}$ can be written as follows:
\begin{eqnarray}
\label{eq:wrench_force}
      {}^{\{C\}}\bm{f}_{i} &=& {}^{\{C\}}R_{\{F_i\}}(\bm{q}) u_i \\
\label{eq:wrench_torque}
      {}^{\{C\}}\bm{\tau}_{i} &=&  {}^{\{C\}}\bm{p}_{\{F_i\}}(\bm{q}) \times {}^{\{C\}}\bm{f}_{i} = {}^{\{C\}}\hat{\bm{p}}_{\{F_i\}}(\bm{q}) {}^{\{C\}}R_{\{F_i\}}(\bm{q}) u_i
\end{eqnarray}
where,  $u_i \in \mathcal{R}$ is the thrust force generated by $i$-th rotor in the rotor frame $\{F_i\}$. Note that, $\hat{\cdot}$ denotes the operation from a vector to a skew-symmetric matrix.

Using \equref{eq:wrench_force} and \equref{eq:wrench_torque}, the total wrench can be summarized as:
\begin{eqnarray}
\label{eq:total_wrench_c}
&\left(
\begin{array}{c}
  {}^{\{C\}}\bm{f} \\
  {}^{\{C\}}\bm{\tau} \\
\end{array}
\right)
=
{\displaystyle \sum_{i = 1}^{N}} \left(
\begin{array}{c}
  {}^{\{C\}}\bm{f}_i \\
  {}^{\{C\}}\bm{\tau}_i \\
\end{array}
\right)
= \left(
\begin{array}{c}
 Q^{'}_{\mathrm{tran}}(\bm{q}) \\
 Q^{'}_{\mathrm{rot}}(\bm{q}) \\
\end{array}
\right)  \bm{u}
=
Q^{'}(\bm{q}) \bm{u}
\end{eqnarray}
where $\bm{u} = \left[\begin{array}{cccc} u_1 & u_2 & u_3 & u_4 \end{array} \right]^{\mathrm{T}} $.

A hovering state is the zero equilibrium that total wrench expressed by \equref{eq:total_wrench_c} balances with gravity. However the allocation matrix $Q^{'}(\bm{q}) \in \mathcal{R}^{6 \times 4}$ reveals the difficulty to use  four dof input to manipulate  six dof output independently, which is also the reason of the under-actuation. Therefore, in most of the case, it is unable to find $\bm{u}$ that satisfies following condition: $\left[\begin{array}{cccccc} 0 &0 & m_{\Sigma}g & 0 & 0 &0 \end{array} \right]^{\mathrm{T}} = Q^{'}(\bm{q}) \bm{u}$.
However, it is possible to find a solution $\bm{u}_s$ for following relaxed equations:
\begin{eqnarray}
  \label{eq:u_s}
  &  \| {}^{\{C\}}\bm{f} \| = \| Q^{'}_{\mathrm{tran}} \bm{u}_s \| = m_{\Sigma}g \\
&  \bm{0} =  Q^{'}_{\mathrm{rot}} \bm{u}_s
\end{eqnarray}
It is notable that $\bm{u}_s$ is the hovering thrust vector, since $Q^{'}_{\mathrm{tran}} \bm{u}_s$ can be converted to $\left[\begin{array}{cccccc} 0 &0 & m_{\Sigma}g \end{array} \right]^{\mathrm{T}}$ by a certain rotational transformation.
To obtain $\bm{u}_s$, we first find a thrust vector $\bm{u}^{'}$ which only balances with unit z axis and torque:
\begin{equation}
  \label{eq:u_dash}
  \bm{u}^{'} = \left(
  \begin{array}{c}
    Q^{'}_{\mathrm{tran_z}}(\bm{q}) \\
    Q^{'}_{\mathrm{rot}}(\bm{q}) \\
  \end{array}
  \right)^{-1} \left[\begin{array}{cccc} 1 & 0 & 0 &0 \end{array} \right]^{\mathrm{T}}
\end{equation}
where, $Q^{'}_{\mathrm{tran_z}}(\bm{q}) \in \mathcal{R}^{1 \times 4}$ is the third row vector of $Q^{'}_{\mathrm{tran}}(\bm{q})$.

Using \equref{eq:u_dash}, $\bm{u}_s$ can be given by
\begin{equation}
  \label{eq:u_s2}
  \bm{u}_s = \frac{m_{\Sigma}g}{\|Q^{'}_{\mathrm{tran}}(\bm{q}) \bm{u}^{'} \|} \bm{u}^{'}
\end{equation}

\subsection{Design of Tilting Angle for Propeller}


Regarding the tilting propeller  as shown in \figref{figure:design}(B), the thrust force can be divided into the vertical component $cos(\beta)u_i$ and the horizontal component $sin(\beta)u_i$. The relationship between the vertical component and the energy efficiency is revealed by \cite{ryll-sync-tilting-hexarotor-iros2016}, whereas the influence of the horizontal component on the yaw rotational motion is qualitatively discussed in \cite{shi-tilt-hydrus-mpc-ar2019}.
However, there are other critical factors, such as the influence on the horizontal motion, should be also taken into account to design the tilting angle.
In this work, we present four factors (i.e., hovering thrust, torque in z axis, horizontal force and aerodynamic interference) associated with the tilting propeller, which are further integrated into an optimal problem to obtain the best tilting angle.
Note that, we apply the model specification of the actual platform as shown in \figref{figure:hardware}  to quantify the model parameters (e.g., the link length, mass). 

\subsubsection{Hovering Thrust}

\begin{figure}[h]
  \begin{center}
    \includegraphics[width=0.9\columnwidth]{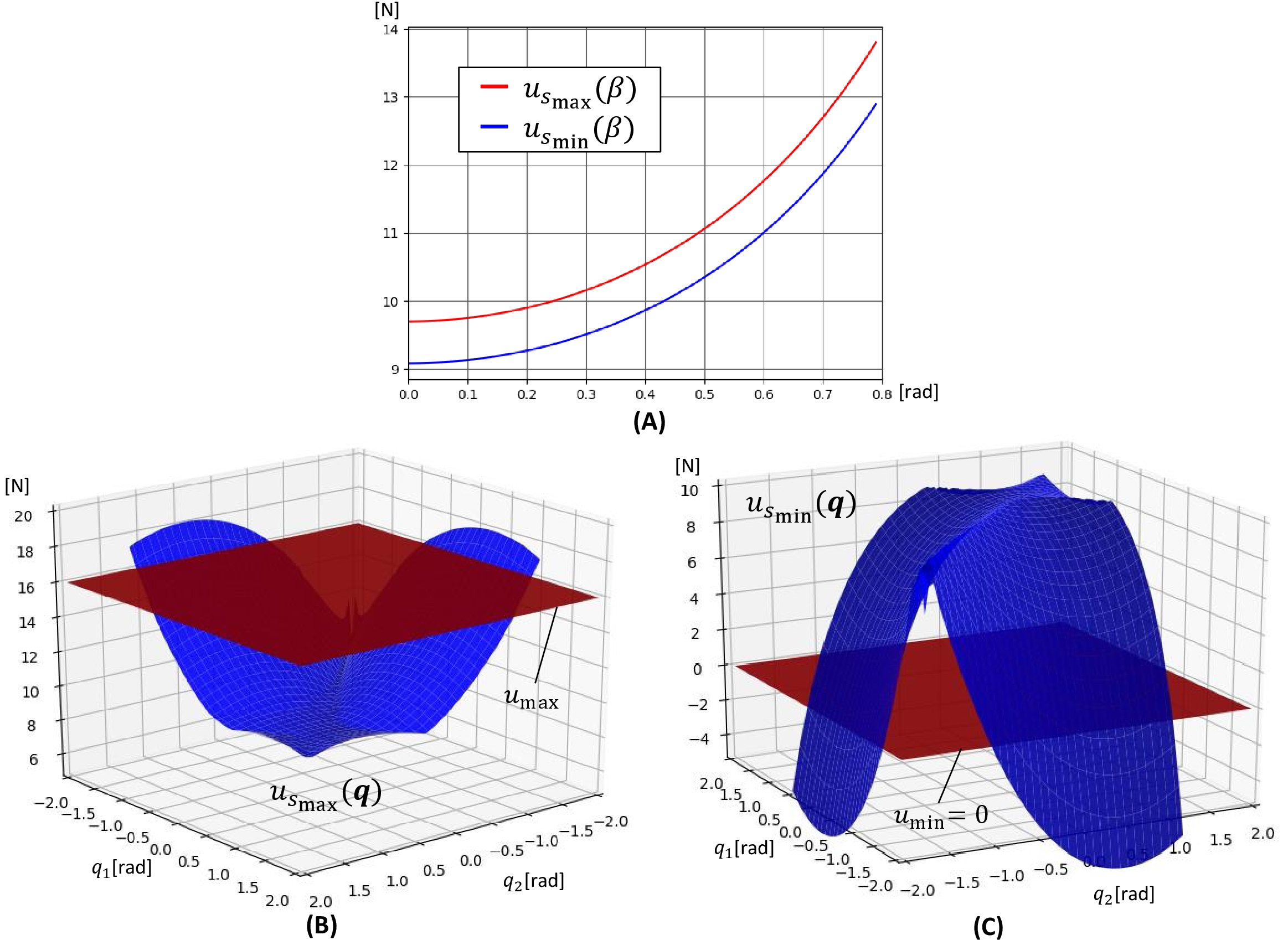}
    \caption{{\bf (A)}: the behaviors regarding the maximum and the minimum components of the hovering thrust vector $\bm{u}_s$ along the change in the tilt angle $\beta$ with fixed joint angles (i.e., $q_i = \frac{\pi}{2}$ rad).  {\bf (B)} the behavior of the maximum component of $\bm{u}_s$ along the change in the joint angles (i.e., $q_i \in \left[ -\frac{\pi}{2} \ \frac{\pi}{2} \right]$) with fixed tilt angle (i.e., $\beta = 10$ deg). The red plane denotes the upper limit of the thrust that the rotor can generate (i.e., 16 N). {\bf (C)} the behavior of the minimum component of $\bm{u}_s$ along  the change in the joint angles with fixed tilt angle (i.e., $\beta = 10$ deg). The red plane denotes the lower limit of thrust (i.e., 0 N).}
    \label{figure:thrust_maxmin_plot}
  \end{center}
\end{figure}

The hovering thrust $\bm{u}_s$ calculated from \equref{eq:u_s2} is the thrust vector in hovering situation, which corresponds to the energy efficiency in most of the situation. We explicitly add the joint angles $\bm{q}$ and the propeller tilt angle $\beta$ as the variables of $\bm{u}_s$, and define the maximum and minimum elements as $u_{s_{\mathrm{max}}}(\bm{q}, \beta)$ and $u_{s_{\mathrm{min}}}(\bm{q}, \beta)$, respectively. Then, we analysis the influence of tilt angle $\beta$ on $u_{s_{\mathrm{min}}}$ and $u_{s_{\mathrm{max}}}$ by changing the angle from 0 rad to 0.8 rad and fixing the joint angles (i.e. $\bm{q} = [\frac{\pi}{2} \ \frac{\pi}{2}]^{\mathrm{T}}$ rad) as shown in \figref{figure:thrust_maxmin_plot}(A). Both plots demonstrate the monotonous increase that are identical to the behaviors  of $\frac{u_{s_{\mathrm{min}}}(\bm{q}, 0)}{cos(\beta)}$ and $\frac{u_{s_{\mathrm{max}}}(\bm{q}, 0)}{cos(\beta)}$. On the other hand, we also fix the tilt angle (i.e., $\beta = 10$ deg), and change the joint angles (i.e., $q_i \in \left[ -\frac{\pi}{2} \ \frac{\pi}{2} \right]$) as shown in \figref{figure:thrust_maxmin_plot}(B) and (C).
From the plot results, we can confirm that the gap between the  $u_{s_{\mathrm{max}}(\bm{q})}$ and $u_{s_{\mathrm{min}}}(\bm{q})$  becomes larger when the joint angle $q_i$ is close to $-\frac{\pi}{2}$ rad, implying the force efficiency and the stability become worse under such form.
Furthermore, the hovering thrust should be also within the valid range of the thrust force, i.e., $u_{s_{\mathrm{min}}}(\bm{q}) \geq 0, u_{s_{\mathrm{max}}}(\bm{q}) \leq u_{\mathrm{max}}$, which is an important factor to clarify the valid deformation range.

\subsubsection{Torque in Z Axis}
The weakness of torque in z axis compared with other axes (i.e., $\tau_z \sim 0.1 \tau_{xy}$) is one of the critical issues regarding the general multirotor, since torque in z axis can only be generated from the rotor's drug moment which is significantly small . This leads to the insufficient robustness against the yaw rotational disturbance \cite{anzai-hydrus-iros2017}, and also induces the thrust force saturation which further influences the stability in other axes.

The  minimum and maximum of the z axis torque can be obtained by solving following problems:
\begin{eqnarray}
  \label{eq:tau_yaw_min}
& \tau_{z_{\mathrm{min}}}(\bm{q}, \beta) = \displaystyle \min_{\bm{u}} Q^{'}_{\mathrm{rot}_z}(\bm{q}, \beta) \bm{u} \\
&  s.t. \hspace{3mm} 0 \leq u_i \leq u_{\mathrm{max}} \nonumber \\
  \label{eq:tau_yaw_max}
& \tau_{z_{\mathrm{max}}}(\bm{q}, \beta) = \displaystyle \max_{\bm{u}} Q^{'}_{\mathrm{rot}_z}(\bm{q}, \beta) \bm{u} \\
&  s.t. \hspace{3mm} 0 \leq u_i \leq u_{\mathrm{max}} \nonumber
\end{eqnarray}
where, $Q^{'}_{\mathrm{rot}_z}(\bm{q}, \beta)$ is the third row vector of the matrix $Q^{'}_{\mathrm{rot}}(\bm{q}, \beta)$. 

\begin{figure}[h]
  \begin{center}
    \includegraphics[width=0.9\columnwidth]{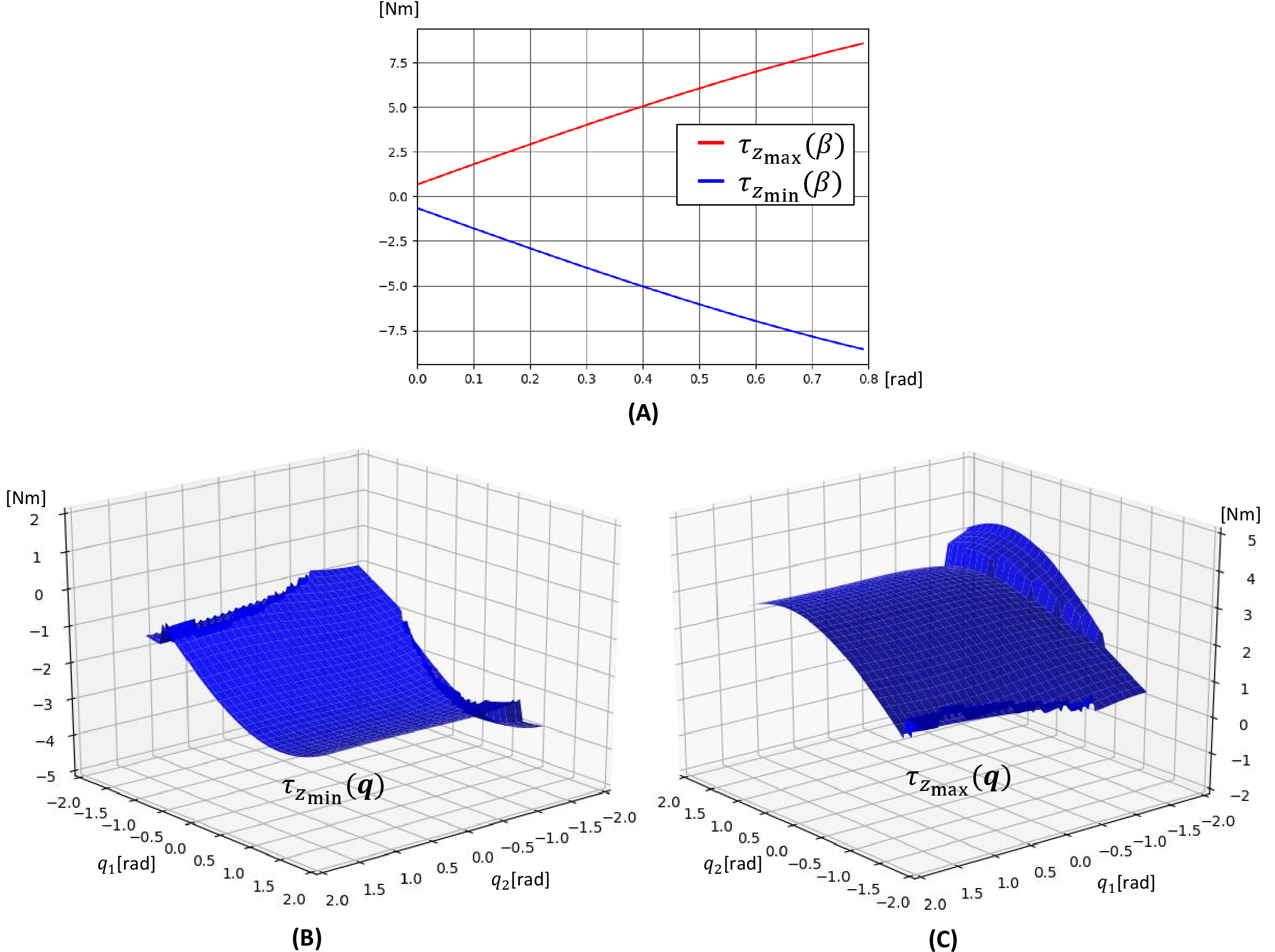}
    \caption{{\bf (A)}: the behaviors of the maximum and the minimum of the z axis torque along the change in the tilt angle with fixed joint angles (i.e., $q_i = \frac{\pi}{2}$ rad).  {\bf (B)} the behavior of the minimum of z axis torque along the change in the joint angles (i.e., $q_i \in \left[ -\frac{\pi}{2} \ \frac{\pi}{2} \right]$) with fixed tilt angle (i.e., $\beta = 10$ deg). {\bf (C)} the behavior of the maximum of z axis torque along the change in the joint angles (i.e., $q_i \in \left[ -\frac{\pi}{2} \ \frac{\pi}{2} \right]$)  with fixed tilt angle (i.e., $\beta = 10$ deg).}
    \label{figure:t_yaw_minmax_plot}
  \end{center}
\end{figure}

Again, we first analysis the influence of tilt angle $\beta$ on these values $\tau_{z_{\mathrm{min}}}$ and $\tau_{z_{\mathrm{max}}}$ by changing the angle from 0 rad to 0.8 rad and fixing the joint angles (i.e. $\bm{q} = [\frac{\pi}{2} \ \frac{\pi}{2}]^{\mathrm{T}}$ rad)  as shown in \figref{figure:t_yaw_minmax_plot}(A). Both absolute values  monotonically increases because z axis torque generated by tilting propeller are associated with the sine of the tile angle. We then fix the tilt angle (i.e., $\beta = 10$ deg)  and change the joint angles (i.e., $q_i \in \left[ -\frac{\pi}{2} \ \frac{\pi}{2} \right]$) as shown in \figref{figure:t_yaw_minmax_plot}(B) and (C). When $q_i \approx -\frac{\pi}{2}$,  the absolute value of either  $\tau_{z_{\mathrm{max}}}(\bm{q})$ or $\tau_{z_{\mathrm{min}}}(\bm{q})$ is  relatively small, leading to  the relatively weak controlability regarding the yaw rotational motion under such robot form.

\subsubsection{Horizontal Force}
The horizontal force $\bm{f}_{xy} \in \mathcal{R}^2$ corresponds to the first two elements of ${}^{\{C\}}\bm{f}$  which can be calculated from \equref{eq:total_wrench_c}.
For fully-/over-actuated model, the horizontal force is beneficial to the control in horizontally translational motion. However, for an under-actuated model, the horizontal motion is controlled by changing the robot attitude, and thus the intervention of this force would break the position stability.
To clarify the 
Given a desired torque ${}^{\{C\}}\bm{\tau}^{\mathrm{des}}$  obtained from a attitude controller, a derived horizontal force $\bm{f}_{xy}$ can be calculated by following equation:
\begin{equation}
  \label{eq:horizontal_force}
\bm{f}_{xy}(\bm{q}, \beta, {}^{\{C\}}\bm{\tau}^{\mathrm{des}}) =   Q^{'}_{\mathrm{tran_{xy}}}(\bm{q}, \beta)Q^{'\#}_{\mathrm{rot}}(\bm{q}, \beta) {}^{\{C\}}\bm{\tau}^{\mathrm{des}}
\end{equation}
where the matrix operation $(\cdot)^{\#}$ denotes the MP inverse  which corresponds to the minimum norm of $\bm{u}$ to generate ${}^{\{C\}}\bm{\tau}^{\mathrm{des}}$. Also note that, $Q^{'}_{\mathrm{tran_{xy}}}(\bm{q}) \in \mathcal{R}^{2 \times 4}$ corresponds to the top two rows of $Q^{'}_{\mathrm{tran}}$.

\begin{figure}[h]
  \begin{center}
    \includegraphics[width=0.9\columnwidth]{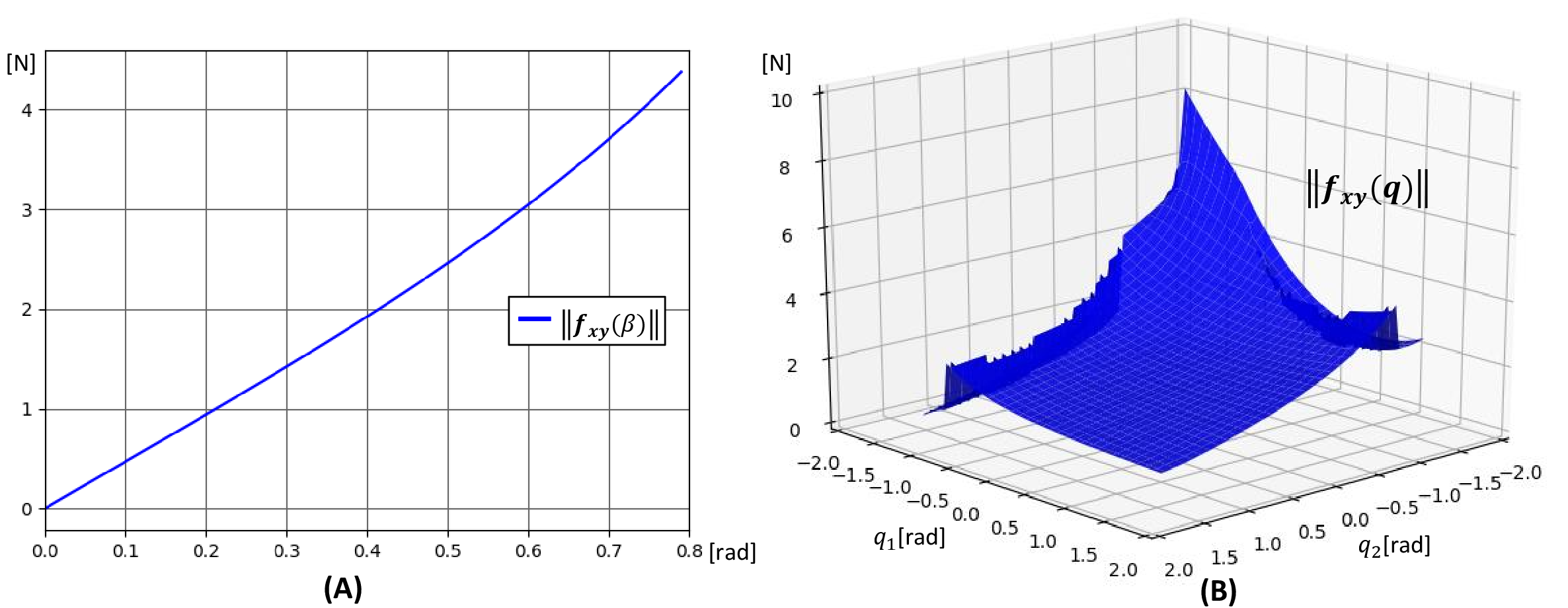}
    \caption{{\bf (A)}: the behavior of the norm of the horizontal force $\| \bm{f}_{xy}(\beta) \|$ calculated from  \equref{eq:horizontal_force} with  $\bm{\tau}^{\mathrm{des}} = \left[1 \ 1 \ 1\right]^{\mathrm{T}}$ and fixed joint angles (i.e., $q_i =  \frac{\pi}{2}$ rad)  along the change in the tilt angle. {\bf (B)} the behavior of  the norm of the horizontal force $\| \bm{f}_{xy}(\bm{q}) \|$ along the change in the joint angles with fixed tilt angle (i.e., $\beta = 10$ deg).}
    \label{figure:f_xy_norm_plot}
  \end{center}
\end{figure}

To simplify the influence of the rotational motion in the translational motion, a fixed desired torque ${}^{\{C\}}\bm{\tau}^{\mathrm{des}} = \left[1 \ 1 \ 1\right]^{\mathrm{T}}$ is introduced. Then it is possible to validate the relationship between the norm of the horizontal force $\| \bm{f}_{xy}(\bm{q}, \beta) \|$ and the tilt angle and joint angles.
Again, we first analysis the influence of tilt angle $\beta$ on  $\| \bm{f}_{xy}(\bm{q}, \beta) \|$  with fixed joint angles (i.e. $\bm{q} = [\frac{\pi}{2} \ \frac{\pi}{2}]^{\mathrm{T}}$ rad)  as shown in \figref{figure:f_xy_norm_plot}(A), which demonstrates a monotonous increase and also an increase in inclination. We consider that $Q^{'}_{\mathrm{tran_{xy}}}(\bm{q}, \beta)$ and $Q^{'\#}_{\mathrm{rot}}(\bm{q}, \beta)$ has the element of $sin(\beta)$ and $\frac{1}{cos(\beta)}$ respectively, thus $\| \bm{f}_{xy}(\beta) \| \propto tan(\beta)$ . Then, we fix the tilt angle (i.e., $\beta = 10$ deg), and change the joint angles (i.e., $q \in \left[ -\frac{\pi}{2} \ \frac{\pi}{2} \right]$) as shown in \figref{figure:f_xy_norm_plot}(B). $\| \bm{f}_{xy}(\bm{q}) \|$ should be as small as possible. However, the value divergence when both $q_1$ and $q_2$ are close to $ -\frac{\pi}{2}$ rad, since this is a singular form where all propeller are aligned on the same line.

\subsubsection{Aerodynamic Interference}
Is is also necessary to consider the aerodynamic interference from the tilting propeller, since the airflow form propeller hits on the downstream link rod as shown \figref{figure:design}(B). The influence range can be given by
\begin{equation}
  \label{eq:aerointerference}
  l_{\mathrm{air}}(\beta) \approx \frac{D_{\mathrm{prop}}}{cos(\beta)}
\end{equation}
where, $D_{\mathrm{prop}}$ is the diameter of the propeller. 
It is also notable that, when the tilt angle increases, the airflow will act not only on  the link rod but also  on the side component, such as, the duct. Thus, it is necessary to suppress such aerointerference.

\subsubsection{Optimization Problem}
Our goal is to propose a design method which takes above four factors into account to find an optimal tilt angle for multilinked aerial robot.
Although we only validate the influence of the change in the tilt angle $\beta$  under a single form (i.e., $q_i = \frac{\pi}{2}$ rad), it can be considered that the behavior in other forms is identical. Thus, the optimization problem to find the best tilt angle is designed under $q_i = \frac{\pi}{2}$ rad. 
Then, a maximization of the following weighted sum function is introduced:
\begin{eqnarray}
  \label{eq:design_optimization}
  &\displaystyle \max_{\beta} \bm{w}^{\mathrm{T}} \bm{x}(\beta) \\
  &\bm{x} = \left[ -\begin{array}{cccc} u_{s_{\mathrm{max}}}(\beta) + u_{s_{\mathrm{min}}}(\beta) & \tau_{z_{\mathrm{max}}}(\beta) - \tau_{z_{\mathrm{min}}}(\beta) & - \| \bm{f}_{xy}(\beta) \| & -l_{a}(\beta)  \end{array}\right]^{\mathrm{T}} \nonumber
\end{eqnarray}
where, $\bm{w}$ is the weight vector to normalize each component in $\bm{x}(\beta)$. Given a robot specification from the actual platform as shown \figref{figure:hardware}, the optimal tilt angle is  $0.1745$ rad ($ = 10$ deg). This is the result  which puts more weight on the suppression of the horizontal force $\bm{f}_{xy}(\bm{q}, \beta, {}^{\{C\}}\bm{\tau}^{\mathrm{des}})$ to guarantee the position stability. More discussion on the stability and tilt angle will be presented in the \secref{control}. Note that, a larger tilt angle (i.e. 20 deg) is applied in other works \cite{shi-tilt-hydrus-mpc-iros2019, shi-tilt-hydrus-mpc-ar2019}. This is due to the different robot specification (e.g., no battery in the experimental validation). Furthermore, the nonlinear model predict control is applied in those works, which can utilize such horizontal force for position control even in under-actuation model, but with the expense of the high computational cost.

\subsection{Valid Deformation Range}
In addition to the hovering thrust, the controlability regarding the rotational motion is another important factor to check the validity of a form. For torque in z axis, the maximum and minimum value should be positive and negative respectively, i.e.,  $\tau_{z_{\mathrm{max}}}(\bm{q}) > 0$ , $\tau_{z_{\mathrm{min}}}(\bm{q}) < 0$. However, these constraints only corresponds to the validity of yaw rotational motion. It is necessary to check the comprehensive  motion in all three axes. Therefore an extended quantity called feasible control torque convex ${\mathcal V}_{T}(\bm{q})$ \cite{odar-tasme2018, anzai-hydrus-xi-iros2019} is applied to validate the rotational controlability which is given by
\begin{eqnarray}
  \label{eq:feasible_control_torque_convex}
        &{\mathcal V}_T(\bm{q}) := \{ {}^{\{C\}}\bm{\tau}(\bm{q}) \in {\mathcal R}^3 | {}^{\{C\}}\bm{\tau}(\bm{q}) = \displaystyle \sum_{i=0}^{N} u_i \bm{v}_i(\bm{q}), 0 \leq u_i \leq u_{\mathrm{max}} \}
\end{eqnarray}
where, $\bm{v}_i(\bm{q}) = {}^{\{C\}}\hat{\bm{p}}_{\{F_i\}}(\bm{q}) {}^{\{C\}}R_{\{F_i\}}(\bm{q}) $ according to \equref{eq:wrench_torque}.

We further introduce the guaranteed minimum control torque $\tau_{\mathrm{min}}(\bm{q})$ which has following property:
\begin{eqnarray}
  \label{eq:guaranteed_minimum_control_torque}
  \|{}^{\{C\}}\bm{\tau}(\bm{q}) \| \leq \tau_{\mathrm{min}}(\bm{q}) \Rightarrow \bm{\tau}(\bm{q}) \in {\mathcal V}_T(\bm{q}),
\end{eqnarray}

Using the distance $d^{\tau}_{ij}(\bm{q})$ which is from the origin to a plane of convex ${\mathcal V}_T$ along its normal vector $\frac{\bm{v}_i \times \bm{v}_j}{\|\bm{v}_i \times \bm{v}_j\|}$, the guaranteed minimum control torque can be given by:
\begin{eqnarray}
  \label{eq:feasible_control_fxy}
  &d^{\tau}_{ij}(\bm{q}) = \displaystyle \sum_{k=0}^{N} \max (0, \frac{(\bm{v}_i \times \bm{v}_j)^{\mathrm{T}}}{\|\bm{v}_i \times \bm{v}_j\|} \bm{v}_k), \\
  \label{eq:feasible_control_fxy}
  &{\tau}_{\mathrm{min}}(\bm{q}) = \displaystyle \min_{i,j \in {\mathcal I}} d^{\tau}_{ij}(\bm{q}).
\end{eqnarray}
where, $\mathcal{I} := \{1, 2, \cdots, N\}$. 
It is obvious that the condition to guarantee the torque controlability is ${\tau}_{\mathrm{min}}(\bm{q}) > 0$.

\begin{figure}[h]
  \begin{center}
    \includegraphics[width=\columnwidth]{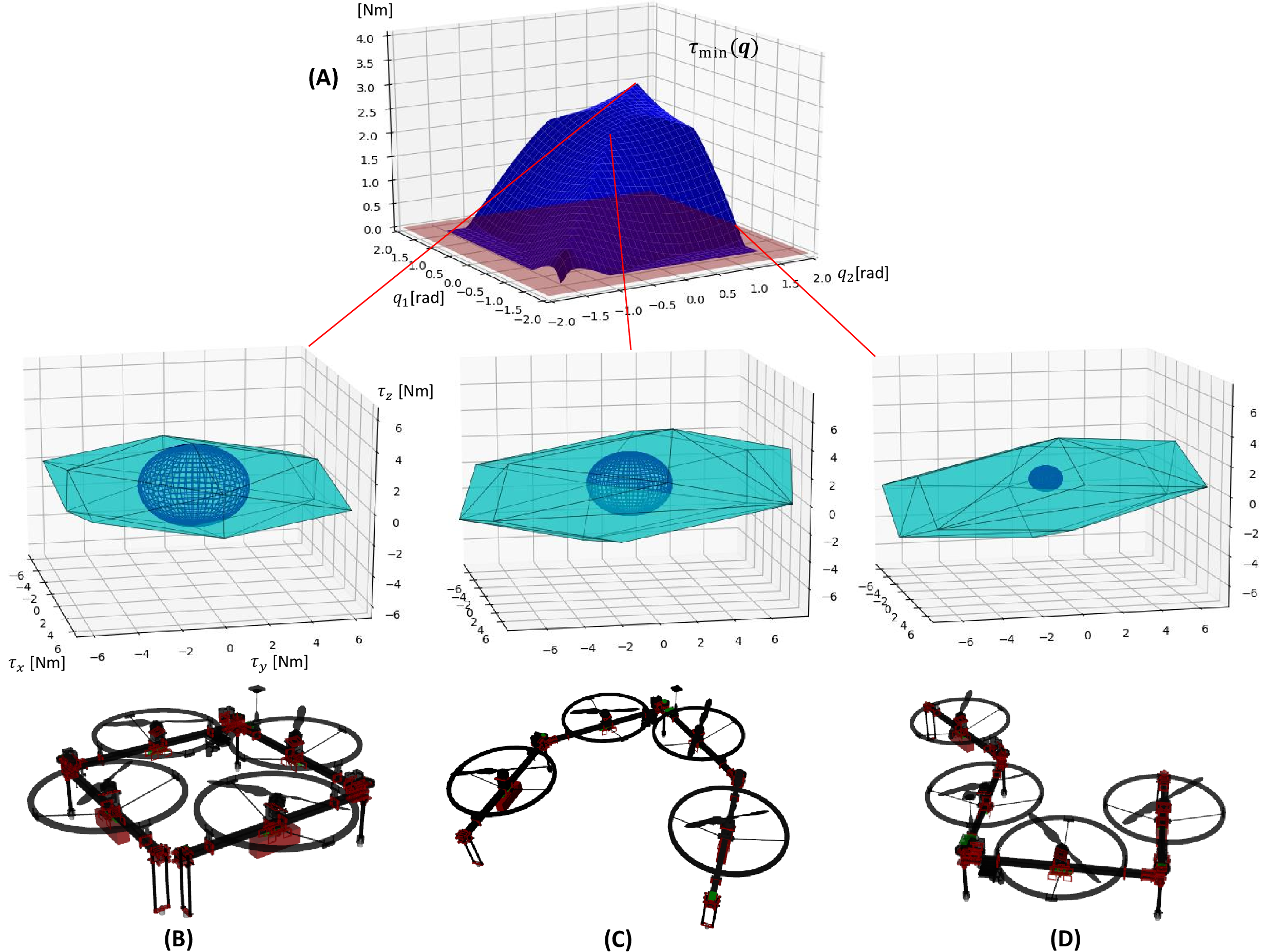}
    \caption{{\bf (A)}: the behavior of the guaranteed minimum control torque ${\tau}_{\mathrm{min}}(\bm{q})$ along the change in joint angles.  {\bf (B)-(D)}: feasible control torque convex  ${\mathcal V}_T$ and the sphere of which the radius is ${\tau}_{\mathrm{min}}(\bm{q})$  under different robot forms  as shown in the bottom images (i.e., {\bf B}: $\bm{q} = \in \left[ \frac{\pi}{2} \ \frac{\pi}{2} \right]$, {\bf C}: $\bm{q} = \in \left[ \frac{\pi}{4} \ \frac{\pi}{4} \right]$, {\bf D}: $\bm{q} = \in \left[ -\frac{\pi}{4} \ \frac{\pi}{2} \right]$).}
    \label{figure:feasible_control_convex_examples}
  \end{center}
\end{figure}

\figref{figure:feasible_control_convex_examples}(A) shows relationship between  ${\tau}_{\mathrm{min}}(\bm{q}))$ and joint angles $\bm{q}$, while \figref{figure:feasible_control_convex_examples}(B)-(D) demonstrate several distinguish cases of  ${\mathcal V}_T$ and ${\tau}_{\mathrm{min}}$ with differential forms. It can be confirmed that the form of \figref{figure:feasible_control_convex_examples}(D)  stretches the convex and thus ${\tau}_{\mathrm{min}}$ becomes significantly small. However, the shrink direction is not perfectly identical to the z axis, which indicates the importance to apply ${\tau}_{\mathrm{min}}$ instead of only evaluating z axis (i.e., $\tau_{z_{\mathrm{max}}}(\bm{q}), \tau_{z_{\mathrm{min}}}(\bm{q})$).


To summarize, the valid range of the joint angles can be given by:
\begin{eqnarray}
  \label{eq:valid_joint_range}
  {\mathcal D}_q := \{ \bm{q} \in {\mathcal R}^2 | u_{s_{\mathrm{min}}}(\bm{q}) \geq u_{\mathrm{thre}}, u_{s_{\mathrm{max}}}(\bm{q}) \leq u_{\mathrm{max}} - u_{\mathrm{thre}}, {\tau}_{\mathrm{min}}(\bm{q}) \geq \tau_{\mathrm{thre}}  \}
\end{eqnarray}
where, $u_{\mathrm{thre}}$ and   $\tau_{\mathrm{thre}}$ are the positive thresholds to provide certain control margin. 

\section{Control}
\label{sec:control}

\begin{figure}[h]
  \begin{center}
    \includegraphics[width=\columnwidth]{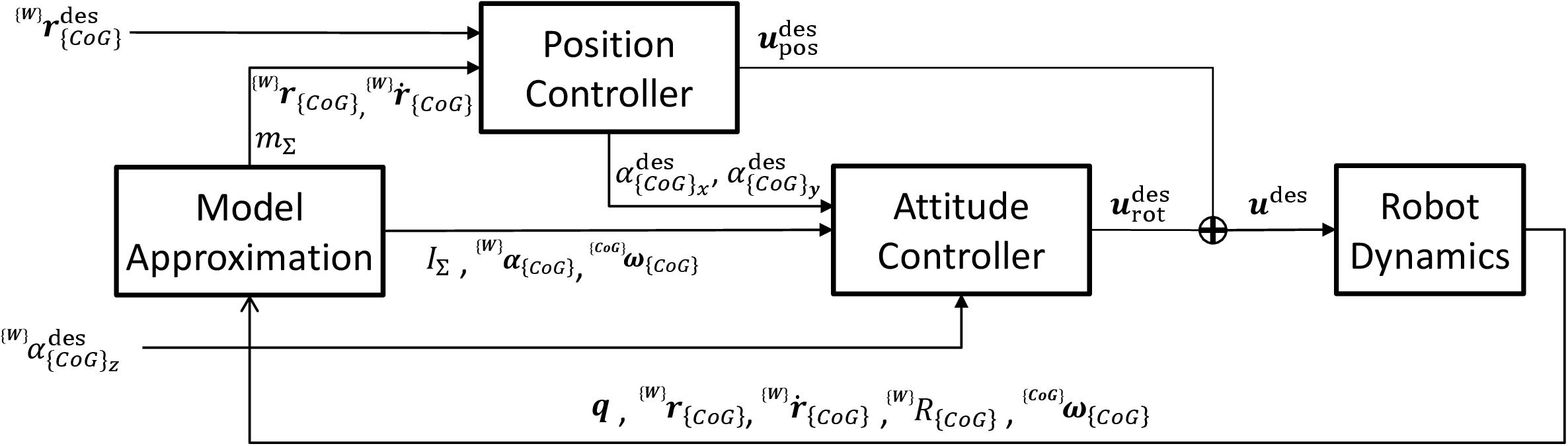}
    \caption{The proposed control framework for multilinked aerial robot  which involves the dynamics model approximation and a cascaded control flow including position and attitude controller.}
    \label{figure:control_diagram}
  \end{center}
\end{figure}

An improved control framework as shown in \figref{figure:control_diagram} based on our previous work \cite{hydrus-ar2016}  has been developed in this work. The dynamics and control is described in a specially defined frame $\{CoG\}$. Subsequently the model approximation is introduced to simply the dynamics, which is followed by a cascaded control flow.

\subsection{Definition of CoG Frame}
As shown in \equref{eq:u_s}, in most cases, the frame  $\{C\}$  could not be level in hovering state. Then, a new frame which is always level in hovering state is introduced to to reduce the nonlinearity of the dynamics and control. We define such a frame as $\{CoG\}$. 

In terms of the relation between the frame of $\{CoG\}$ and $\{C\}$, there must  exist a rotation matrix ${}^{\{CoG\}}R_{\{C\}}$ which satisfies following transformation:
\begin{eqnarray}
  \label{eq:R_CoG}
        {}^{\{CoG\}}R_{\{C\}} Q^{'}_{\mathrm{tran}} \bm{u}_s = {}^{\{CoG\}}R_{\{C\}}  {}^{\{C\}}\bm{f}_s = m_{\Sigma}\bm{g}
\end{eqnarray}
where, $\bm{g} = \left[\begin{array}{ccc} 0 & 0 & g \end{array} \right]^{\mathrm{T}}$, and $\bm{u}_s$ is the hovering thrust vector calculated from \equref{eq:u_s2}.

Then, Euler angles $\bm{\alpha} = (\begin{array}{ccc}\alpha_x  & \alpha_y & \alpha_z \end{array})^{\mathrm{T}}$ are introduced to represent   ${}^{\{CoG\}}R_{\{C\}}$:
\begin{eqnarray}
  \label{eq:R_CoG2}
        {}^{\{CoG\}}R_{\{C\}} &=& R_Y(\alpha_y) R_Y(\alpha_x) \\
        \label{eq:R_CoG_roll}
        \alpha_x &=& tan^{-1}({}^{\{C\}}f_{s_y}, {}^{\{C\}}f_{s_z}) \\
        \label{eq:R_CoG_pitch}
        \alpha_y &=& tan^{-1}(-{}^{\{C\}}f_{s_x}, \sqrt{{}^{\{C\}}f_{s_y}^2 + {}^{\{C\}}f_{s_z}^2})
\end{eqnarray}
where, $R_{X}(\cdot), R_{Y}(\cdot)$ are the special rotation matrix which only rotates along $x$ and $y$ axis, respectively.

Finally, the allocation matrix Q w.r.t the frame $\{CoG\}$ can be given by
\begin{eqnarray}
\label{eq:total_wrench}
Q(\bm{q})
= \left(
\begin{array}{c}
  Q_{\mathrm{tran}}(\bm{q}) \\
  Q_{\mathrm{rot}}(\bm{q}) \\
\end{array}
\right)
= \left(
\begin{array}{c}
  {}^{\{CoG\}}R_{\{C\}}  Q^{'}_{\mathrm{tran}}(\bm{q}) \\
{}^{\{CoG\}}R_{\{C\}}  Q^{'}_{\mathrm{rot}}(\bm{q}) \\
\end{array}
\right)
\end{eqnarray}

\subsection{Dynamics}
As stated in our previous work\cite{hydrus-ijrr2018}, we assume the joint motion is sufficiently slow. Then the multilinked model can be regarded as a time-variant rigid body. Thus, the dynamics regarding the CoG frame can be simplified as follows:
\begin{eqnarray}
  \label{eq:translational_dynamics}
  & m_{\Sigma} ({}^{\{W\}}\ddot{\bm r}_{\{CoG\}} + \bm g) = {}^{\{W\}}R_{\{CoG\}} {}^{\{CoG\}}\bm{f} + \bm{\Delta}_{\mathrm{tran}}  = {}^{\{W\}}R_{\{CoG\}} Q_{\mathrm{tran}}(\bm{q}) \bm{u} + \bm{\Delta}_{\mathrm{tran}} \\
  \label{eq:rotational_dynamics}
  &{}^{\{CoG\}}I_{\Sigma}(\bm{q})  {}^{\{CoG\}}\dot{\bm \omega}_{\{CoG\}} +  {}^{\{CoG\}}{\bm \omega}_{\{CoG\}} \times  {}^{\{CoG\}}I_{\Sigma}(\bm{q})  {}^{\{CoG\}}{\bm \omega}_{\{CoG\}} = {}^{\{CoG\}}{\bm \tau} + \bm{\Delta}_{\mathrm{rot}}  =  Q_{\mathrm{rot}}(\bm{q}) \bm{u} + \bm{\Delta}_{\mathrm{rot}} \nonumber \\
\end{eqnarray}
where ${}^{\{W\}}\bm r_{\{CoG\}}$ and ${}^{\{W\}}R_{\{CoG\}}$ are the position and attitude of the frame $\{CoG\}$  w.r.t the world frame $\{W\}$, respectively.
These states can be calculated based on the forward-kinematics from the states of the root link (i.e., ${}^{\{W\}}\bm r_{\{L_1\}}$, ${}^{\{W\}}\dot{\bm r}_{\{L_1\}}$ and ${}^{\{W\}}R_{\{L_1\}}$), while the angular velocity can be obtained by ${}^{\{CoG\}}\dot{\bm \omega}_{\{CoG\}} =   {}^{\{CoG\}}R_{\{C\}}  {}^{\{L_1\}}{\bm \omega}_{\{L_1\}} $.
The total inertial matrix ${}^{\{CoG\}}I_{\Sigma}(\bm{q})$ w.r.t the frame $\{CoG\}$ can be also calculated from the forward-kinematics process.
We explicitly consider the unstructured but fixed uncertainties $\bm{\Delta}_{\mathrm{tran}}$ and $\bm{\Delta}_{\mathrm{rot}}$ (e.g., the model offset caused by the additional weight in object grasping task) to show the compensation ability of the proposed control framework.

We further assume that most of the flight tasks are performed in near-hover condition. Thus, the approximation between the differential of Euler angles and angular velocity $ {}^{\{W\}}\dot{\bm \alpha}_{\{CoG\}} \approx {}^{\{CoG\}}{\bm \omega}_{\{CoG\}}$ is available, since ${}^{\{W\}}\alpha_{\{CoG\}_x} \approx 0$, ${}^{\{W\}}\alpha_{\{CoG\}_y} \approx 0$. Then the rotational dynamics expressed in \equref{eq:rotational_dynamics} can be further linearized in the near-hover condition:
\begin{equation}
  \label{eq:rotational_dynamics2}
  {}^{\{CoG\}}I_{\Sigma}(\bm{q})  {}^{\{W\}}\ddot{\bm \alpha}_{\{CoG\}} + {}^{\{CoG\}}{\bm \omega}_{\{CoG\}} \times  {}^{\{CoG\}}I_{\Sigma}(\bm{q}) {}^{\{CoG\}}{\bm \omega}_{\{CoG\}} = Q_{\mathrm{rot}}(\bm{q}) \bm{u}  + \bm{\Delta}_{\mathrm{rot}}
\end{equation}
Finally, \equref{eq:translational_dynamics} and \equref{eq:rotational_dynamics2} are the fundamental dynamics which are used in the proposed control framework.
It is also notable that $Q_{\mathrm{tran}}(\bm{q}) \bm{u}$  in \equref{eq:translational_dynamics}  implies the full influence on all translational axes without rotation $ {}^{\{W\}}R_{\{CoG\}}$. Such special property is the main difference from a general quadrotor \cite{quadtor-upenn-ijrr} and is also the crucial issue to solve in flight control.

\subsection{Attitude Control}
We first present the attitude control part in the cascaded control flow as shown in \figref{figure:control_diagram}. In order to compensate the unstructured but fixed uncertainties $\bm{\Delta}_{\mathrm{rot}}$, the integrated control term is required. Although the general PID control can be applied, the optimal control with integral control, called LQI \cite{LQI0}, is more suitable for our model, since it is possible to design the cost function in the optimal control framework.

We first rewrite the rotational dynamics \equref{eq:rotational_dynamics2} as follows:
\begin{eqnarray}
\label{eq:rot_state_equation1}
&\dot{\mbox{\boldmath $x$}} = A \mbox{\boldmath $x$} + B \mbox{\boldmath $u$} + D (\bm{\Delta}_{\mathrm{rot}} - I^{-1}\bm{\omega} \times I \bm{\omega}) \\
\label{eq:rot_state_equation2}
&\mbox{\boldmath $y$} = C\mbox{\boldmath $x$}\\
&\mbox{\boldmath $x$} \in R^{6}, \hspace{3mm} \mbox{\boldmath $u$} \in R^{4}, \hspace{3mm} \mbox{\boldmath $y$} \in R^{3} \nonumber
\end{eqnarray}
where
\begin{eqnarray*}
&
\bm{x} = \left[
  \begin{array}{cccccccc}
    {}^{\{W\}}\alpha_{\{CoG\}_x} & {}^{\{W\}}\dot{\alpha}_{\{CoG\}_x} & {}^{\{W\}}\alpha_{\{CoG\}_y} & {}^{\{W\}}\dot{\alpha}_{\{CoG\}_y} & {}^{\{W\}}\alpha_{\{CoG\}_z} & {}^{\{W\}}\dot{\alpha}_{\{CoG\}_z}
  \end{array}
  \right]^{\mathrm{T}} \\
  & A =  \left[
    \begin{array}{cccccccc}
      0 & 1 & 0 & 0 & 0 & 0  \\
      0 & 0 & 0 & 0 & 0 & 0  \\
      0 & 0 & 0 & 1 & 0 & 0  \\
      0 & 0 & 0 & 0 & 0 & 0  \\
      0 & 0 & 0 & 0 & 0 & 1  \\
      0 & 0 & 0 & 0 & 0 & 0  \\
    \end{array}
    \right]
;
 B =  \left[
    \begin{array}{c}
      {\bf 0} \\ Q_{\mathrm{rot}_x}(\bm{q}) \\ {\bf 0} \\ Q_{\mathrm{rot}_y}(\bm{q}) \\ {\bf 0} \\ Q_{\mathrm{rot}_z}(\bm{q})
    \end{array}
    \right]^{\mathrm{T}}
;
 C =  \left[
    \begin{array}{cccccccc}
      1 & 0 & 0 & 0 & 0 & 0 \\
      0 & 0 & 1 & 0 & 0 & 0 \\
      0 & 0 & 0 & 0 & 1 & 0 \\
    \end{array}
    \right]
 ;
 D =  \left[
   \begin{array}{ccc}
     0 & 0 & 0  \\
     1 & 0 & 0  \\
     0 & 0 & 0  \\
     0 & 1 & 0  \\
     0 & 0 & 0  \\
     0 & 0 & 1  \\
   \end{array}
    \right]
\end{eqnarray*}
Note that, $Q_{\mathrm{rot}}(\bm{q}) = \left[\begin{array}{ccc} Q_{\mathrm{rot}_x}(\bm{q})^{\mathrm{T}} &  Q_{\mathrm{rot}_y}(\bm{q})^{\mathrm{T}} & Q_{\mathrm{rot}_z}(\bm{q})^{\mathrm{T}} \end{array} \right]^{\mathrm{T}} $.

We then introduce a tracking error $\mbox{\boldmath $e$}$ between the desired value and the system output, along with its integral value:
\begin{equation}
\label{eq:tracking_error}
\dot{\bm{e}}_{I_{\alpha}}  = \bm{y}^{\mathrm{des}} - \bm{y} = C \bm{x}^{\mathrm{des}} - C\bm{x} = C \bm{e}_x
\end{equation}

Using \equref{eq:tracking_error}, \equref{eq:rot_state_equation1} and  \equref{eq:rot_state_equation2} can be extended as follows:
\begin{eqnarray}
\label{eq:ex_rot_state_equation2}
& \dot{\bar{\bm{x}}} = \bar{A} \bar{\bm{x}} + \bar{B} \bm{u} + \bar{D} (\bm{\Delta}_{\mathrm{rot}} - I^{-1}\bm{\omega} \times I \bm{\omega})\\
&  \bar{\mbox{\boldmath $x$}} = \left[
  \begin{array}{c}
    \bm{e}_x \\
    \bm{e}_{I_{\alpha}}
    \end{array}
  \right]  ; \hspace{3mm}
  \bar{A} =\left[
  \begin{array}{cc}
    A & 0_{8 \times 3}\\
    C & 0_{3 \times 3}
    \end{array}
  \right]  ; \hspace{3mm}
  \bar{B} =\left[
  \begin{array}{c}
    -B  \\
    0_{3 \times 4}
    \end{array}
  \right] ; \hspace{3mm}
  \bar{D} =\left[
  \begin{array}{c}
    -D \\
    0_{3 \times 3}
    \end{array}
  \right] \nonumber
\end{eqnarray}

In the cost function of a general optimal control, there are two basic terms: the term regarding the state to change the convergence performance  and the term regarding the control input to restrict the input magnitude.
However, in this work, the suppression of the force generated by the thrust force, especially the horizontal force (i.e., \equref{eq:horizontal_force}) should be taken into account in the attitude control. Then an original cost function for this model is designed as follows:
\begin{eqnarray}
\label{eq:lqi-cost-func}
J = \int^{\infty}_0 \left(\bar{\bm{x}}^{\mathrm{T}} M  \bar{\bm{x}} +  \tilde{\bm{u}}^{\mathrm{T}} N \tilde{\bm{u}} \right)dt \\
\label{eq:lqi-input-cost}
N = W_1 +  Q_{\mathrm{tran}}^{\mathrm{T}}(\bm{q}) W_2 Q_{\mathrm{tran}}(\bm{q})
\end{eqnarray}
where, the second term in \equref{eq:lqi-input-cost} corresponds to the minimization of the norm of the force generate by the attitude control, since $\|{}^{\{CoG\}}\bm{f}\|^2 = {}^{\{CoG\}}\bm{f}^{\mathrm{T}}{}^{\{CoG\}}\bm{f} = u^{\mathrm{T}} Q_{\mathrm{tran}}^{\mathrm{T}}(\bm{q}) ^{\mathrm{T}} Q_{\mathrm{tran}}^{\mathrm{T}}(\bm{q})  u $. The diagonal weight matrices $M$, $W_1$ and $W_2$ balance the performance of the convergence to the desired state, the suppression of the control input and the suppression of the translational force generated by the attitude control.

Then, a constant feed-back gain matrix $K_x$ can be obtained by solving the related algebraic Riccati equation (ARE) driven by the equation  \equref{eq:ex_rot_state_equation2} and the cost function \equref{eq:lqi-cost-func}. Finally, the desired control input regarding the attitude control $\bm{u}^{\mathrm{des}}_{\mathrm{att}}$ can be given by:
\begin{eqnarray}
  \label{eq:deisred_att_thrust_force}
  \bm{u}^{\mathrm{des}}_{\mathrm{att}} = K_x\bar{\bm {x}} + Q^{\#}_{\mathrm{rot}}(\bm{q}) I^{-1} \bm{\omega} \times I \bm{\omega}
\end{eqnarray}

\subsection{Position Control}
The position control follows the method presented in \cite{lee-quadrotor-se3-control, lee-quadrotor-pid-se3-control}, which first calculates the desired total force based on the common PID control, then converts the desired total force to the desired thrust vector and the deisred roll and pitch angles.

The desired total force can be given by:
\begin{eqnarray}
  \label{eq:pid_pos}
        {\bm f}^{\mathrm{des}} &=& m_{\Sigma} (K_P \bm{e}_r + K_I \int (\dot{\bm{e}}_{r} + c\bm{e}_{r} )dt + K_D \dot{\bm{e}}_{r} + \ddot{\bm r}^{\mathrm{des}}) - {}^{\{W\}}R_{\{CoG\}} Q_{\mathrm{tran}}(\bm{q}) Q^{\#}_{\mathrm{rot}}(\bm{q})  I^{-1} \bm{\omega} \times I \bm{\omega} \nonumber \\
        &=& m_{\Sigma} (K_P \bm{e}_r + K_I \bm{e}_{I_r} + K_D \dot{\bm{e}}_{r} + \ddot{\bm r}^{\mathrm{des}}) + \bm{\phi}
\end{eqnarray}
where $\bm{e}_r = \bm{r}^{\mathrm{des}} - \bm{r}$, and $c$ is a positive constant. Note that the offset term $\bm{\phi}$ is added to compensate the force caused by the attitude control (i.e., the second term of \equref{eq:deisred_att_thrust_force}).

Then the desired roll and pitch angles can be also calculated from ${\bm f}^{\mathrm{des}}$:
\begin{eqnarray}
  \label{eq:desired_roll}
  &\alpha_{\{CoG\}_x}^{\mathrm{des}} = atan^{-1}(-\bar{f}_y, \sqrt{\bar{f}^2_x + \bar{f}^2_z}) \\
  \label{eq:desired_pitch}
  &\alpha_{\{CoG\}_y}^{\mathrm{des}} = atan^{-1}(\bar{f}_x, \bar{f}_z) \\
  & \bar{\bm f} = R_z^{-1}(\alpha_z^{\mathrm{des}}) {\bm f}^{\mathrm{des}} \nonumber
\end{eqnarray}
where, $R_{Z}(\cdot)$ is a special rotation matrix which only rotates along $z$ axis. Note that these two values are subsequently transmitted to the attitude controller as shown in \figref{figure:control_diagram}.

On the other hand,  the desired collective thrust force can be calculated as follows:
\begin{eqnarray}
  \label{eq:collective thrust}
  f_T^{\mathrm{des}} = ( {}^{\{W\}}R_{\{CoG\}} \bm{b}_3)^{\mathrm{T}} {\bm f}^{\mathrm{des}} ,
\end{eqnarray}
where,  $\bm{b}_3$ is a unit vector $\left[\begin{array}{ccc}0 & 0& 1 \end{array} \right]^{\mathrm{T}}$.

Using \eqref{eq:collective thrust}, the allocation from the collective thrust force to the thrust force vector can be performed as follows:
\begin{eqnarray}
  \label{eq:collective_thrust2}
  \bm{u}^{\mathrm{des}}_{\mathrm{pos}} = \frac{\bm{u}_s}{m_{\Sigma} g} f_T^{\mathrm{des}},
\end{eqnarray}
where, $\bm{u}_s$ is the thrust force vector under the zero equilibrium as expressed by \equref{eq:u_s2}, which only balances with the gravity force and thus does not affect rotational motion. Eventually, the final deisred thrust force for each rotor should be the sum of output from bthe attitude controller and the position control: $\bm{u}^{\mathrm{des}} = \bm{u}^{\mathrm{des}}_{\mathrm{att}} + \bm{u}^{\mathrm{des}}_{\mathrm{pos}}$.

\subsection{Stability Analysis}
For a general multirotor, the exponential stability of the cascaded control method is well-studied by \cite{lee-quadrotor-se3-control, lee-quadrotor-pid-se3-control}.
However, the full influence   $Q_{\mathrm{tran}}(\bm{q}) \bm{u}$  on all translational axes as shown in \equref{eq:translational_dynamics} makes the problem difficult. Thus, it is required to clarify the new stable condition regarding the feed-back control gains in \equref{eq:lqi-cost-func} and    \equref{eq:pid_pos}.

\subsubsection{Attitude Stability}
Is it well-known that the LQI control framework guarantees the exponential stability. In order to show the stability of the complete system, we design a proper lyapunov function candidate for attitude control based on the error dynamics  \equref{eq:ex_rot_state_equation2}  as follows:
\begin{eqnarray}
  \label{eq:rot-lyapunov}
& \mathcal{V}_1 = \frac{1}{2} \bar{\bm{e}}_{\mathrm{rot}}^2 \\
 & \bm{e}_{\mathrm{rot}} = \bar{\bm{x}}  - \left[\begin{array}{cc} \bm{0}_6 &  \bm{\Delta}_{\mathrm{rot}} \circ \frac{1}{Q_{\mathrm{rot}}(\bm{q}) K_{x_I}}  \end{array}\right]^{\mathrm{T}} \nonumber.
\end{eqnarray}
where $(\cdot) \circ (\cdot)$ denote the element-wise multiplication of two vectors, and $K_{x_I} \in \mathcal{R}^{4 \times 3}$ corresponds to the right three columns of the gain matrix $K_x$. Thus  $\bm{\Delta}_{\mathrm{rot}} \circ \frac{1}{Q^{\#}_{\mathrm{rot}}(\bm{q}) K_{x_I}}$ is the convergent integral value of $\bm{e}_{I_{\alpha}}$.

Then, the time derivative of $\mathcal{V}_1$ is given by:
\begin{eqnarray}
  \label{eq:rot-dot-lyapunov}
  \dot{\mathcal{V}}_1 &=& \bm{e}_{\mathrm{rot}}^{\mathrm{T}} \dot{\bm{e}}_{\mathrm{rot}} = \bm{e}_{\mathrm{rot}}^{\mathrm{T}} \dot{\bar{\bm{x}}} = \bm{e}_{\mathrm{rot}}^{\mathrm{T}} (\bar{A} \bar{\bm{x}} + \bar{B} \bm{u} + \bar{D}(\bm{\Delta}_{\mathrm{rot}} - I^{-1}\bm{\omega} \times I \bm{\omega})) \\ \nonumber
  &=&  \bm{e}_{\mathrm{rot}}^{\mathrm{T}} ((\bar{A} +  \bar{B} K_x) \bar{\bm{x}}  +  \bar{D} \bm{\Delta}_{\mathrm{rot}})) = \bm{e}_{\mathrm{rot}}^{\mathrm{T}} (\bar{A} +  \bar{B} K_x) \bm{e}_{\mathrm{rot}} < 0.
\end{eqnarray}
$\dot{\mathcal{V}}_1$ should be always negative since all eigenvalues of $(\bar{A} +  \bar{B} K)$ are negative, and thus the attitude control guarantees the exponential stability. However, the derivation of \equref{eq:rot-dot-lyapunov} is only valid in the near-hover state (i.e., $\bm{\omega} \approx \dot{\bm{\alpha}}$), otherwise the state equation \equref{eq:rot_state_equation1} would not be established.

\subsubsection{Complete Stability}

The dynamics of position error can be given by:
\begin{eqnarray}
  \label{eq:pos-err-dynamics2}
  m \ddot{\bm{e}}_{r}  = - m K_P \bm{e}_r - m K_I \bm{e}_{I_r} - m K_D \dot{\bm{e}}_r + bm{X} + R Q^{'}_{\mathrm{tran}} \bm {e}_{\mathrm{rot}}  + \bm{\Delta}_{\mathrm{tran}} + \bm{\Delta}^{'}_{\mathrm{rot}},
\end{eqnarray}
where  $\bm{\Delta}^{'}_{\mathrm{rot}} =  Q_{\mathrm{tran}} Q^{\#}_{\mathrm{rot}} \bm{\Delta}_{\mathrm{rot}}$, $Q^{'}_{\mathrm{tran}} = Q_{\mathrm{tran}} K_x$ and $\bm{X} = \|{\bm f}^{\mathrm{des}}\| ( ((R^{\mathrm{des}} \bm{b}_3)^{\mathrm{T}}  R  \bm{b}_3 )R \bm{b}_3  -  R^{\mathrm{des}} \bm{b}_3 )$. For convenience  ${}^{\{W\}}R_{\{CoG\}}$ and  $m_{\Sigma}$  are simplified as $R$ and $m$. We refer the reader to Appendix A, where the we provide the detailed derivation of \equref{eq:pos-err-dynamics2}. 

Then the integral Lyapunov candidate $\mathcal{V} = \mathcal{V}_1 + \mathcal{V}_2$ for the complete system is written as follows:
\begin{eqnarray}
  \label{eq:integral_lyapunov_candidate}
&  \mathcal{V} = \mathcal{V}_1 + \mathcal{V}_2 \\
  \label{eq:pos-lyapunov}
&  \mathcal{V}_2 = \frac{1}{2} \bm{e}_{r}^{\mathrm{T}} K_P \bm{e}_{r} +  \frac{1}{2} \|\dot{\bm{e}}_{r}\|^2 + c \bm{e}^{\mathrm{T}}_{r} \dot{\bm{e}}_{r} + \frac{1}{2} (\bm{e}_{I_r} - \bm{\Delta} \circ \frac{1} {\bm{k}_{I_r}})^{\mathrm{T}} K_{I_r} (\bm{e}_{I_r} - \bm{\Delta} \circ \frac{1} {\bm{k}_{I_r}})
\end{eqnarray}
where, $\mathcal{V}_2$ is the  Lyapunov candidate  regarding the position error dynamics, and the gain matrix $K_P$ and the positive constant $c$ corresponds to   \equref{eq:pid_pos}.
Also note that, $\bm{k}_{I_r} \in \mathcal{R}^3$ is the diagonal elements of $K_{I_r}$: $K_{I_r} = diag(\bm{k}_{I_r})$, and $\bm{\Delta} = \frac{\bm{\Delta}_{\mathrm{tran}} + \bm{\Delta}^{'}_{\mathrm{rot}}}{m}$.

In order to guarantee $\mathcal{V} > 0$ and $\dot{\mathcal{V}} < 0$, following constraints  should be satisfied
\begin{eqnarray}
  \label{eq:kp_constraints}
  &k_{P_{\mathrm{min}}} > \gamma k_{P_{\mathrm{max}}} \\
  \label{eq:c_constraints}
  &c < \mathrm{min}\{\frac{4(k_{P_{\mathrm{min}}} - \gamma k_{P_{\mathrm{max}}})(k_{D_{\mathrm{min}}} - \gamma k_{D_{\mathrm{max}}})}{k_{D_{\mathrm{max}}}^2 (1+\gamma)^2 + 4(k_{P_{\mathrm{min}}} - \gamma k_{P_{\mathrm{max}}})} ,  k_{D_{\mathrm{min}}} - \gamma k_{D_{\mathrm{max}}}, \sqrt{k_{P_{\mathrm{min}}}}\} \\
  \label{eq:gains_constraints}
  &\lambda_{\mathrm{min}}(W_1) \lambda_{\mathrm{min}}(W_2) > \frac{\sigma^2_{\mathrm{max}}(W_{12})}{4}
\end{eqnarray}
where, 
\begin{eqnarray}
  W_1 = \frac{1}{2}\left[\begin{array}{cc} c(k_{P_{\mathrm{min}}} - \gamma k_{P_{\mathrm{max}}}) & - \frac{ck_{D_{\mathrm{max}}}}{2}(1 + \gamma) \\ - \frac{ck_{D_{\mathrm{max}}}}{2}(1 + \gamma) & -c + k_{D_{\mathrm{min}}} - \gamma k_{D_{\mathrm{max}}}) \end{array}\right],
  W_{12} = \left[\begin{array}{c} \frac{c(\sigma_{max}(Q^{'}_{\mathrm{tran}}) + O)}{m} \\  \frac{\sigma_{max}(Q^{'}_{\mathrm{tran}}) + B}{m} + k_{P_{\mathrm{max}}} \bm{e}_{r_{\mathrm{max}}} \end{array}\right],
  W_2 = - (\bar{A} +  \bar{B} K_x) \nonumber
\end{eqnarray}
Note that, $\lambda_{\mathrm{max}}(\cdot)$  and $\lambda_{\mathrm{min}}(\cdot)$ are the maximum and minimum eigenvalue of a matrix, while $\sigma_{max}(\cdot)$ denotes the maximum singular value of a matrix.  $k_{P_{\mathrm{max}}}$, $k_{P_{\mathrm{min}}}, k_{D_{\mathrm{max}}}$ and $k_{D_{\mathrm{min}}}$ are the maximum and minimum elements in the gain matrix  $K_P$ and $K_D$, respectively. $\gamma$ is a positive constant, i.e., $\gamma \leq sin(\|\bm{e}_{\alpha}\|) < 1$. while $O$ is also a positive constant which satisfies \equref{eq:acc-cond1}. $\bm{e}_{r_{\mathrm{max}}}$ is the upper bound of the postion error. We refer the reader to Appendix B, where the we provide the detailed derivation of \equref{eq:kp_constraints} $\sim$   \equref{eq:gains_constraints}.

According to \equref{eq:gains_constraints}, if we increase the propeller tilt angle, $\sigma^2_{\mathrm{max}}(W_{12})$ would increase because of the positive correlation with the sine of tilt angle. Then, it is necessary to increase either the attitude or position gains and further increase the left side of \equref{eq:gains_constraints}. However, the increase of the position gains would make the change of desired attitude  more aggressive according to  \equref{eq:desired_roll} and \equref{eq:desired_pitch},  and thus  the violation of  the near-hover assumption would be easier to occur. Therefore,  changing the attitude gains should be more effective.


\section{State Estimation}
\label{sec:estimation}
\begin{figure}[h]
  \begin{center}
    \includegraphics[width=\columnwidth]{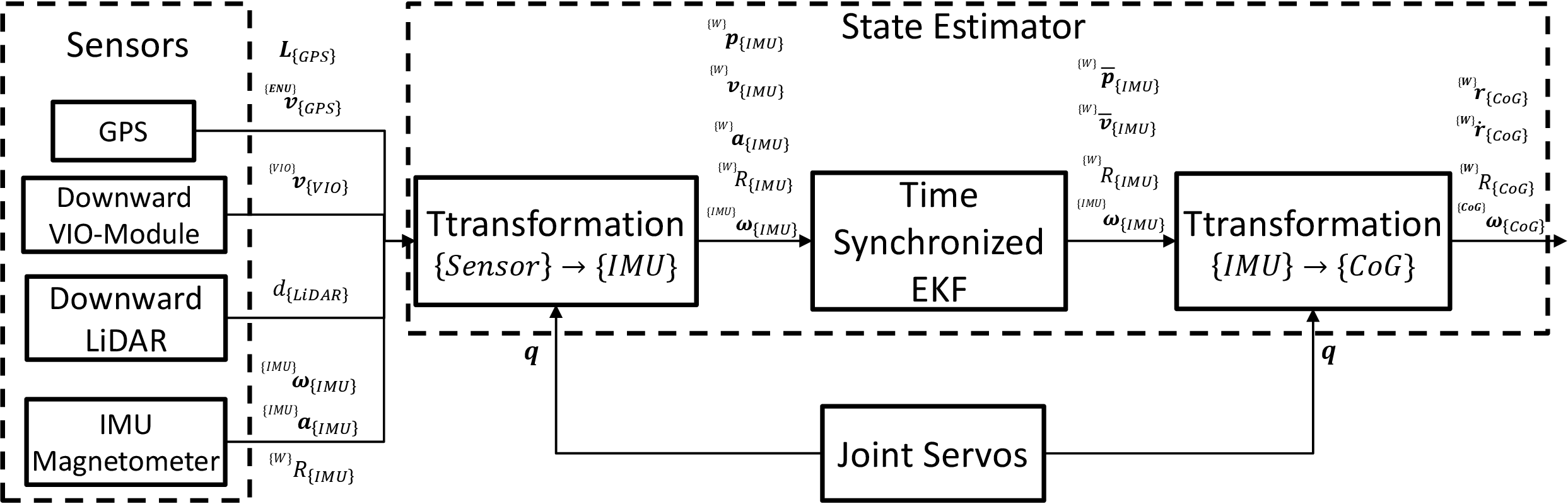}
    \caption{The framework of the state estimation for fully autonomous flight. Sensors provide sensor values w.r.t. each sensor frame. Then the transformation from each sensor frame to the frame $\{IMU\}$ is performed to unify these sensor values according to the kinematics involving joint angles $\bm{q}$,  which is followed by a time-synchronized extended Kalman filter to generate the fused odometry. Note that the symbols with bar are the estimated output of EKF. Finally the odometry is transformed to the  frame $\{CoG\}$ for flight control.}
    \label{figure:state-estimation}
  \end{center}
\end{figure}

In this work, the state estimation  for the multilinked aerial robot to achieve fully autonomous flight in outdoor environment is also developed. The crucial issue regarding the multilinked model is the necessity to transform the sensor value from each sensor frame to a common frame for the sensor fusion and then transform again to the frame $\{ CoG\}$ for control. These transformation processes involve the joint angles $\bm{q}$ as shown in \figref{figure:state-estimation}. On the other hand, the time synchronization is also considered in the extended Kalman filter to solve the delay of sensor measurement. 

\subsection{Core Sensors}

\subsubsection{GPS}
Global positioning system (GPS) mainly provides the latitude and longitude $L_{\{GPS\}} = \left[lat \ lon\right]^{\mathrm{T}}$ based on WGS84 \footnote[1]{WGS84: WORLD GEODETIC SYSTEM 1984}, which can be straightforwardly converted into a metric position around a reference point with an east-north-up (ENU) coordinate. In other words,  GPS can provide the global horizontal position, namely ${}^{\{W\}}\bm{p}_{\{GPS\}_{xy}}$ w.r.t a world frame $\{W\}$ of which the x and y axis coincide with the north and east direction, respectively. Such definition also benefits the integration with  magnetometer. Furthermore, the latest GPS module can also provide the global velocity  w.r.t ENU coordinate which is calculated according to relative motion between the GPS module and each satellite.

On the other hand, there are couple of problems with GPS module. First, the  accuracy highly depends on the satellite number and the weather condition, which however can be solved by fuseing multiple sensors. Second, the delay of measurement is most significant compared with other sensors, which can reach 0.5 s. Given that such a delay would worsen the performance of the sensor fusion, the time-synchronized EKF framework is developed in this work.

\subsubsection{Downward VIO-Module}
A VIO Module can provide a 6DoF odometry from either a monocular or a stereo camera combined with IMU, which is call visual-inertial-odometry (VIO). In an outdoor environment, the visual processing from a horizontal has the problem of the insufficient feature points. Then the downward view is applied. However, the density of valid feature points is still lower than an indoor environment. Thus the position estimated by a VIO module is omitted since it highly depends on the feature map, and only the relative velocity ${}^{\{VIO\}}\bm{v}_{\{VIO\}}$  is used in our framework.

\subsubsection{Downward LiDAR}
A downward Light Detection and Ranging (LiDAR) sensor measures  one dimension distance $d_{\{LiDAR\}}$ from the sensor to the ground. In most cases, the outdoor flight is performed upon a flat ground and the robot orientation would not change significantly from the hovering state. Thus the height ${}^{\{W\}}{p}_{\{LiDAR\}_{z}}$ from the ground can be given by
\begin{equation}
  \label{eq:aerointerference}
  {}^{\{W\}}{p}_{\{LiDAR\}_{z}} = \left[0 \ 0 \ 1\right]  {}^{\{W\}}R_{\{LiDAR\}} \left[d_{\{LiDAR\}} \ 0 \ 0\right]^{\mathrm{T}}
\end{equation}
where,  ${}^{\{W\}}R_{\{LiDAR\}}$ denotes the orientation of the LiDAR sensor. Note that the x axis of the frame $\{LiDAR\}$ coincides with the direction of light emission. 

\subsubsection{IMU and Magnetometer}
The rotational motion can be estimated by the accelerometer and gyroscope in IMU with magnetometer. Among various effective estimation algorithms (e.g. \cite{madgwick, ekf-att-est}), the most computationally light algorithm, namely the complementary filter \cite{complementary_filter_martin}, is applied in this work to calculate the orientation in a microcontroller. The orientation estimated by IMU and magnetometer is also based on the ENU coordinate. On the other hand, the angular velocity and the linear acceleration measured by IMU are also used in subsequent frame transformation and the extended Kalman filter as shown in \figref{figure:state-estimation}.
It is also notable that, the VIO algorithm  can also provide the estimated orientation; however, IMU and Magnetometer are the most robust and reliable sensors in fields, and thus a separated orientation estimation only by these two sensors is designed as shown in \figref{figure:state-estimation}.

\subsection{Transformation from Sensor Frame to IMU Frame}
The extended Kalman filter requires all sensor values to be described at a common frame. Then the frame  $\{IMU\}$  is chosen as the common frame to avoid the necessity of the linear and angular accleration in the transformation process. The position and velocity provided by each sensor is converted  with following rules:
\begin{eqnarray}
  \label{eq:pos_imu_convert}
  & {}^{\{W\}}\bm{p}_{\{IMU\}} = {}^{\{W\}}\bm{p}_{\{S\}} - {}^{\{W\}}R_{\{IMU\}} {}^{\{IMU\}}\bm{p}_{\{IMU\} \rightarrow \{S\}}(\bm{q}) \\
  \label{eq:vel_imu_convert}
  & {}^{\{W\}}\bm{v}_{\{IMU\}} = {}^{\{W\}}R_{\{IMU\}} ( {}^{\{IMU\}}R_{\{S\}}(q){}^{\{S\}}\bm{v}_{\{S\}}  - {}^{\{IMU\}}\bm{\omega}_{\{IMU\}} \times  {}^{\{IMU\}}\bm{p}_{\{IMU\} \rightarrow \{S\}}(\bm{q}))
\end{eqnarray}
where, $\{S\}$ denotes the sensor frame, and ${}^{\{W\}}\bm{p}_{\{IMU\} \rightarrow \{S\}}(\bm{q})$ is the position vector from  $\{IMU\}$ to $\{S\}$ which involves the joint angles $\bm{q}$.
Note that, \equref{eq:pos_imu_convert} is used for the conversion about  the GPS and Downward LiDAR, while \equref{eq:vel_imu_convert} is used for the conversion of the VIO module.

\subsection{Time Synchronized Extended Kalman Filter}
The estimation state $\bm{x}$  in EKF holds the position ${}^{\{W\}}\bm{p}_{\{IMU\}}$  and velocity ${}^{\{W\}}\bm{v}_{\{IMU\}}$ of the  frame $\{IMU\}$ and the bias of the acceleration $\bm{b}_{\mathrm{acc}}$:
$  \bm{x} := \left[\begin{array}{ccc} {}^{\{W\}}\bm{p}_{\{IMU\}}^{\mathrm{T}}  & {}^{\{W\}}\bm{v}_{\{IMU\}}^{\mathrm{T}}  &  \bm{b}_{\mathrm{acc}}^{\mathrm{T}} \end{array}\right]^{\mathrm{T}}
$.
The input for prediction is the accleration obtained from IMU sensor ${}^{\{IMU\}}\bm{a}_{\{IMU\}}$, while the accleration bias $\bm{b}_{\mathrm{acc}}$ is modeled as random walk with their derivatives being white gaussian noise. Then the prediction model can be expressed as a simple dynamics only involving position, velocity and acceleration. On the other hand, the measurement vector $\bm{z} := \left[\begin{array}{cc} {}^{\{W\}}\bm{p}_{\{IMU\}}^{\mathrm{T}}  & {}^{\{W\}}\bm{v}_{\{IMU\}}^{\mathrm{T}} \end{array}\right]^{\mathrm{T}}$ contains the position and velocity which are obtained from sensors other than IMU, implying the observation model is a simple linear matrix.

\begin{figure}[h]
  \begin{center}
    \begin{minipage}{0.49\hsize}
      \begin{center}
        \includegraphics[width=1.0\columnwidth]{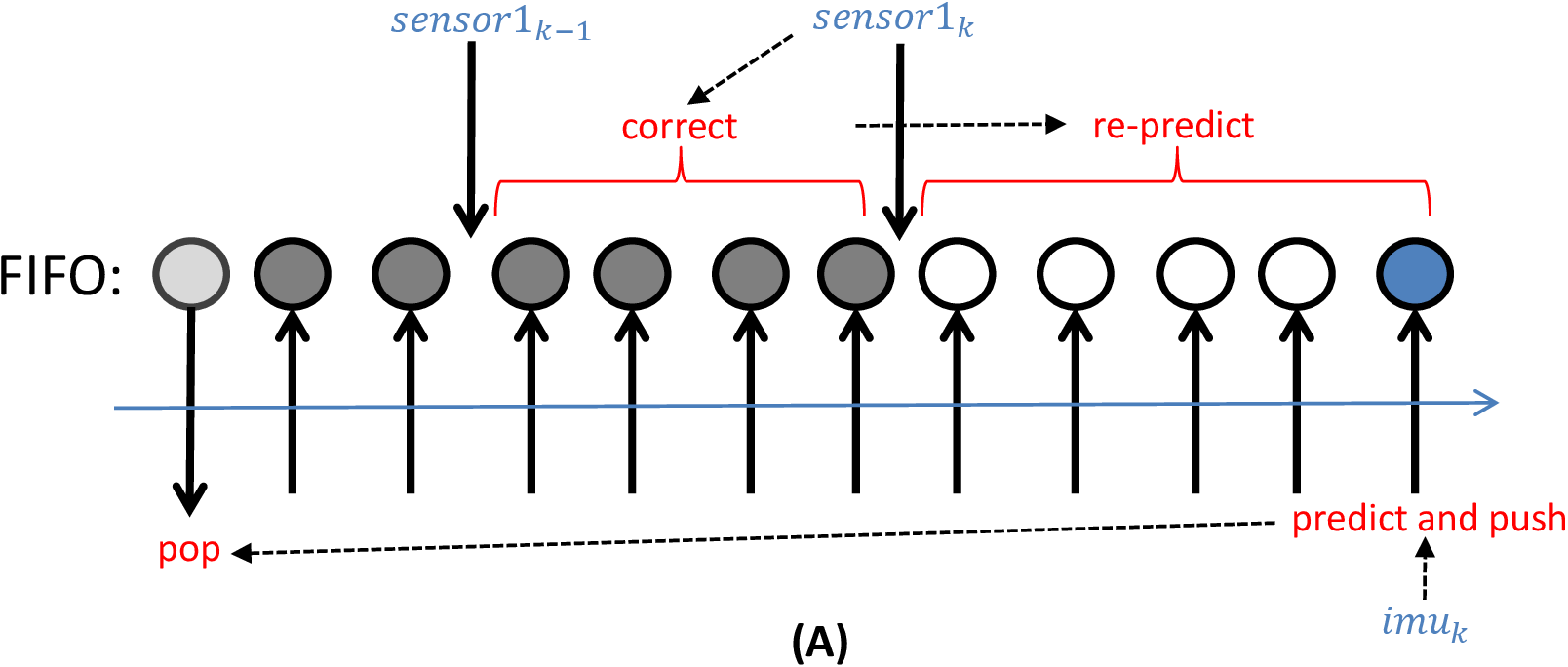}
      \end{center}
    \end{minipage}
    \begin{minipage}{0.49\hsize}
      \begin{center}
        \includegraphics[width=1.0\columnwidth]{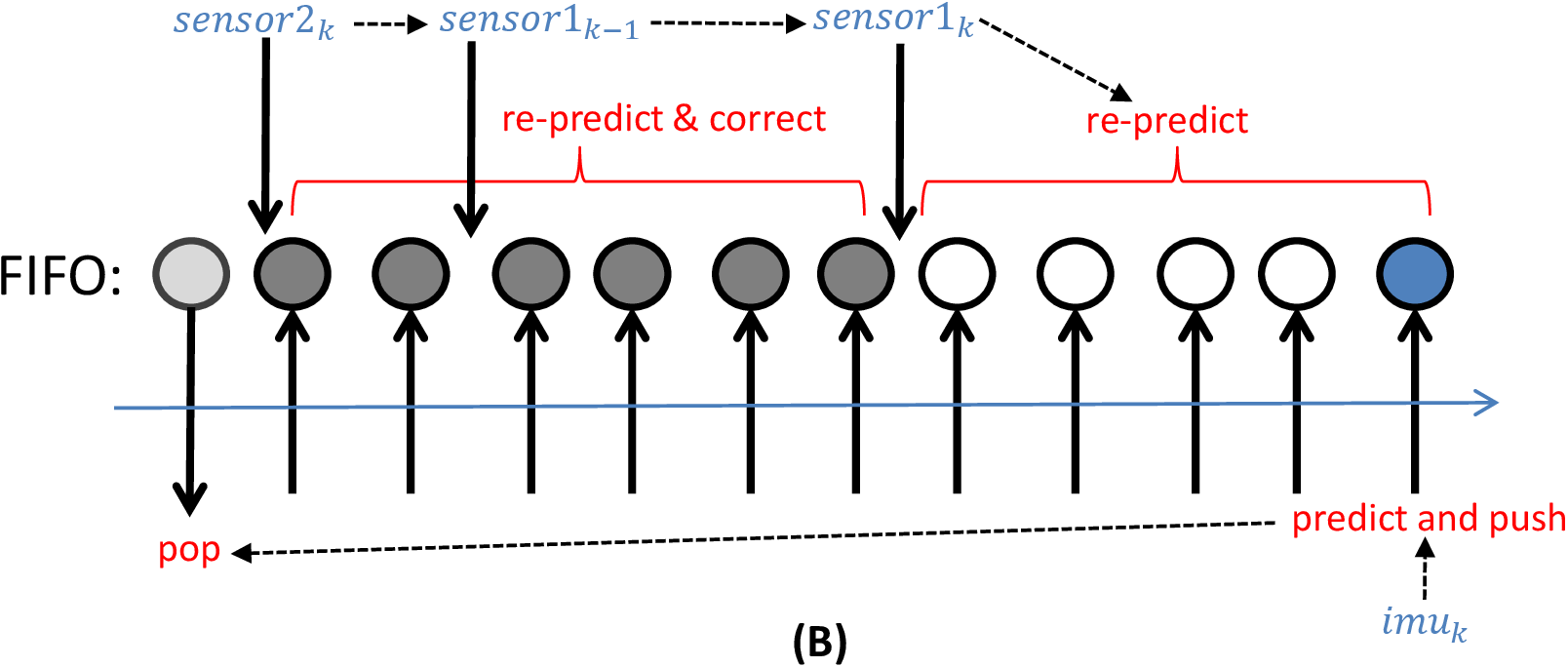}
      \end{center}
    \end{minipage}
    \caption{The time-synchronized extended Kalman filter framework which contains a FIFO buffer sorted by timestamp. The push motion is performed when a new imu value arrives, and the pop motion is performed in the same time to keep the buffer length fixed. There are two cases of correction with delayed sensor value.  {\bf (A)}:  the case when the sensor value with latest sensor timestamp $sensor1_k$ arrives. {\bf (B)}: the case when the further delayed sensor value which is older than the latest sensor timestamp $sensor2_k$ arrives. The dashed arrows indicate the process sequence when either a new imu value or a sensor value arrives.}
    \label{figure:time-sync-ekf}
  \end{center}
\end{figure}

An important issue in sensor fusion is time synchronization among sensors, since ignoring the measurement delay would significantly decrease in estimation performance. Therefore, a time-synchronized Kalman Filter framework is developed based on \cite{time-sync-ekf-eth} as shown in \figref{figure:time-sync-ekf}. The key of this framework is the FIFO structure which enables the correction in the past node. Given that the sensor value from IMU has the smallest delay, the arrival of new IMU state serves as the trigger to perform prediction. On the other hand, the correction process is performed on-demand  when a new sensor value from other sensors is arrived as shown in \figref{figure:time-sync-ekf}. There two two cases of correction: (A) shows the case when a new sensor value with the latest timestamp $sensor1_k$ arrives which requires correction, whereas (B) shows the case when a further delayed sensor value $sensor2_k$  arrives which requires both re-prediction and re-correction to the latest sensor timestamp. In both of cases, re-prediction from the latest sensor timestamp to the latest imu timestamp is also required, and the covariance is not predicted during this phase since it is not necessary for the control. Thus, the successive prediction of covariance  is performed on-demand during the correction phase to reduce the computational cost.

\subsection{Transformation from IMU Frame to CoG Frame}
The estimated odometry regarding the frame $\{IMU\}$ should be finally converted to the frame  $\{CoG\}$ with following rule
\begin{eqnarray}
  \label{eq:r_cog_convert}
  & {}^{\{W\}}R_{\{CoG\}} =  {}^{\{W\}}R_{\{IMU\}} {}^{\{IMU\}}R_{\{CoG\}}(\bm{q}) \\
  \label{eq:omega_cog_convert}
  & {}^{\{CoG\}}\bm{\omega}_{\{CoG\}} =   {}^{\{CoG\}}R_{\{IMU\}}(\bm{q}) {}^{\{IMU\}}\bm{\omega}_{\{IMU\}} \\
  \label{eq:pos_cog_convert}
  & {}^{\{W\}}\bm{p}_{\{CoG\}} = {}^{\{W\}}\bar{\bm{p}}_{\{IMU\}} + {}^{\{W\}}R_{\{IMU\}} {}^{\{IMU\}}\bm{p}_{\{IMU\} \rightarrow \{CoG\}}(\bm{q}) \\
  \label{eq:vel_cog_convert}
  & {}^{\{W\}}\bm{v}_{\{CoG\}} =  {}^{\{W\}}\bar{\bm{v}}_{\{IMU\}} + {}^{\{W\}}R_{\{IMU\}} ({}^{\{IMU\}}\bm{\omega}_{\{IMU\}} \times  {}^{\{IMU\}}\bm{p}_{\{IMU\} \rightarrow \{S\}}(\bm{q}))
\end{eqnarray}
where, ${}^{\{W\}}\bar{\bm{p}}_{\{IMU\}}$ and ${}^{\{W\}}\bar{\bm{v}}_{\{IMU\}}$ are the estimated result from the time-synchronized extend EKF, and ${}^{\{IMU\}}R_{\{CoG\}}(\bm{q})$ is obtained from the forward-kinematics and \equref{eq:R_CoG2}. Finally the converted odometry regarding the frame  $\{CoG\}$ is used as the feed-back state in the control system as shown in \figref{figure:control_diagram}.

\section{Platform}
\label{sec:platform}

\subsection{Mechanical Specification}
\begin{figure}[h]
    \begin{center}
      \includegraphics[width=\columnwidth]{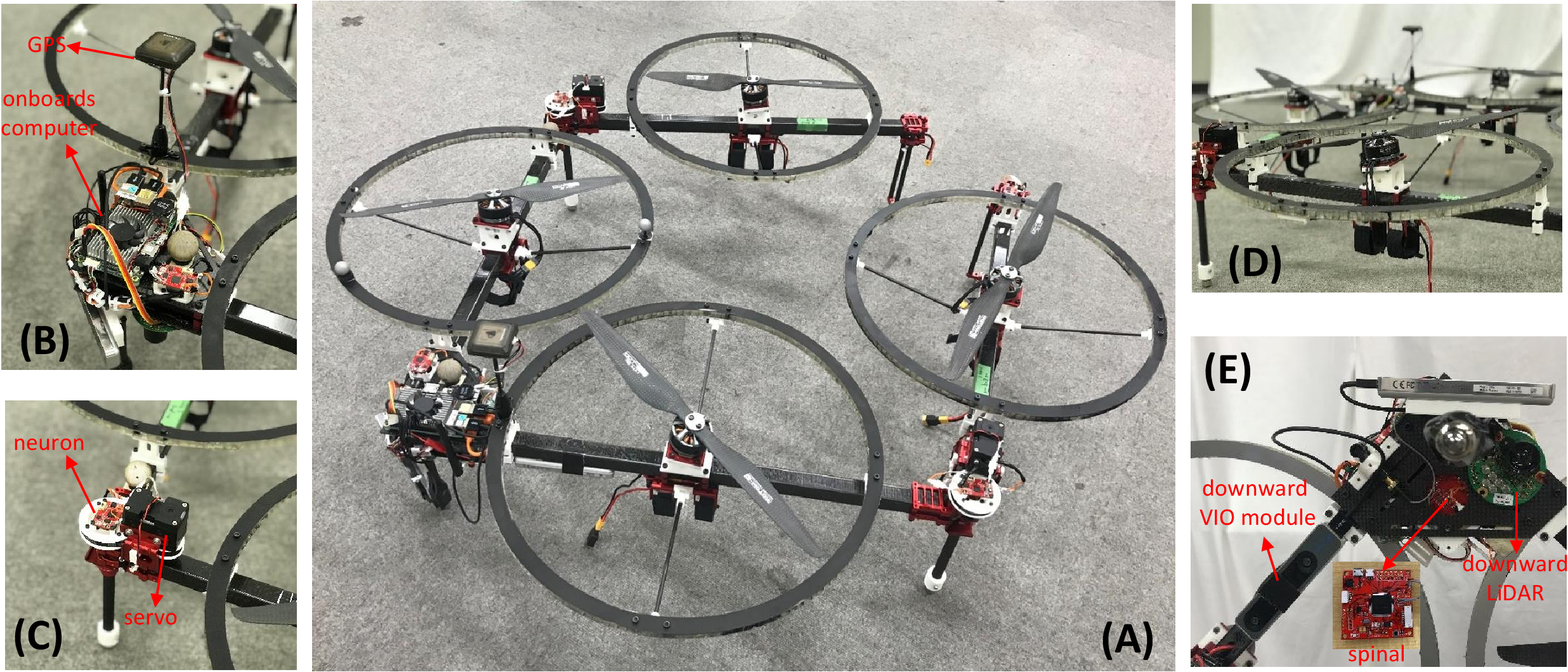}
    \end{center}
    \caption{{\bf (A)}: robot platform composed from three parts which are connected by two joints. For convenience, we regard the central part as two separated links, namely link2 and link3. The link length is 0.6m. {\bf (B)}: the top of the connection part between link2 and link3. {\bf (C)}: the top of the joint part which is  actuated by a servo motor. {\bf (D)}: the tilting propeller in each link. {\bf (E)}: the bottom of the connection part between link2 and link3. The original micro controller units called {\it spinal} and {\it neuron}s are connected by Controller Area Network (CAN) for the internal communication between link modules.}
    \label{figure:hardware}
\end{figure}

The hardware decisions  are led by the need to create the synergy among different tasks, which leads to an original hardware platform as shown in \figref{figure:hardware}.
The link rod of which the length is 0.6 m is make from the carbon square pipe, and cables pass inside the rod.  Other white and red components are made from PLA and Aluminum respectively, which are used in different places according to the desired strength.

Two joints are actuated by servo motors (Dynamixel XH430-W350R \footnote[2]{See http://www.robotis.us/dynamixel-xh430-w350-r.}) of which the maximum torque are 4.2 Nm at 14.8 V. However, the maximum  torque might be not sufficient for grasping task since  large friction is required during delivery. Thus an enhanced joint design is developed in this work which applies pulleys (white disk-like parts in \figref{figure:hardware}(C)) to further amplify the torque. The pulley ratio is 1:2 which enables double torque output. It is notable that, such design enables to employ different servos and different pulleys  to achieve different joint torque characteristics. Such a configurable design is the main improvement compared with the servo-embedded structure as developed in our previous work \cite{hydrus-ijrr2018}.

The propulsion system is built based on T-motor products (rotors: T-motor MN3510 KV360\footnote[3]{See https://store-en.tmotor.com/goods.php?id=337.}; ESC: T-motor Air40A\footnote[4]{See https://store-en.tmotor.com/goods.php?id=368.}; propeller: T-motor P14$\times$4.8 Prop\footnote[5]{See https://store-en.tmotor.com/goods.php?id=380.})
, and the maximum thrust generated by this propulsion system is 16 N.
As shown in \figref{figure:hardware}(D), the propeller and rotor are mounted at a PLA component of which the top surface is inclined at an angle of 10 deg for tilting.
On the other hand, battery can be mounted below the propulsion system where there is a power cable connecting to the battery. Generally, a 6 s (22.2 V) Turnigy 1300 mAh battery is connected to each link module. Then the total weight of this basic platform with batteries is 3.4 Kg, which results in 15 min flight. However, it is possible to change the arrangement of batteries along with the number and capacity to different task, since all batteries are wired in parallel. In addition, a propeller protect duct with an aluminum honeycomb structure is employed not only for safety, but also for acting as gripper tip in grasping task.


\subsection{System Architecture}

\subsubsection{Processors}
\begin{figure}[!t]
  \begin{center}
    \includegraphics[width=0.8\columnwidth]{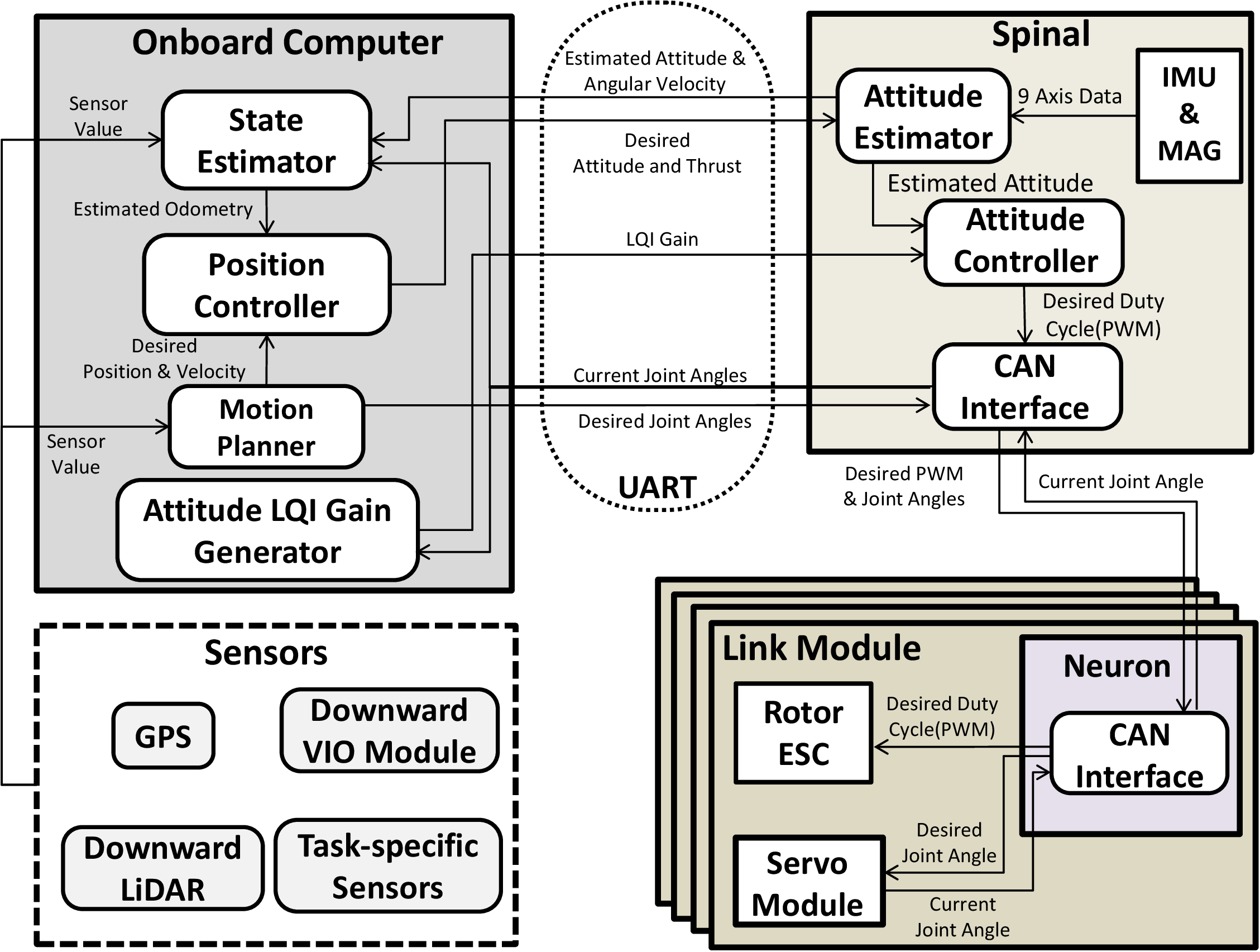}
  \end{center}
    \caption{The system architecture to achieve fully autonomous flight. }
    \label{figure:system}
\end{figure}

The onboard computer as shown in \figref{figure:hardware}(B) is UP Board\footnote[6]{https://www.aaeon.com/jp/p/up-board-computer-board-for-professional-makers.} with Intel Atom CPU running Ubuntu and the robot operating system (ROS)\footnote[7]{http://www.ros.org/.} as middleware to execute the state estimation, flight control and task-specific motion planning as shown in \figref{figure:system}. Other  onboard computers with higher computational performance (e.g. Intel NUC\footnote[8]{https://www.intel.com/content/www/us/en/products/boards‐kits/nuc.html.}) can be alternatively used for the task requiring the vision or point cloud processing.

On the other hand, the original printed circuit board (PCB) called {\it Spinal} as shown in \figref{figure:hardware}(E) is a micro controller unit (MCU) with a STM32F7 core to fulfill the real-time processing such as the attitude estimation and controller. As shown in \figref{figure:system}, a IMU and Magnetometer unit (InvenSense MPU9250\footnote[9]{https://invensense.tdk.com/products/motion-tracking/9-axis/mpu-9250/}) are embedded in {\it Spinal} to achieve the zero delay data transmission for attitude estimation to control. The message exchange between onboard computer and {\it Spinal} is achieved by the UART and rosserial based protocol\footnote[10]{http://wiki.ros.org/ja/rosserial}.

Another type of original MCU called {\it Neuron} with STM32F4 core is designed to directly connect to actuators (e.g., rotor ESC, joint servo) as shown in \figref{figure:system}. An internal communication system based on Controller Area Network (CAN)\cite{CAN} which is developed in our previous work \cite{hydrus-ijrr2018} connects {\it Spinal} and {\it Neuron}s, and an original message protocol inside CAN is designed  to pass ROS messages from the computer to each actuators as shown in  \figref{figure:system}.

\subsubsection{Sensors}
The core sensors for localization are an embedded IMU \& magnetometer, a GPS (u-blox M8 module\footnote[11]{https://www.u-blox.com/en/product/neo-m8-series}), a downward LiDAR (LedderOne\footnote[12]{https://leddartech.com/lidar/leddarone/}), and a downward-facing light VIO module (RealSense T265\footnote[13]{https://www.intelrealsense.com/tracking-camera-t265/})  as shown in \figref{figure:hardware}. Regarding the VIO module, RealSense T265 contains internal processor to calculate the visual odometry from an internal IMU and stereo fisheye cameras, which can significantly save the external computational resource. Furthermore, This device is relatively light ($\sim$ 80g) which is suitable for the aerial application.
On the other hand,  RTK-GPS is not applied in our platform since it can not afford the fully autonomous flight in wide area. Nevertheless, the proposed sensor employment can still promise the sufficient localization accuracy which will be shown in our outdoor experiments in \secref{experiments}.


\section{Experiments}
\label{sec:experiments}

We evaluated our robot platform by experiments in both indoor and outdoor environment. The video of our evaluation can be found at https://youtu.be/LkDGP82sg1I.
The control gain parameters for the real robot platform are summarized as follows:
\vspace{-5mm}
\begin{table}[!h]
  \begin{center}
    \caption{Control Gain Parameters for the Real Platform}
    \begin{tabular}{|c|c|c|}
    \hline
    Equation &  Parameter & Value \\ \hline
    \equref{eq:lqi-cost-func} & $M$  & $diag(1100, 80, 1100, 80, 100, 50, 10, 10, 0.5)$ \\ \hline
    \equref{eq:lqi-input-cost}  & $W_1$  & $diag(1, 1, 1, 1)$ \\ \cline{2-3}
    & $W_2$  & $diag(100, 100, 100)$ \\ \hline
    & $K_{P}$  & $diag(2.3, 2.3, 3.6)$ \\ \cline{2-3}
    \equref{eq:pid_pos}  & $K_I$  & $diag(0.02, 0.02, 3.4)$ \\ \cline{2-3}
    & $K_D$  & $diag(4.0, 4.0, 1.55)$ \\ \hline
    \end{tabular}
    \label{table:parameter}
  \end{center}
\end{table}
\vspace{-5mm}
\subsection{Indoor Experiments}
In order to evaluate the feasibility of proposed control method, several indoor flight experiments were conducted to testify the stability during deformation, the compensation ability regarding the unknown model error, and the robustness against the strong disturbance. A motion capture system was employed  instead of using the state estimator in all indoor experiments for the ground truth regarding robot odometry.

\subsubsection{Deformation Stability}
\begin{figure}[!h]
  \begin{center}
    \includegraphics[width=0.93\columnwidth]{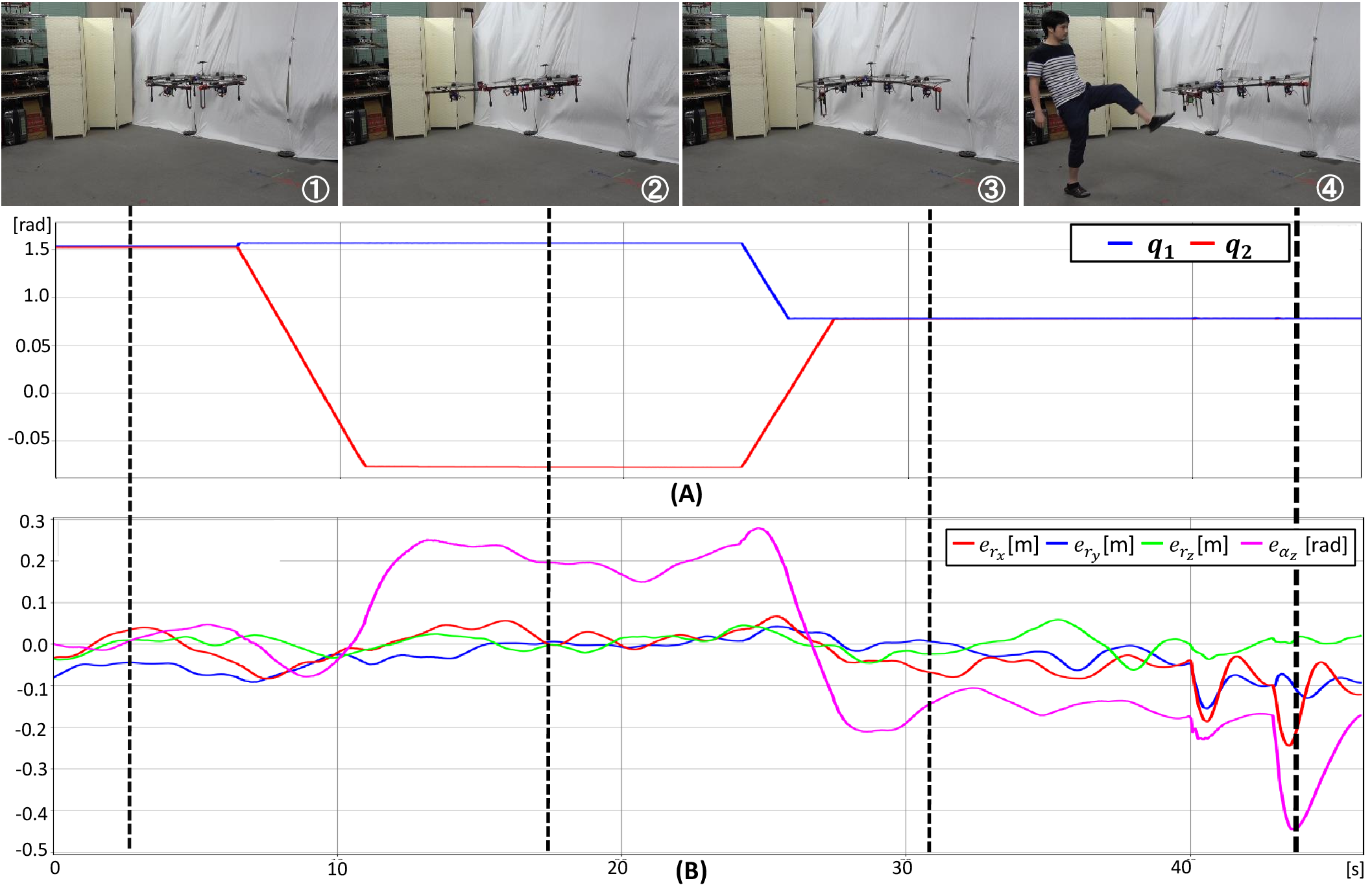}
    \caption{The evaluation of stable deformation (\textcircled{\scriptsize 2} and \textcircled{\scriptsize 3}) in an indoor environment using a motion capture system, along with the robustness against large impact \textcircled{\scriptsize 4}. {\bf (A)}: the change in joint angles during this experiment. {\bf (B)}: the tracking errors regarding the translational and yaw rotational motion. }
    \label{figure:indoor_hovering_transformation_kick}
    \vspace{-3mm}
  \end{center}
\end{figure}

The flight stability during deformation was evaluated as shown in \figref{figure:indoor_hovering_transformation_kick}. During deformation, the form with $q_1 = - \frac{\pi}{4}$ rad, $q_2 = \frac{\pi}{2}$ rad (\figref{figure:indoor_hovering_transformation_kick}\textcircled{\scriptsize 2}) corresponds to the smallest $\tau_{\mathrm{min}}$ as shown in \figref{figure:feasible_control_convex_examples}(D), while the form with $q_1 =  q_2 = \frac{\pi}{4}$ rad (\figref{figure:indoor_hovering_transformation_kick}\textcircled{\scriptsize 3})  serves as the initial form to grasp an object. The overall tracking errors are relatively small as shown in \figref{figure:indoor_hovering_transformation_kick}(B); however, the deviation of the yaw motion (i.e., $\alpha_z$) can be confirmed under the form of \textcircled{\scriptsize 2} and \textcircled{\scriptsize 3}. We consider this is due to the model error caused by the slight bending and torsion of the multilinked structure. Nevertheless, the deviation was slowly reduced by the integral control in our proposed attitude control, and the stability regarding other axes was guaranteed regardless of the deformation. These results demonstrate the effectiveness of the proposed control method for the multilinked aerial robot with the tilting propeller. Furthermore, the robustness against the sudden impact was also evaluated as shown in \figref{figure:indoor_hovering_transformation_kick}\textcircled{\scriptsize 4}. The large external wrench caused by human kicks induced the temporal divergence in horizontal and yaw motion (41 s and 43 s in \figref{figure:indoor_hovering_transformation_kick}(B)); however, the hovering was recovered quickly, which showed the sufficient robustness of this platform.

\begin{figure}[!h]
  \begin{center}
    \includegraphics[width=0.93\columnwidth]{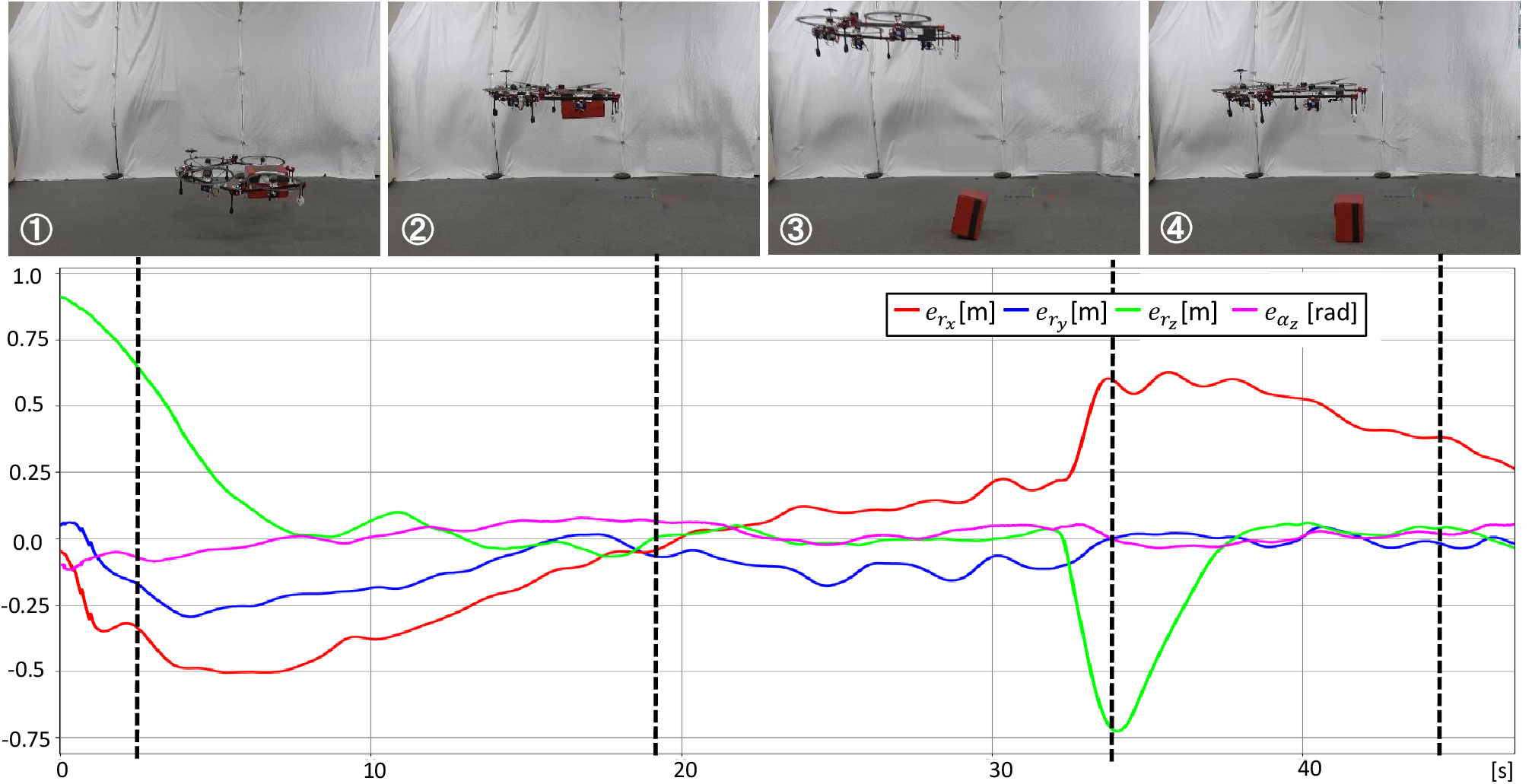}
    \caption{Grasping and dropping an object with unknown weight (the actual weight is 1.0 Kg) from takeoff phase. The tracking errors confirm the stable hovering, demonstrating the compensation ability of our control method regarding the unknown and large model error. } 
    \label{figure:indoor_grasping_takeoff_release}
  \end{center}
\end{figure}

\subsubsection{Grasping and Releasing Object}

Our multilinked robot  can be regarded as an entire gripper to grasp an object using two end links. In order to increase the contact area, an extended gripper tip attached on the propeller duct was designed as shown in \figref{figure:mbzirc_task2_sensors}(A).
Then an hovering experiment with a grasped object was conducted as shown in \figref{figure:indoor_grasping_takeoff_release}.
The weight of grasped object is 1.0Kg,  which was not contained in the dynamics model \equref{eq:translational_dynamics} and \equref{eq:rotational_dynamics2} for control system. Therefore, the temporal position divergence because of the deviation of the CoG frame occurred after takeoff as shown in \figref{figure:indoor_grasping_takeoff_release}\textcircled{\scriptsize 1}, which was recovered by the integral control as shown in \figref{figure:indoor_grasping_takeoff_release}\textcircled{\scriptsize 2}. Furthermore, a sudden ascending and deviation in x axis also occurred right after releasing the object as shown \figref{figure:indoor_grasping_takeoff_release}\textcircled{\scriptsize 3}. However, the robot was back to the desired point smoothly as shown \figref{figure:indoor_grasping_takeoff_release}\textcircled{\scriptsize 4}. Therefore, these result demonstrates the compensation ability of our control method regarding the unknown and large model error.

\subsubsection{Opening Sheet by Deformation}
\begin{figure}[!th]
  \begin{center}
    \includegraphics[width=0.95\columnwidth]{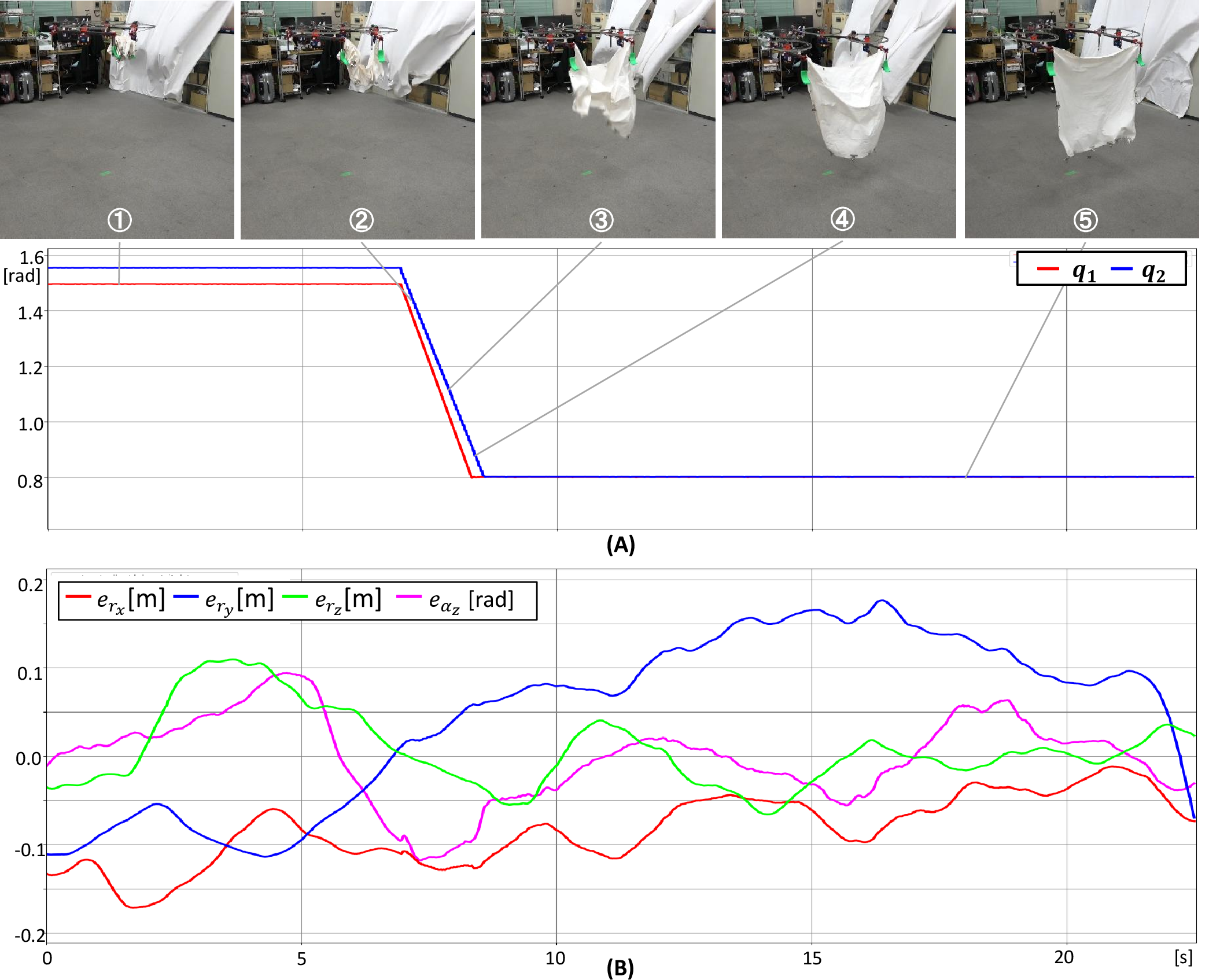}
    \caption{The stable sheet expanding motion by deformation (\textcircled{\scriptsize 1} $\sim$ \textcircled{\scriptsize 5}).
       {\bf (A)}: the change in joint angles during this experiment. {\bf (B)}: the tracking errors regarding the translational and yaw rotational motion.
      The video of comparison with a robot model without the tilting propeller can be found at https://youtu.be/LkDGP82sg1I.}
    \label{figure:indoor_sheet_openning}
  \end{center}
\end{figure}
The proposed multilinked structure can also be regarded as a manipulator to expand sheet, and one of the application using this motion is to cover a fire spot from the air which will be presented later in detail. However, the aerodynamics disturbance caused by the airflow acting on the expanded sheet surface can significantly prevent the hovering stability. In the case of the robot model without the tilting propeller, such disturbance can easily induce the yaw control divergence because of the low controllability. Then, the improvement of stability achieved by the robot model with proposed tilting propeller was evaluated  as shown in \figref{figure:indoor_sheet_openning}. The sheet of which the weight is 0.68 Kg and width is 1 m  was folded in the beginning as shown in \figref{figure:indoor_sheet_openning}\textcircled{\scriptsize 1}, and then fully expanded as shown in \figref{figure:indoor_sheet_openning}\textcircled{\scriptsize 2}$\sim$\textcircled{\scriptsize 5}. The tracking errors as shown in \figref{figure:indoor_sheet_openning}(B) demonstrate the hovering stability when sheet was fully expanded.
In addition to the aerodynamics disturbance,  the CoG of the sheet would suddenly change at the expanding moment. Nevertheless, the stable flight during \textcircled{\scriptsize 2}$\sim$\textcircled{\scriptsize 4}  was guaranteed, which also indicates the robustness against a large disturbance.

\subsection{Outdoor Experiments}

In the outdoor experiments, we first performed the fundamental hovering test with the onboards sensor to confirm the feasibility of the state estimation, which was followed by a circle trajectory tracking experiment to show the feasibility of the control method for a relatively fast and wide-range motion. Then, three task-specific experiments were conducted. Given than the main focus of this paper is the evaluation on the effectiveness of the proposed robot platform in fully autonomous flight involving aerial deformation, the specific target/environment detection method and motion planning algorithm for each task  will be not presented in this paper; however the task-specific sensors and the platform customization for each task  will be introduced in each experiment.

\subsubsection{Hovering and Circle Trajectory Tracking}
\begin{figure}[!h]
  \begin{center}
    \includegraphics[width=\columnwidth]{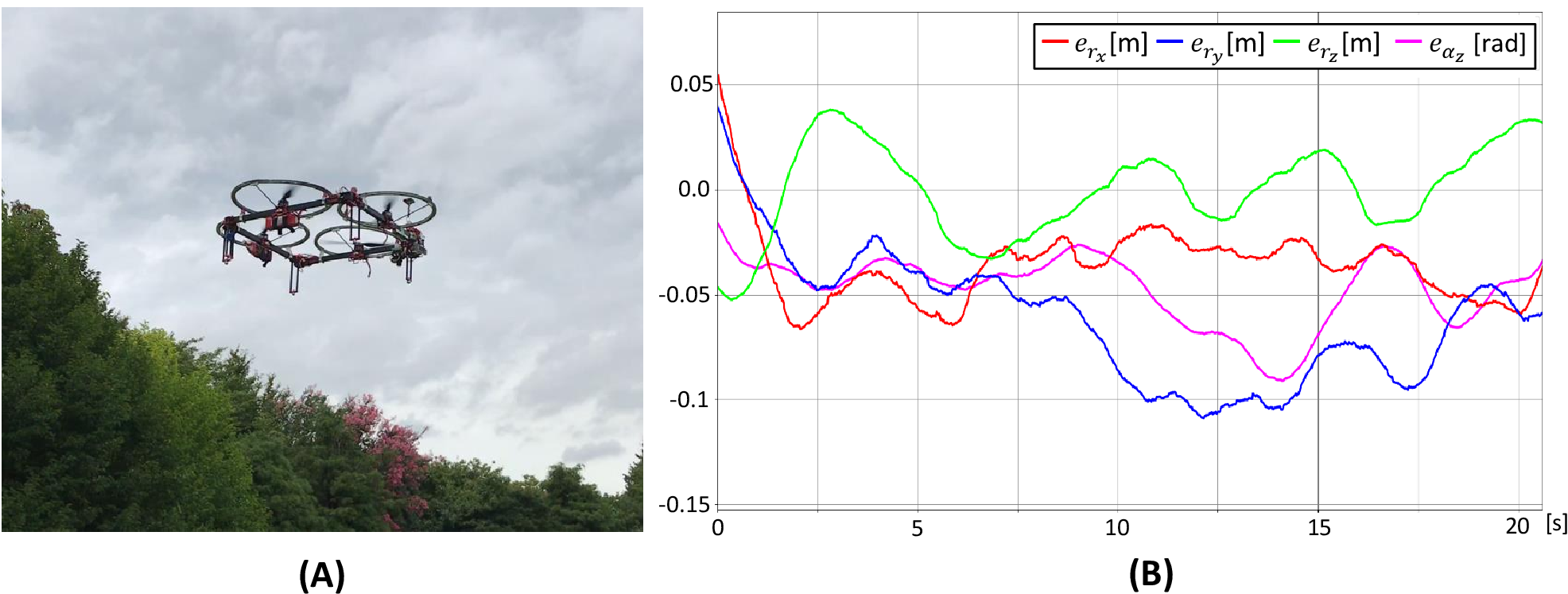}
    \caption{{\bf (A)} the hovering test at a height of 3 m in outdoor environment using the onboard sensors and developed state estimation method. {\bf (B)} the tracking errors regarding the translational and yaw rotational motion.}
    \label{figure:outdoor_hovering}
  \end{center}
\end{figure}

The autonomous hovering at a height of 3 m was first conducted as shown in \figref{figure:outdoor_hovering}(A). The relatively small tracking errors as shown in \figref{figure:outdoor_hovering}(B) demonstrate the stable flight  around a desired  point. Although there was no ground truth provided in this experiment to evaluate the accuracy of the proposed state estimation, the convergence of flight control can indirectly confirm the performance of state estimation, since the uncertainty of state estimation would induce  the divergence of flight control. However, a relatively constant offset in $x$ and $y$ axes might always exist, since the global position in $x$ and $y$ axes for the state estimation framework  is only available from GPS sensor (${}^{\{W\}}\bm{p}_{\{GPS\}_{xy}}$  in \figref{figure:state-estimation}) which generally contains a certain offset in positioning. Nevertheless, most of the outdoor application involves visual servoing which can guarantee the expected tracking performance towards a desired position.

\begin{figure}[!h]
  \begin{center}
    \includegraphics[width=\columnwidth]{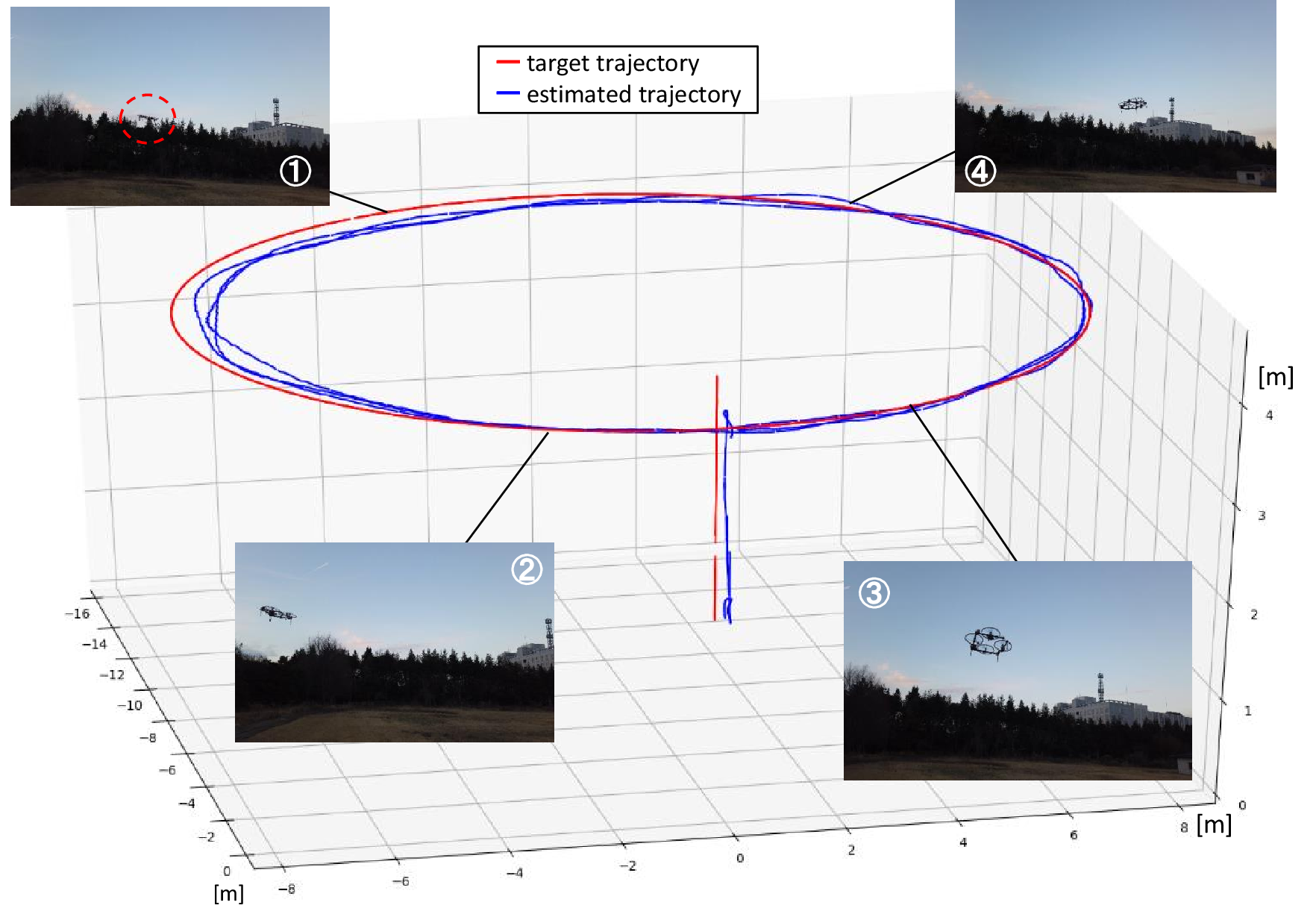}
    \caption{The tracking performance in outdoor environment which tracked a circle trajectory of which the radius is 8m and the height is 4m.  The tracking desired velocity was gradually increased from 0.5 m/s to 3.0 m/s over three laps.}
    \label{figure:outdoor_circle_tracking}
  \end{center}
\end{figure}

\begin{figure}[!h]
  \begin{center}
    \includegraphics[width=0.95\columnwidth]{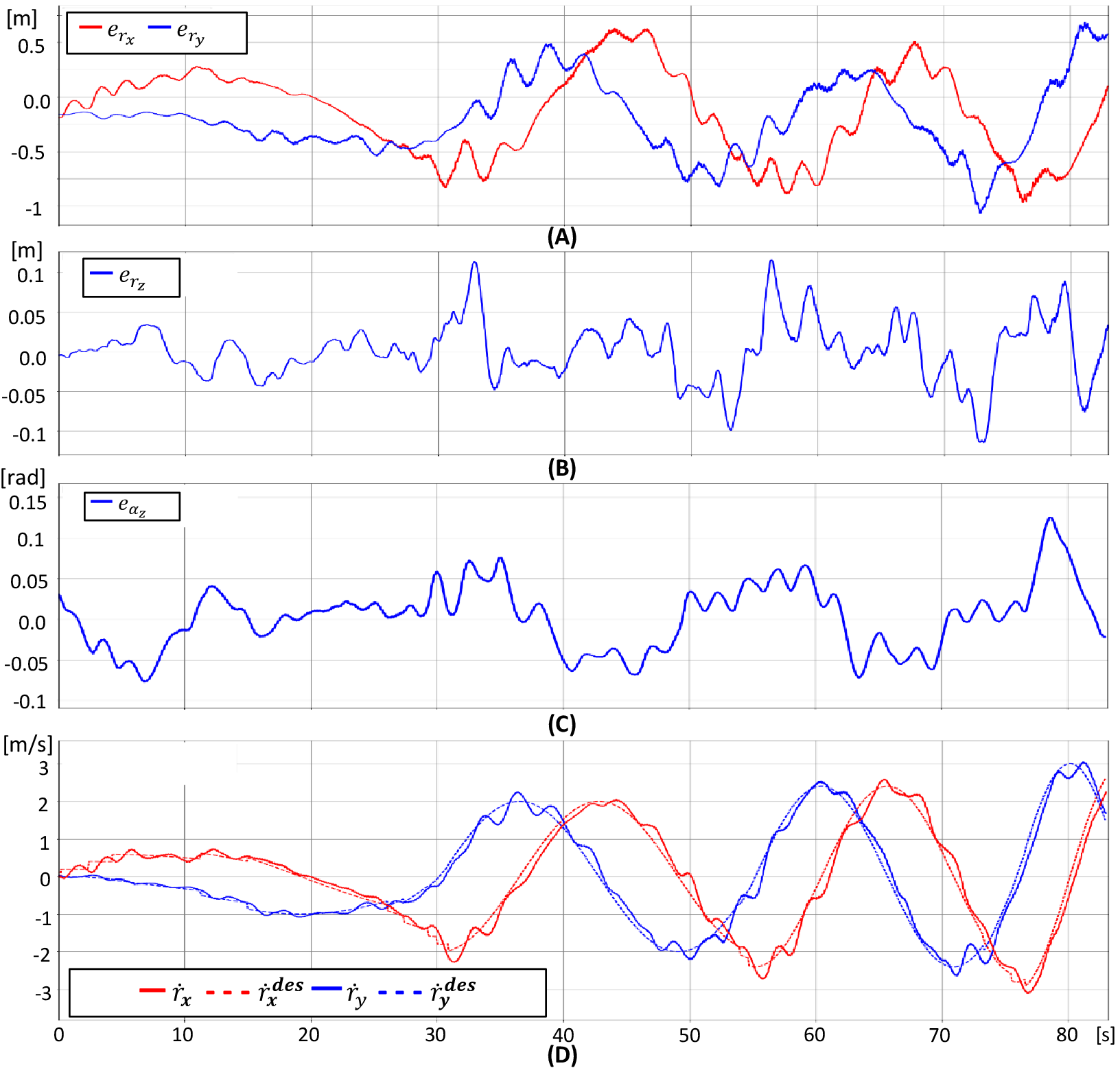}
    \caption{{\bf (A)} $\sim$ {\bf (C)}: the tracking errors (i.e., $e_{r_x}, e_{r_y}, e_{r_z}, e_{\alpha_z}$) regarding experiment as shown in \figref{figure:outdoor_circle_tracking}.  {\bf (D)}: the velocity tracking performance. The desired velocity was gradually increased from 0.5 m/s to 3.0 m/s.}
    \vspace{-4mm}
    \label{figure:outdoor_circle_tracking_err_plot}
  \end{center}
\end{figure}

Then, the feasibility of the control system to track a relatively wide and aggressive trajectory was evaluated.  The desired trajectory is a circle of which the radius is 8 m and the height is 4 m  as shown in \figref{figure:outdoor_circle_tracking}, and the desired tracking velocity was gradually increased from 0.5 m/s to 3.0 m/s over 3 laps. As shown in \figref{figure:outdoor_circle_tracking_err_plot}(A), the maximum horizontal error reached 1m at a desired tracking velocity of 3.0m.
Such a large deviation is due to the insufficient proportional control in \equref{eq:pid_pos}  when the motion becomes aggressive. However, increasing $K_P$ in \equref{eq:pid_pos} might induce unexpected vibration in hovering flight.  Thus, in order to guarantee the tracking performance, the maximum horizontal velocity is limited to 2.0m/s in most of tasks.
In comparison with the position tracking, the velocity tracking performance demonstrated a better result as shown in \figref{figure:outdoor_circle_tracking_err_plot}(D), implying the potential to perform an aggressive maneuvering.
On the other hand, the expected tracking performance on z and yaw motion around the fixed desired values (i.e., $r^{\mathrm{des}}_z = 4$ m, $\alpha^{\mathrm{des}}_z = 0$ rad) can be confirmed from \figref{figure:outdoor_circle_tracking_err_plot}(B) and (C).

\subsubsection{Interception with a Fast Flying Target}

\begin{figure}[!t]
  \begin{center}
    \includegraphics[width=0.9\columnwidth]{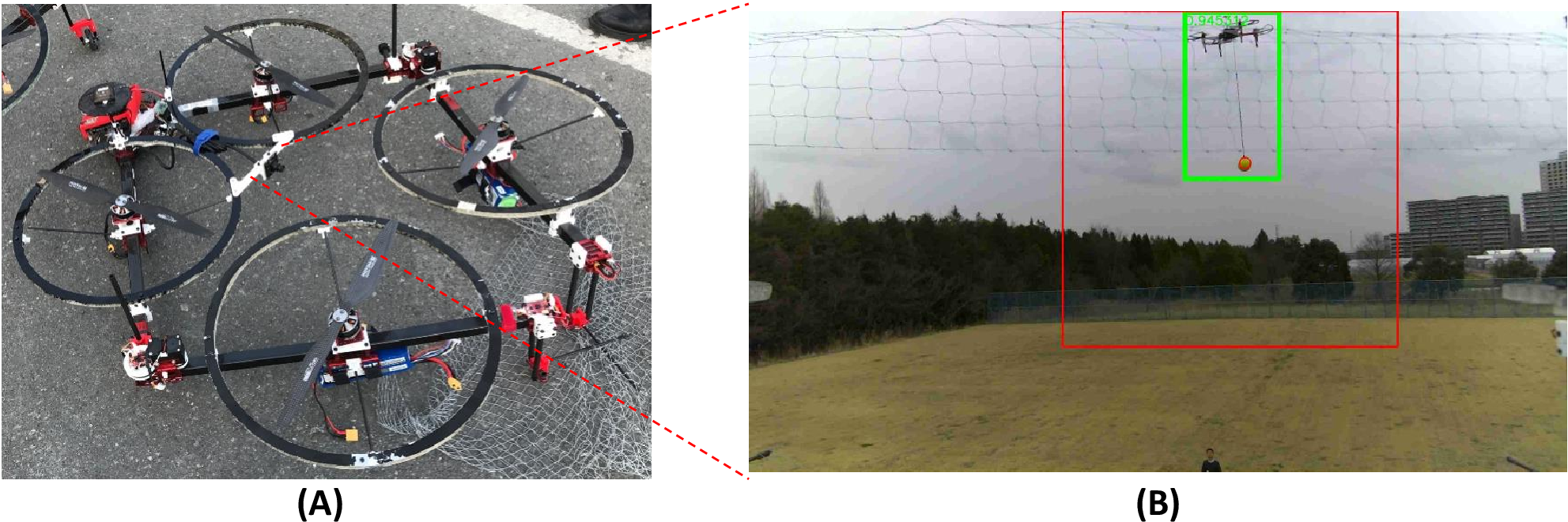}
    \vspace{-3mm}
    \caption{{\bf (A)} the front-facing camera for the moving target detection and the net between two ends to perform dropping and catching. {\bf (B)} the image detection on the moving target (a moving quadrotor with a hung yellow ball) by using the edge computing device Google Coral. }
    \label{figure:mbzirc_task1_sensors}
  \end{center}
\end{figure}

In order to evaluate the performance of our developed platform on an aggressive task, an experiment to intercept a fast flying target was conducted. Regarding the task-specific sensor, a front-facing monocular camera (ELP-SUSB1080P01-LC1100\footnote[14]{http://www.webcamerausb.com/}) was equipped to detect the moving target as shown in \figref{figure:mbzirc_task1_sensors}(A), and a edge computing device (Google Coral\footnote[15]{https://coral.ai/products/accelerator}) is connected to the onboard computer (Intel NUC7I7DNHE\footnote[16]{https://www.intel.com/content/www/us/en/products/boards-kits/nuc/kits/nuc7i7dnhe.html}) to perform SSD detection \cite{SSD} as shown in \figref{figure:mbzirc_task1_sensors}(B). On the other hand, a front net between two ends was also equipped to enable dropping or catching target.

\begin{figure}[!h]
  \begin{center}
    \includegraphics[width=0.9\columnwidth]{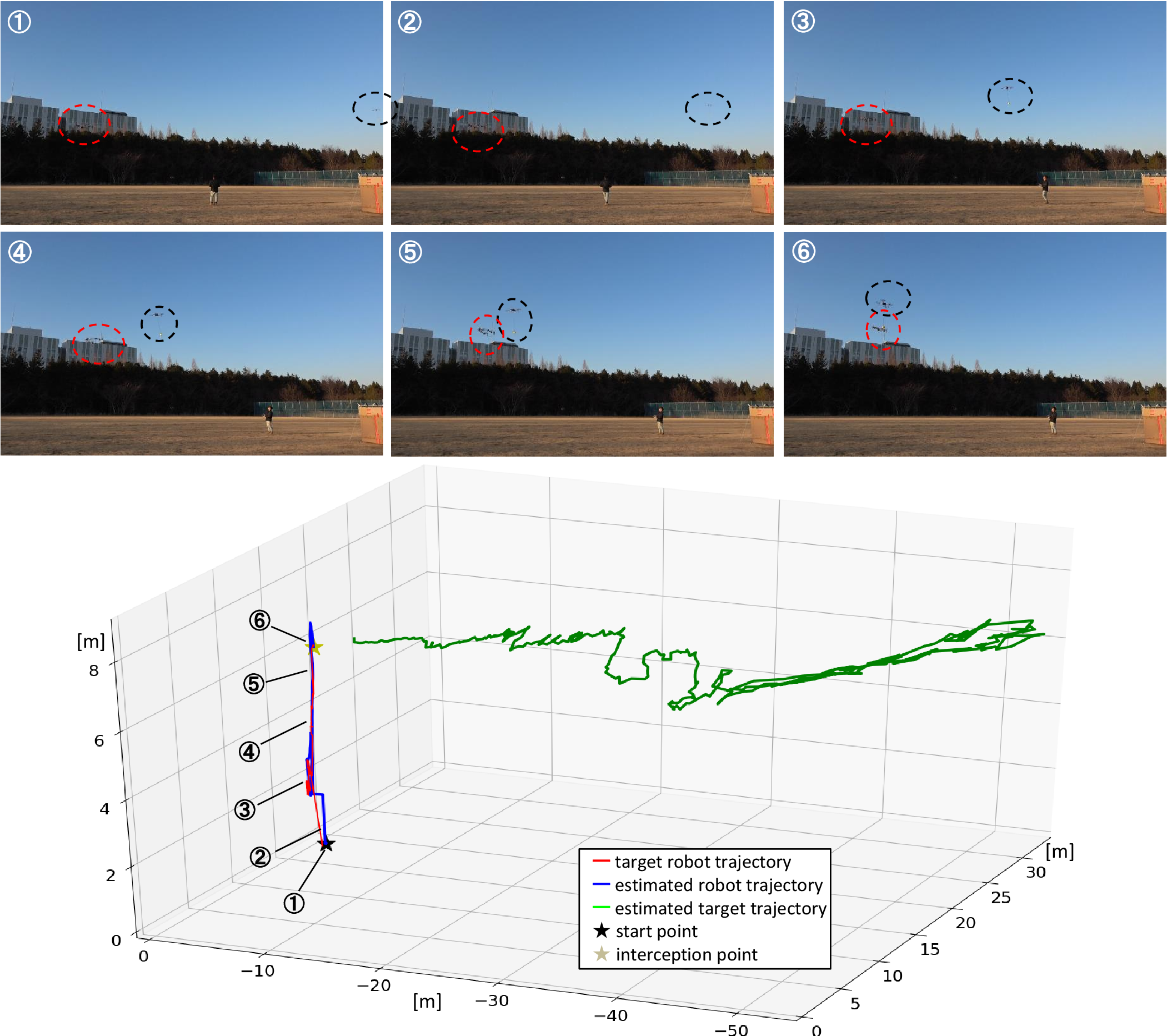}
    \caption{The trajectory of our robot  along with the estimated trajectory of the target in the autonomous interception task as shown in \textcircled{\scriptsize 1}$\sim$\textcircled{\scriptsize 6}. Our robot increased the height from 3 m to 8 m, and succeeded to intercept the target. The duration from starting ascending to interception was 6 s.}
    \label{figure:mbzirc_task1_plot}
  \end{center}
\end{figure}

In this experiment, the target is a yellow ball hung from a quadrotor. As shown \figref{figure:mbzirc_task1_plot}, the moving trajectory of the this quadrotor is straight line when viewed from above, and the flight height changes between 6 m to 10 m. The moving speed is 5 m/s. On the other hand, our robot was waiting at a height of 3 m at the beginning. Once the target was detected with a certain duration, the desired interception point could be predicted based on the estimated target trajectory. Subsequently, the desired interception motion for the robot was planned and further fulfilled.
As shown in \figref{figure:mbzirc_task1_plot}\textcircled{\scriptsize 3}$\sim$\textcircled{\scriptsize 6} and the plotted trajectory, the robot was ascending quickly to reach the same  height with the yellow ball, and an adjustment in horizontal motion was also performed simultaneously. Finally our robot  succeeded to hit the yellow ball  by the net and drop it from the moving quadrotor, which took 6 s from starting ascending to the interception. The accurate of the target detection and position projection was relatively low leading to the unreliable estimated trajectory when the target was far from the robot (the green trajectory in \figref{figure:mbzirc_task1_plot}). However, the estimation  of the target height was relatively reliable, which enables early ascending and thus promises the good visual servoing  when the target becomes closer.
The success of interception as shown in \figref{figure:mbzirc_task1_plot}  confirmed the feasibility of the proposed control system and state estimation to perform an aggressive task involving fast ascending and quick horizontal motion.

\subsubsection{Searching, Grasping and Delivering a Brick}
\begin{figure}[!th]
  \begin{center}
    \includegraphics[width=\columnwidth]{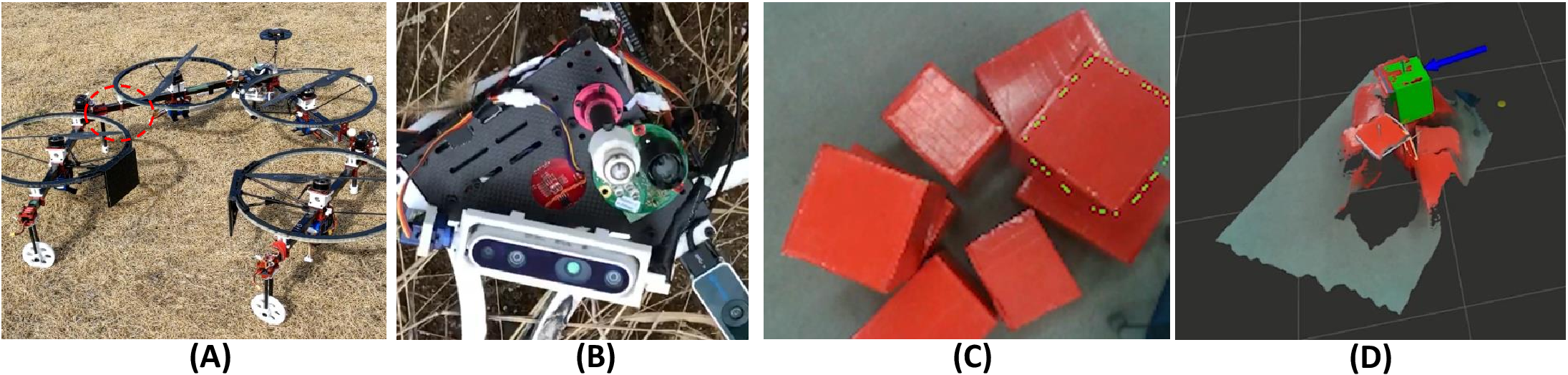}
    \vspace{-6mm}
    \caption{{\bf (A)}: the customized platform which can grasp object by using two black  tips attached on the propeller duct. {\bf (B)}: the downward RGBD sensor (RealSense D435i) for the ground object detection. {\bf (C)}: the downward image from the sensor.  {\bf (D)}: the downward point cloud from the sensor, where the detected object is highlighted by green color.}
    \label{figure:mbzirc_task2_sensors}
    \vspace{-5mm}
  \end{center}
\end{figure}

One of the novel characteristics of our platform is the manipulation ability. Then a task to grasp object from the ground and subsequently deliver to a designated location was performed. The grasping strategy was the same as shown in \figref{figure:indoor_grasping_takeoff_release}, which used two ends to contact with the object surfaces. A RBGD sensor (RealSense D435i\footnote[17]{https://www.intelrealsense.com/depth-camera-d435i/}) was equipped as shown in \figref{figure:mbzirc_task2_sensors}(B) to detect the ground object from both  color image and point cloud as shown in \figref{figure:mbzirc_task2_sensors}(C) and (D). The onboard processor to process the image and point cloud is LattePanda Alpha 864s\footnote[18]{https://www.lattepanda.com/products/lattepanda-alpha-864s.html}. 
\begin{figure}[!h]
  \begin{center}
    \includegraphics[width=0.95\columnwidth]{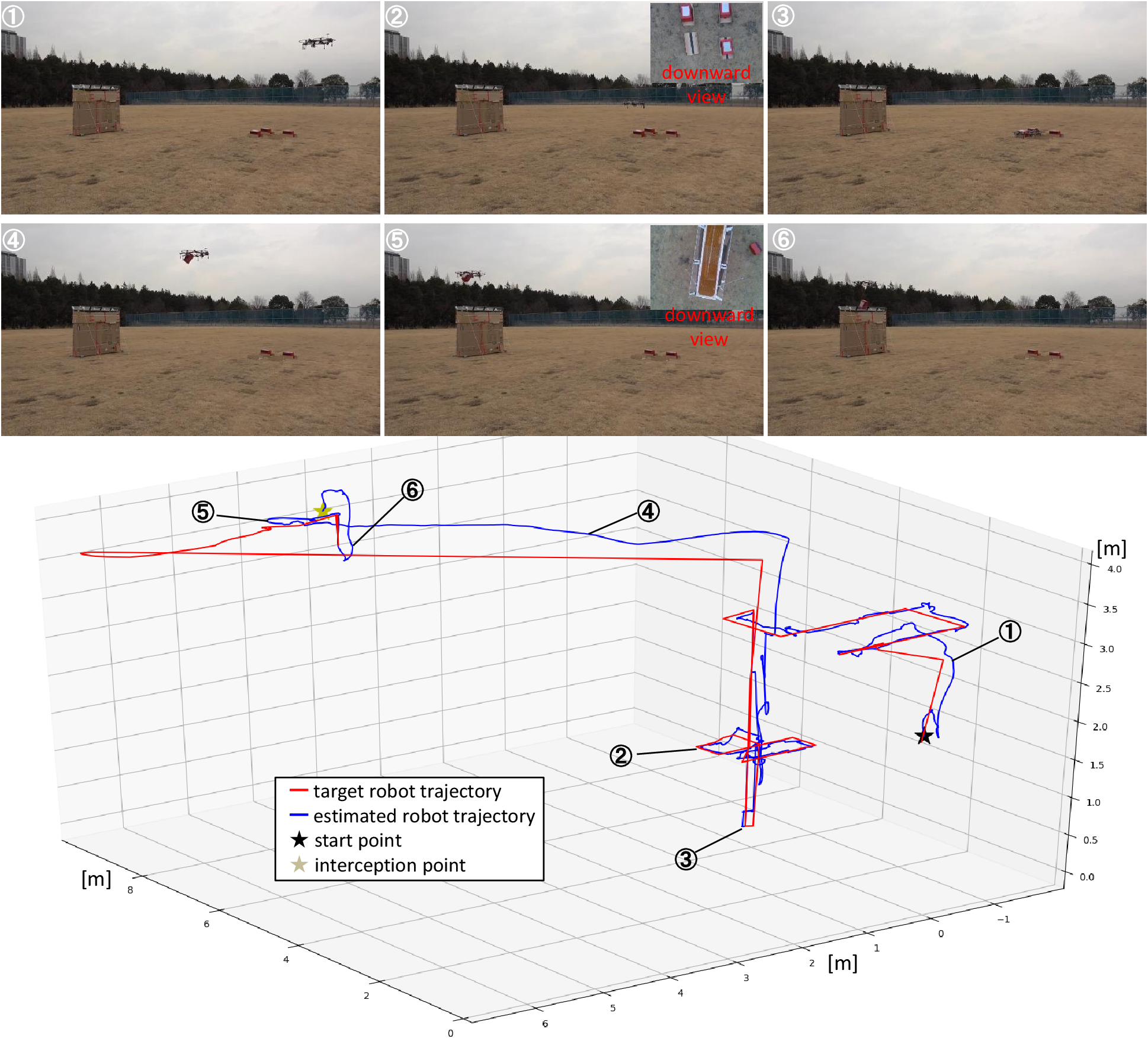}
    \vspace{-4mm}
    \caption{The trajectory of our robot in the autonomous task involving brick searching, grasping, delivering and placing  as shown in \textcircled{\scriptsize 1}$\sim$\textcircled{\scriptsize 6}. Our robot increased the height from 3 m to 8 m, and succeeded to intercept the target. The duration from starting ascending to interception was 6 s. The robot succeeded to grasp and place a brick with a weight 0f 1.0 Kg to a wall with a height of 2 m.}
    \label{figure:mbzirc_task2_plot}
  \end{center}
  \vspace{-5mm}
\end{figure}

In this autonomous task, the robot was first required to find the ground red bricks with a weight of 1.0 Kg by moving to several designated waypoints as shown in \figref{figure:mbzirc_task2_plot}\textcircled{\scriptsize 1}$\sim$ \textcircled{\scriptsize 2}. Once the bricks  were detected and  target brick was selected, the robot fully landed to grasp the target brick as shown in \figref{figure:mbzirc_task2_plot}\textcircled{\scriptsize 3}. Subsequently, the robot took off again to deliver the brick to the wall as shown in \figref{figure:mbzirc_task2_plot}\textcircled{\scriptsize 4}.
A relatively large tracking error can be confirmed from the deviation between the blue and red trajectories as shown in \figref{figure:mbzirc_task2_plot} during this delivery phase. This is because the control system was switched to a velocity control mode (i.e., $K_P = 0$ in \equref{eq:pid_pos}), and thus the position error was allows in this phase.  Nevertheless, the robot converged to a desired position once the control system was switched back to the potions control model as shown in  \figref{figure:mbzirc_task2_plot}\textcircled{\scriptsize 5}.
Finally, the robot placed the brick on the top of the wall with a height of 2 m (\figref{figure:mbzirc_task2_plot}\textcircled{\scriptsize 6})  by performing the channel detection using the downward RGBD sensor. A heuristic solution to switch off the downward LiDAR while flying upon the wall was applied to solve the height gap problem in height estimation.
The success of the whole task showed the feasibility of the developed platform to autonomously gasp, deliver object, along with the effectiveness of state estimation and the flight control in a task involving landing and takeoff on the way.

\subsubsection{Firefighting by Using Blanket}
\begin{figure}[!h]
  \begin{center}
    \includegraphics[width=0.95\columnwidth]{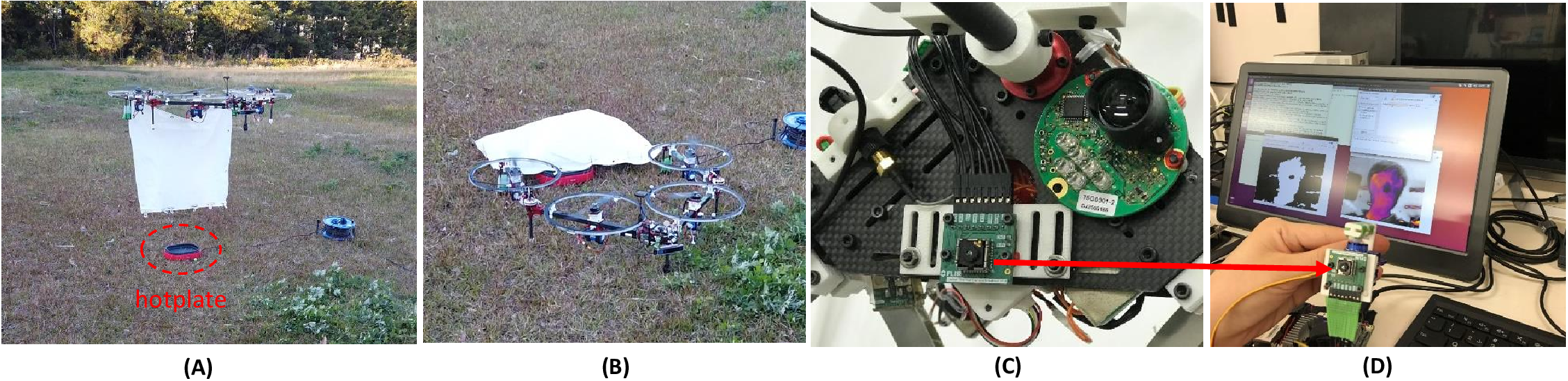}
    \vspace{-4mm}
    \caption{{\bf (A) - (B)}: firefighting with a blanket expanded by our robot. The hotplate is regarded as a virtual fire spot. {\bf (C) - (D)} the downward thermal sensor (FLIR Radiometric Lepton Dev Kit) for the ground heat source detection.}
    \label{figure:mbzirc_task3_sensors}
    \vspace{-4mm}
  \end{center}
\end{figure}

In comparison to water, using a blanket does not require accurate shooting to distinguish fire, and our platform can expand a blanket by openning joints as shown in \figref{figure:indoor_sheet_openning}.  A hotplate was prepared to serve as a fire spot as shown in \figref{figure:mbzirc_task3_sensors}(A) and it is assumed that this virtual fire can be extinguished by the covering motion as shown in \figref{figure:mbzirc_task3_sensors}(B). To detect the heat from the hotplate, a downward thermal sensor (FLIR Radiometric Lepton Dev Kit\footnote[19]{https://www.sparkfun.com/products/15948}) was equipped  as shown in \figref{figure:mbzirc_task3_sensors}(C) and (D). In this task, the image processing cost is relatively low, thus the default onboard computer, UP Board$^6$ was selected.

\begin{figure}[!h]
  \begin{center}
    \includegraphics[width=0.91\columnwidth]{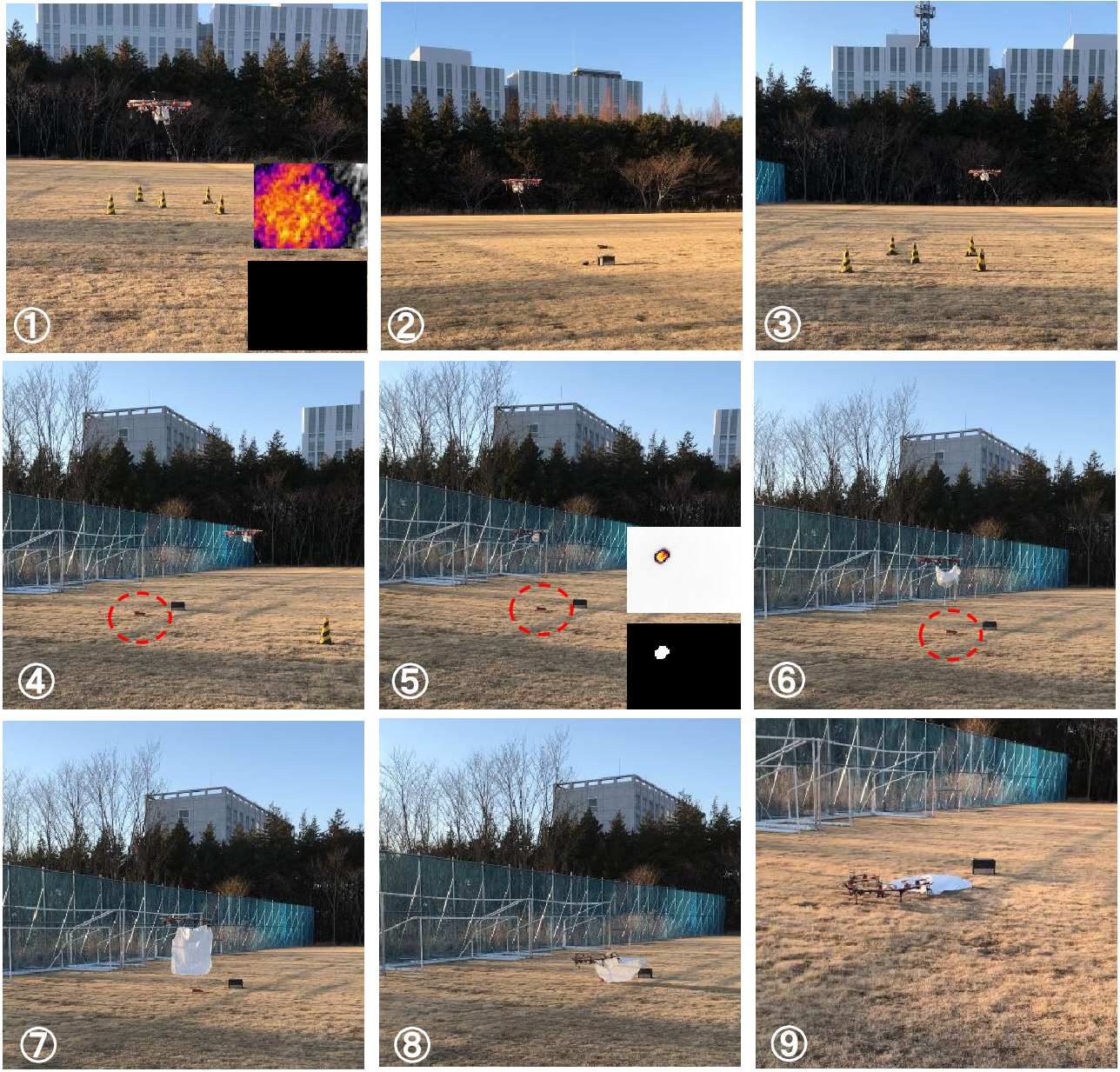}
    \caption{The autonomous task to find a heat source and perform firefighting motion by covering a blanket. The sub images in \textcircled{\scriptsize 1} and \textcircled{\scriptsize 5} correspond to the colored thermal images and the thresholded images by temperature. The hotplate surrounded in a dashed circle was found at \textcircled{\scriptsize 5}, and then the covering motion was performed \textcircled{\scriptsize 6} $\sim$ \textcircled{\scriptsize 9}.}
    \label{figure:mbzirc_task3_image}
  \end{center}
\end{figure}

\begin{figure}[!h]
  \begin{center}
    \includegraphics[width=\columnwidth]{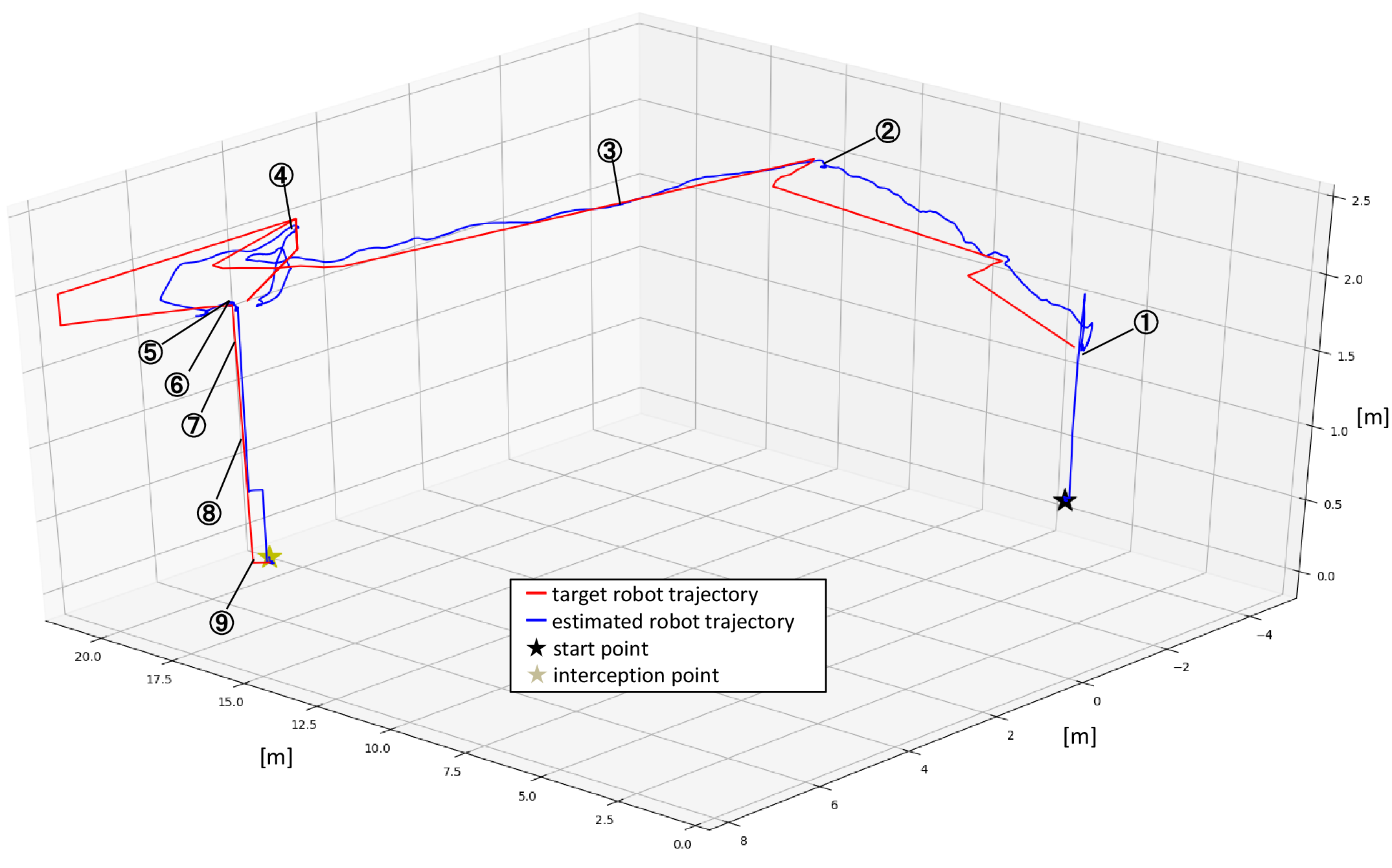}
    \vspace{-4mm}
    \caption{The trajectory of our robot related to the autonomous firefighting task as shown in \figref{figure:mbzirc_task3_image}.}
    \label{figure:mbzirc_task3_plot}
  \end{center}
\end{figure}

Similar to the object grasping and deliver task as shown in \figref{figure:mbzirc_task2_plot}, the firefighting task also started with an autonomous search by patrolling between several designated waypoints as shown in \figref{figure:mbzirc_task3_image}\textcircled{\scriptsize 1}$\sim$ \textcircled{\scriptsize 4}.
Again, a relatively large tracking error can be confirmed from the blue and red trajectories in \figref{figure:mbzirc_task3_plot} during this patrol phase, which is due to the same reason as explained in the second task.
Once the heat source was found by the thermal sensor as shown in the sub-images inside \figref{figure:mbzirc_task3_image}\textcircled{\scriptsize 5}, the expanding motion followed by a rapid descending was performed as shown in \figref{figure:mbzirc_task3_image}\textcircled{\scriptsize 6}$\sim$ \textcircled{\scriptsize 9}.
The success of the whole task demonstrated the feasibility of the developed platform to autonomous patrol in a relatively wide area (i.e. 12 m $\times$ 20 m as shown in \figref{figure:mbzirc_task3_plot}).
Moreover, the proposed state estimation by fusing multiple sensors guaranteed the estimation quality and also the flight stability  even after the blanket was expanded and thus large occlusion area occurred which decreased the accuracy of the output from VIO module.


\subsection{Result in MBZIRC 2020}
At the competition, we participated in all challenges with  multilinked aerial robots in fully autonomous mode as shown in \figref{figure:abst}(C)$\sim$(E), and the platforms were customized as shown in \figref{figure:mbzirc_task1_sensors}, \figref{figure:mbzirc_task2_sensors} and \figref{figure:mbzirc_task3_sensors} for each challenge, respectively.

The performance in the competition was not as ideal or as good as our experiments because we did not have enough time to debug and fine-tune our system by the time of the competition. But nonetheless, we were still able to rank third place in Challenge 1 (one of the three teams to succeed to intercept the moving target) and sixth place in Challenge 3. Regarding Challenge 2, we succeeded to grasp, deliver and place a brick to the goal in rehearsal, which would deserve  a high score in the real challenge. Those results sufficiently demonstrated not only the feasibility of the proposed methods for desgin, control and state estimation to achieve the fully autonomous maneuvering, but also the versatility of our proposed deformable aerial robot in  different tasks.

\section{Conclusion and Lessons Learned}
\label{sec:conclusion}
In this paper, we presented a multilinked aerial robot platform comprised from two joints and four tilting rotor. The design, modeling and control method has been developed to achieve the flight stability, while the state estimation based on the multilinked kinematics has been also developed  to enable the fully autonomous flight in the fields. Various on-site experiments, including the fast maneuvering for target interception, the aerial grasping for delivery and the blanket manipulation for firefighting, has been evaluated to demonstrate the feasibility of the proposed platform.

This robot platform has been also evaluated in MBZIRC 2020 which is a highly demanding and successful event for us. An on-site testbed similar to the real competition also helped us to properly improve our platform from daily experiments.  Furthermore, unification of the platform foundation in all challenges and suppression of task-specific customization are the strategic advantages in our development, which leads to the synergy between different tasks such as the efficiency of the robot operation and  maintenance.  In addition to the versatility, our robot platform also demonstrates the novelty in aerial manipulation, which can be considered as an important contribution to the field robotics community.

Several open issues remain to be address in future work. First, an improved position control method should be developed  to enable more aggressive maneuvering.
Second, the measurement delay of a sensor which is affected by the whether and location was manually tuned by human in this work. Hence, a temporal calibration method should be developed to autonomously identify the measurement delay, and further improve the accuracy of the state estimation. Last but not least, a cooperative control system should be investigated to achieve the manipulation or transportation of a large object by using multiple robots.

\section*{Appendix A: Postion Error Dynamics}
The position error dynamics is given by:
\begin{eqnarray}
  \label{eq:pos-err-dynamics}
  m \ddot{\bm{e}}_{r}
  &=& m\ddot{\bm r} - m\ddot{\bm r}^{\mathrm{des}} \nonumber \\
  &=& m\bm{g} + R (f_T^{\mathrm{des}} \bm{b}_3 + Q_{\mathrm{tran}} \bm{u}^{\mathrm{des}}_{\mathrm{att}}) - m\ddot{\bm r}^{\mathrm{des}} + \bm{\Delta}_{\mathrm{tran}} \nonumber \\
  &=& m\bm{g} +  ((R \bm{b}_3 )^{\mathrm{T}}  {\bm f}^{\mathrm{des}})R \bm{b}_3 + R Q_{\mathrm{tran}} \bm{u}^{\mathrm{des}}_{\mathrm{att}} - m\ddot{\bm r}^{\mathrm{des}} + \bm{\Delta}_{\mathrm{tran}} \nonumber \\
  &=& m\bm{g} +  \|{\bm f}^{\mathrm{des}}\| ((R^{\mathrm{des}} \bm{b}_3)^{\mathrm{T}}  R  \bm{b}_3 )R \bm{b}_3 + R Q_{\mathrm{tran}} \bm{u}^{\mathrm{des}}_{\mathrm{att}} - m\ddot{\bm r}^{\mathrm{des}} + \bm{\Delta}_{\mathrm{tran}} \nonumber \\
  &=& m\bm{g} +  \|{\bm f}^{\mathrm{des}}\| R^{\mathrm{des}} \bm{b}_3 - \|{\bm f}^{\mathrm{des}}\| R^{\mathrm{des}} \bm{b}_3 +  \|{\bm f}^{\mathrm{des}}\| ((R^{\mathrm{des}} \bm{b}_3)^{\mathrm{T}}  R  \bm{b}_3 )R \bm{b}_3 + R Q_{\mathrm{tran}} \bm{u}^{\mathrm{des}}_{\mathrm{att}} - m\ddot{\bm r}^{\mathrm{des}} + \bm{\Delta}_{\mathrm{tran}} \nonumber \\
  &=& m\bm{g} +  \|{\bm f}^{\mathrm{des}}\| R^{\mathrm{des}} \bm{b}_3 + \|{\bm f}^{\mathrm{des}}\| ( ((R^{\mathrm{des}} \bm{b}_3)^{\mathrm{T}}  R  \bm{b}_3 )R \bm{b}_3  -  R^{\mathrm{des}} \bm{b}_3 ) + R Q_{\mathrm{tran}} \bm{u}^{\mathrm{des}}_{\mathrm{att}} - m\ddot{\bm r}^{\mathrm{des}} + \bm{\Delta}_{\mathrm{tran}} \nonumber \\
  &=& m\bm{g} +  \|{\bm f}^{\mathrm{des}}\| R^{\mathrm{des}} \bm{b}_3 + \bm{X} + R Q_{\mathrm{tran}} \bm{u}^{\mathrm{des}}_{\mathrm{att}} - m\ddot{\bm r}^{\mathrm{des}} + \bm{\Delta}_{\mathrm{tran}}
\end{eqnarray}
where $ \bm{X} = \|{\bm f}^{\mathrm{des}}\| ( ((R^{\mathrm{des}} \bm{b}_3)^{\mathrm{T}}  R  \bm{b}_3 )R \bm{b}_3  -  R^{\mathrm{des}} \bm{b}_3 )$. Note that, ${}^{\{W\}}R_{\{CoG\}}$ and  $m_{\Sigma}$  are simplified as $R$ and $m$ for convenience.

Substituting  \equref{eq:pid_pos} into \equref{eq:pos-err-dynamics}, the further derivation can be given by
\begin{align}
  m \ddot{\bm{e}}_{r} &= - m K_P \bm{e}_r - m K_I \bm{e}_{I_x} - m K_D \dot{\bm{e}}_r  + \bm{X} + R Q_{\mathrm{tran}} K_x\bar{\bm {x}} + \bm{\Delta}_{\mathrm{tran}} \nonumber \\
  &= - m K_P \bm{e}_r - m K_I \bm{e}_{I_r} - m K_D \dot{\bm{e}}_r + \bm{X} + R Q^{'}_{\mathrm{tran}} \bm {e}_{\mathrm{rot}}  + \bm{\Delta}_{\mathrm{tran}} + \bm{\Delta}^{'}_{\mathrm{rot}}   \tag{\ref{eq:pos-err-dynamics2}}
\end{align}
where  $\bm{\Delta}^{'}_{\mathrm{rot}} =  Q_{\mathrm{tran}} Q^{\#}_{\mathrm{rot}} \bm{\Delta}_{\mathrm{rot}}$ and $Q^{'}_{\mathrm{tran}} = Q_{\mathrm{tran}} K_x$. Note that $R$ is omitted from $\bm{\Delta}^{'}_{\mathrm{rot}}$,  since such converted fixed uncertainty from rotational space should be also constant in translational space regardless of the change of $R$. For instance,  the offset of the CoG origin which induces a moment in rotational motion should be converted to be a constant force in translational force.

\section*{Appendix B: Complete Stability}
As shown in \equref{eq:pos-lyapunov}, the  Lyapunov candidate  regarding the position error dynamics is given by
\begin{align}
  \mathcal{V}_2 = \frac{1}{2} \bm{e}_{r}^{\mathrm{T}} K_P \bm{e}_{r} +  \frac{1}{2} \|\dot{\bm{e}}_{r}\|^2 + c \bm{e}^{\mathrm{T}}_{r} \dot{\bm{e}}_{r} + \frac{1}{2} (\bm{e}_{I_r} - \bm{\Delta} \circ \frac{1} {\bm{k}_{I_r}})^{\mathrm{T}} K_{I_r} (\bm{e}_{I_r} - \bm{\Delta} \circ \frac{1} {\bm{k}_{I_r}})   \tag{\ref{eq:pos-lyapunov}}
\end{align}
Then, the time derivative can be given by
\begin{eqnarray}
  \label{eq:pos-dot-lyapunov}
  \dot{\mathcal{V}}_2 &=&  \bm{e}_{r}^{\mathrm{T}} K_P \dot{\bm{e}}_{r} +  \dot{\bm{e}}_{r}^{\mathrm{T}} \ddot{\bm{e}}_{r}  + c \dot{\bm{e}}^{\mathrm{T}}_{r} \dot{\bm{e}}_{r} + c \bm{e}^{\mathrm{T}}_{r} \ddot{\bm{e}}_{r} +  \dot{\bm{e}}^{\mathrm{T}}_{I} K_{I_r}  (\bm{e}_{I} - \bm{\Delta} \circ \frac{1} {\bm{k}_{I_r}})  \nonumber \nonumber \\
  &=&  c \|\dot{\bm{e}}^{\mathrm{T}}_{r}\|^2 + \bm{e}_{r}^{\mathrm{T}} K_P \dot{\bm{e}}_{r} +  (\dot{\bm{e}}_{r}^{\mathrm{T}}  + c \bm{e}^{\mathrm{T}}_{r} ) \ddot{\bm{e}}_{r} + (\dot{\bm{e}}^{\mathrm{T}}_r +  c\bm{e}^{\mathrm{T}}_{r} )  (K_{I_r} \bm{e}_{I_r} - \bm{\Delta}) \nonumber  \\
  &=&  c \|\dot{\bm{e}}^{\mathrm{T}}_{r}\|^2 + \bm{e}_{r}^{\mathrm{T}} K_P \dot{\bm{e}}_{r}
  + (\dot{\bm{e}}_{r}^{\mathrm{T}}  +   c \bm{e}^{\mathrm{T}}_{r}) (- K_P  \bm{e}_r - K_{I_r} \bm{e}_{I_r} - K_D \dot{\bm{e}}_r + \frac{\bm{X} + R Q^{'}_{\mathrm{tran}} \bm {e}_{\mathrm{rot}}}{m} + \bm{\Delta}   + K_{I_r} \bm{e}_{I_r} - \bm{\Delta})  \nonumber \\
  &=&  - c \bm{e}^{\mathrm{T}}_{r} K_P \bm{e}_{r} +   \dot{\bm{e}}^{\mathrm{T}}_{r} (C - K_D) \dot{\bm{e}} - c \bm{e}^{\mathrm{T}}_{r}  K_D \dot{\bm{e}}_{r} + (\dot{\bm{e}}_{r}^{\mathrm{T}}  + c \bm{e}^{\mathrm{T}}_{r}) \frac{\bm{X} + R Q^{'}_{\mathrm{tran}} \bm {e}_{\mathrm{rot}}}{m}
\end{eqnarray}
where,  $C = c E_{3 \times 3}$.

Then the upper bound of $\dot{\mathcal{V}}_2$ can be given by:
\begin{eqnarray}
  \label{eq:pos-dot-lyapunov2}
  \dot{\mathcal{V}}_2  &\leq& - c k_{P_{\mathrm{min}}} \|\bm{e}_{r}\|^2 +  (c - k_{D_{\mathrm{min}}}) \|\dot{\bm{e}}_{r}\|^2 + c k_{D_{\mathrm{max}}} \|\bm{e}_{r} \| \| \dot{\bm{e}}_{r} \| + (\|\dot{\bm{e}}_{r}\|  + c \|\bm{e}_{r}\|) \frac{\|\bm{X}\| + \|R Q^{'}_{\mathrm{tran}} \bm {e}_{\mathrm{rot}}\|}{m}   \nonumber \\
  &\leq& - ck_{P_{\mathrm{min}}} \|\bm{e}_{r}\|^2 +  (c - k_{D_{\mathrm{min}}}) \|\dot{\bm{e}}_{r}\|^2 + c k_{D_{\mathrm{max}}} \|\bm{e}_{r} \| \| \dot{\bm{e}}_{r} \| + (\|\dot{\bm{e}}_{r}\|  + c \|\bm{e}_{r}\|) \frac{\|\bm{X}\| + \sigma_{max}(Q^{'}_{\mathrm{tran}} ) \|\bm {e}_{\mathrm{rot}}\|}{m}, \nonumber \\
\end{eqnarray}
where, $k_{P_{\mathrm{min}}}$ and $k_{D_{\mathrm{min}}}$ are the minimum element in the gain matrix  $K_P$ and $K_D$ respectively, while $k_{D_{\mathrm{max}}}$ is the maximum element in the gain matrix   $K_D$. Also note that $\sigma_{max}(\cdot)$ denotes the maximum singular value of a matrix.

In terms of $\bm{X}$,  $((R^{\mathrm{des}} \bm{b}_3)^{\mathrm{T}}  R  \bm{b}_3 )R \bm{b}_3  -  R^{\mathrm{des}} \bm{b}_3 = (R \bm{b}_3) \times ( (R \bm{b}_3) \times (R^{\mathrm{des}} \bm{b}_3))$,  and $\|(R \bm{b}_3) \times ( (R \bm{b}_3) \times (R^{\mathrm{des}} \bm{b}_3))\|$ represents the sine angle between $R \bm{b}_3$ and $R^{\mathrm{des}} \bm{b}_3$. Besides we assume the following constraint is available for translational motion:
\begin{eqnarray}
  \label{eq:acc-cond1}
  \| mK_{I_r} \bm{e}_{I_r} + m\ddot{\bm r}^{\mathrm{des}}  - R Q_{\mathrm{tran}} Q^{\#}_{\mathrm{rot}}  I^{-1} \bm{\omega} \times I \bm{\omega} \| < O
\end{eqnarray}
Thus following constraints are available for $X$:
\begin{eqnarray}
  \label{eq:X-cond1}
  \|\bm{X}\| &\leq& (mk_{P_{\mathrm{max}}} \| \bm{e}_{\mathrm{r}} \| + mk_{D_{\mathrm{max}}} \| \dot{\bm{e}}_{\mathrm{r}} \|  + O) \gamma ; \gamma \in (0,1) \\
  \label{eq:X-cond2}
  \|\bm{X}\| &\leq& (mk_{P_{\mathrm{max}}} \| \bm{e}_{\mathrm{r}} \| + mk_{D_{\mathrm{max}}} \| \dot{\bm{e}}_{\mathrm{r}} \|  + O) \| \bm{e}_{\mathrm{rot}} \|
\end{eqnarray}
Substituting \equref{eq:X-cond1} and \equref{eq:X-cond2} into   \equref{eq:pos-dot-lyapunov}:
\begin{eqnarray}
  \label{eq:pos-dot-lyapunov3}
  \dot{\mathcal{V}}_2
  \leq & -ck_{P_{\mathrm{min}}} \|\bm{e}_{r}\|^2 +  (c - k_{D_{\mathrm{min}}}) \|\dot{\bm{e}}_{r}\|^2 + ck_{D_{\mathrm{max}}} \|\bm{e}_{r} \| \| \dot{\bm{e}}_{r} \| + (\|\dot{\bm{e}}_{r}\|  + c \|\bm{e}_{r}\|)  \frac{\sigma_{max}(Q^{'}_{\mathrm{tran}})}{m} \|\bar{\bm {e}}_{\mathrm{rot}}\| \nonumber \\
  &+ \frac{O }{m} \| \bm{e}_{\mathrm{rot}} \| (\|\dot{\bm{e}}_{r}\|  + c \|\bm{e}_{r}\|) +  k_{P_{\mathrm{max}}} \| \bm{e}_{\mathrm{rot}} \| \|\dot{\bm{e}}_{r}\|  \|\bm{e}_{r}\| + \gamma c k_{P_{\mathrm{max}}}   \|\bm{e}_{r}\|^2 +   \gamma c k_{D_{\mathrm{max}}} \|\dot{\bm{e}}_{r}\|  \|\bm{e}_{r}\| + \gamma k_{D_{\mathrm{max}}} \|\dot{\bm{e}}_{r}\|^2 \nonumber \\
\end{eqnarray}
Regarding the third term in \equref{eq:pos-dot-lyapunov3}, namely, $k_{P_{\mathrm{max}}} \| \bm{e}_{\mathrm{rot}} \| \|\dot{\bm{e}}_{r}\|  \|\bm{e}_{r}\|$, a upper bound for the position error is introduced to reduce the order: $ \|\bm{e}_{r} \| < \bm{e}_{r_{\mathrm{max}}}$.  Finally, the upper bound of $\dot{\mathcal{V}}_2$ can be given by
\begin{eqnarray}
  \label{eq:pos-dot-lyapunov4}
  \dot{\mathcal{V}}_2 &\leq&   (-ck_{P_{\mathrm{min}}} + \gamma c k_{P_{\mathrm{max}}}) \|\bm{e}_{r}\|^2 +  (c - k_{D_{\mathrm{min}}} + \gamma k_{D_{\mathrm{max}}}) \|\dot{\bm{e}}_{r}\|^2 + (1 + \gamma)ck_{D_{\mathrm{max}}} \|\bm{e}_{r} \| \| \dot{\bm{e}}_{r} \| \nonumber \\
  &+& (\|\dot{\bm{e}}_{r}\|  + c \|\bm{e}_{r}\|)  \frac{\sigma_{max}(Q^{'}_{\mathrm{tran}}) + O}{m} \|\bm{e}_{\mathrm{rot}}\|  + k_{P_{\mathrm{max}}} \bm{e}_{r_{\mathrm{max}}} \|\dot{\bm{e}}_{r}\| \| \bm{e}_{\mathrm{rot}} \|
\end{eqnarray}

Then, the integral Lyapunov candidate for the complete system as shown in \equref{eq:integral_lyapunov_candidate} is rewritten with $ \bm{z}_1 = \left[\begin{array}{cc}\|\bm{e}_{r} \| & \|\dot{\bm{e}}_{r} \|  \end{array}\right]^{\mathrm{T}} \in \mathsf{R}^2$, $ z_2 = \| \bm{e}_{\mathrm{rot}} \|$. The lower bound of $\mathcal{V}$ can be given by
\begin{eqnarray}
  \label{eq:complete-lyapunov}
  \bm{z}_1^{\mathrm{T}} M_1 \bm{z}_1 +  \mathcal{V}_1 + \mathcal{V}_I \leq \mathcal{V}
\end{eqnarray}
where, 
\begin{eqnarray}
  \mathcal{V}_I  = \frac{1}{2} (\bm{e}_{I_r} - \bm{\Delta} \circ \frac{1} {\bm{k}_{I_r}})^{\mathrm{T}} K_{I_r} (\bm{e}_{I_r} - \bm{\Delta} \circ \frac{1} {\bm{k}_{I_r}}), \hspace{3mm}
  M_1 = \frac{1}{2}\left[\begin{array}{cc} k_{P_{\mathrm{min}}} & -c \\ -c & 1 \end{array}\right] \nonumber
\end{eqnarray}
Note that, $M$ should be positive-defined to guarantee the $\mathcal{V}_I \geq 0$. This leads to the constraint about the constant $c$: $c < \sqrt{k_{P_{\mathrm{min}}}}$.

The time derivative of $\dot{\mathcal{V}}$ is given by
\begin{eqnarray}
  \label{eq:complete-dot-lyapunov}
  \dot{\mathcal{V}} &\leq& -\bm{z}_1^{\mathrm{T}} W_1 \bm{z}_1 + \bm{z}_1^{\mathrm{T}} W_{12} z_2 - \bm{e}_{\mathrm{rot}}^{\mathrm{T}} W_2 \bm{e}_{\mathrm{rot}} \nonumber \\
  &\leq& - \lambda_{\mathrm{min}}(W_1) \|\bm{z}_1 \|^2 +  \sigma_{\mathrm{max}}(W_{12}) \|\bm{z}_1 \| z_2 - \lambda_{\mathrm{min}}(W_2) z_2^2
\end{eqnarray}
where,
\begin{eqnarray*}
  W_1 = \frac{1}{2}\left[\begin{array}{cc} c(k_{P_{\mathrm{min}}} - \gamma k_{P_{\mathrm{max}}}) & - \frac{ck_{D_{\mathrm{max}}}}{2}(1 + \gamma) \\ - \frac{ck_{D_{\mathrm{max}}}}{2}(1 + \gamma) & -c + k_{D_{\mathrm{min}}} - \gamma k_{D_{\mathrm{max}}}) \end{array}\right],
  W_{12} = \left[\begin{array}{c} \frac{c(\sigma_{max}(Q^{'}_{\mathrm{tran}}) + O)}{m} \\  \frac{\sigma_{max}(Q^{'}_{\mathrm{tran}}) + O}{m} + k_{P_{\mathrm{max}}} \bm{e}_{r_{\mathrm{max}}} \end{array}\right],
  W_2 = - (\bar{A} +  \bar{B} K_x) \nonumber
\end{eqnarray*}
$\lambda_{\mathrm{max}}(\cdot)$  and $\lambda_{\mathrm{min}}(\cdot)$ are the maximum and minimum eigenvalue of a matrix. Note that all eigenvalues of $W_2$ are negative.

Define $\bm{z} = \left[\begin{array}{cc}\|\bm{z}_{1} \| & z_2 \|  \end{array}\right]^{\mathrm{T}} \in \mathsf{R}^2$, then \equref{eq:complete-dot-lyapunov} can be further summarized as follows:
\begin{eqnarray}
  \label{eq:complete-dot-lyapuno2}
  \dot{\mathcal{V}} &\leq& - \bm{z}^{\mathrm{T}} W \bm{z} \nonumber \\
  &\leq& - \lambda_{\mathrm{min}}(W) \|\bm{z}\|^2
\end{eqnarray}
where the matrix $W \in \mathcal{R}^{2 \times 2}$ is given by,
\begin{eqnarray}
  W = \frac{1}{2}\left[\begin{array}{cc}  \lambda_{\mathrm{min}}(W_1) &  -\frac{\sigma_{\mathrm{max}}(W_{12})}{2} \\  -\frac{\sigma_{\mathrm{max}}(W_{12})}{2} & \lambda_{\mathrm{min}}(W_2) \end{array}\right] \nonumber
\end{eqnarray}
In order to guarantee $\dot{\mathcal{V}} < 0$,  $W$ is required to be positive-defined. In other words, all eigenvalues should be positive. This derives following constraints:
\begin{align}
  \label{eq:W_1_postive_defined}
  &\lambda_{\mathrm{min}}(W_1) > 0  \\
  &\lambda_{\mathrm{min}}(W_1) \lambda_{\mathrm{min}}(W_2) > \frac{\sigma^2_{\mathrm{max}}(W_{12})}{4}  \tag{\ref{eq:gains_constraints}}
\end{align}
Note that \equref{eq:W_1_postive_defined} also implies $W_1$ should be positive defined. Therefore, the constraints regarding  positive constant $c$  and control gains can be given by
\begin{align}
  &k_{P_{\mathrm{min}}} > \gamma k_{P_{\mathrm{max}}}  \tag{\ref{eq:kp_constraints}} \\
  &c < \mathrm{min}\{\frac{4(k_{P_{\mathrm{min}}} - \gamma k_{P_{\mathrm{max}}})(k_{D_{\mathrm{min}}} - \gamma k_{D_{\mathrm{max}}})}{k_{D_{\mathrm{max}}}^2 (1+\gamma)^2 + 4(k_{P_{\mathrm{min}}} - \gamma k_{P_{\mathrm{max}}})} ,  k_{D_{\mathrm{min}}} - \gamma k_{D_{\mathrm{max}}}, \sqrt{k_{P_{\mathrm{min}}}}\}  \tag{\ref{eq:c_constraints}}
\end{align}

With the given constraints \equref{eq:gains_constraints} $\sim$ \equref{eq:c_constraints},  the zero equilibrium of the tracking errors is exponentially stable with respect to $\bm{e}_{r}, \dot{\bm{e}}_{r}, \bm{e}_{x}$, and the integral terms $\bm{e}_{I_x}, \bm{e}_{I_r}$ are uniformly bounded.


\bibliographystyle{apalike}
\bibliography{main}

\end{document}